%% file: ms.tex
\documentclass{article}

\usepackage{fullpage}
\usepackage{hyperref}       
\usepackage{url}            

\usepackage{amsmath,amsfonts,amssymb,amsthm}
\usepackage{tabularx}
\usepackage{graphicx}
\usepackage{mathrsfs}
\usepackage{algorithm, algorithmicx}
\usepackage[noend]{algpseudocode}

\usepackage{natbib}
\usepackage{color,cases}
\usepackage{tikz}
\usepackage{xspace}
\usepackage{enumerate}
\usepackage{nameref}
\usepackage{bbm}
\usepackage{thmtools,thm-restate}
\usetikzlibrary{calc,shapes}
\usepackage{subfigure}
\usepackage{hyperref}
\usepackage{multirow}

\newtheorem{claim}{Claim}

\newtheorem{lemma}{Lemma}

\newtheorem{theorem}{Theorem}
\theoremstyle{definition}

 \newcommand{\IGNORE}[1]{}

\newcommand{\eat}[1]{}

\newcommand{\calE}{{\mathcal E}}

\newcommand{\inner}[2]{\left\langle #1, #2 \right\rangle}
\newcommand{\ns}[1]{\left\| #1 \right\|^2}
\newcommand{\n}[1]{\left\| #1 \right\|}
\newcommand{\fns}[1]{\left\| #1 \right\|^2_F}
\newcommand{\fn}[1]{\left\| #1 \right\|_F}
\newcommand{\floor}[1]{\lfloor #1 \rfloor}

\newcommand{\indic}[1]{\mathbbm{1}_{#1}}

\newcommand{\diag}{\mbox{diag}}

\newcommand{\indicc}{\indic{c_i''=\sqrt{d(m+K)}/\n{u_i''}}}
\newcommand{\indiccb}{\indic{c_i''=1/\n{u_i''}}}
\newcommand{\indiccj}{\indic{c_j''=\sqrt{d(m+K)}/\n{u_j''}}}
\newcommand{\indiccbj}{\indic{c_j''=1/\n{u_j''}}}

\newcommand{\indiccU}{\indic{c_i=\sqrt{d(m+K)}/\n{u_i}}}
\newcommand{\indiccbU}{\indic{c_i=1/\n{u_i}}}

\newcommand{\ol}[1]{\overline{#1}}

\newcommand{\pr}[1]{\left( #1 \right)}
\newcommand{\br}[1]{\left[ #1 \right]}

\newcommand{\absr}[1]{\left| #1 \right|}

\newcommand{\poly}{\text{poly}}

\newcommand{\mat}{\text{mat}}
\newcommand{\var}{\text{Var}}
\newcommand{\hessian}{\mathcal{H}}
\newcommand{\hu}{\hat{u}}
\newcommand{\vvv}{\text{vec}}

        {\hspace*{\fill}$\Box$\par\vspace{4mm}}
\newenvironment{proofof}[1]{\smallskip\noindent{\bf Proof of #1.}}%
        {\hspace*{\fill}$\Box$\par}

\def\shownotes{1}  \ifnum\shownotes=1
\newcommand{\authnote}[2]{{[#1: #2]}}
\else
\newcommand{\authnote}[2]{}
\fi

\newcommand{\xnote}[1]{{\color{orange}\authnote{XW}{#1}}}
\hypersetup{
    colorlinks=true,       
    linkcolor=black,    
    citecolor=black,        
    filecolor=magenta,     
    urlcolor=blue
}

\newcommand\E{\mathbb{E}}
\newcommand\R{\mathbb{R}}
\newcommand\N{\mathcal{N}}

\makeatletter
\newcommand{\printfnsymbol}[1]{%
  \textsuperscript{\@fnsymbol{#1}}%
}
\makeatother

\title{Beyond Lazy Training for Over-parameterized Tensor Decomposition}

\author{\textbf{Xiang Wang}\thanks{Equal contribution.} \\ Duke University \\ \texttt{xwang@cs.duke.edu}  \and \textbf{Chenwei Wu}\printfnsymbol{1} \\ Duke University \\ \texttt{cwwu@cs.duke.edu} \and \textbf{Jason D. Lee} \\ Princeton University \\ \texttt{jasonlee@princeton.edu} \and \textbf{Tengyu Ma} \\ Stanford University \\ \texttt{tengyuma@stanford.edu} \and \textbf{Rong Ge} \\ Duke University \\ \texttt{rongge@cs.duke.edu} }

\date{}

\begin{document}
\maketitle

\begin{abstract}
Over-parametrization is an important technique in training neural networks. In both theory and practice, training a larger network allows the optimization algorithm to avoid bad local optimal solutions. In this paper we study a closely related tensor decomposition problem: given an $l$-th order tensor in $(\R^d)^{\otimes l}$ of rank $r$ (where $r\ll d$), can variants of gradient descent find a rank $m$ decomposition where $m > r$? We show that in a lazy training regime (similar to the NTK regime for neural networks) one needs at least $m = \Omega(d^{l-1})$, while a variant of gradient descent can find an approximate tensor when $m = O^*(r^{2.5l}\log d)$. Our results show that gradient descent on over-parametrized objective could go beyond the lazy training regime and utilize certain low-rank structure in the data.
\end{abstract}

\section{Introduction}

The success of training neural networks has sparked theoretical research in understanding non-convex optimization. Over-parametrization \--- using more neurons than the number of training data or than what is necessary for expressivity 
\--- 
is crucial to the success of optimizing neural networks~\citep{livni2014computational,jacot2018neural,mei2018mean}. The idea of over-parametrization also applies to other related or simplified problems, such as matrix factorization and tensor decomposition, which are of their own interests and also serve as testbeds for analysis techniques of non-convex optimization. We focus on over-parameterized tensor decomposition in this paper (which is closely connected to over-parameterized neural networks~\citep{ge2017learning}).

Concretely, given an order-$l$ symmetric tensor $T^*$ in $(\R^{d})^{\otimes l}$ with rank $r$, 
we aim to decompose it into a sum of rank-1 tensors with as few components as possible. 
Finding the low-rank decomposition with the smallest possible rank $r$ is known to be NP-hard \citep{hillar2013most}. The problem becomes easier if we relax the goal to finding a decomposition with $m$ components where $m$ is allowed to be larger than $r$. The natural approach is to optimize the following objective using gradient descent
\begin{equation}
\min_{u_i\in \R^d, c_i\in \R}\fns{\sum_{i=1}^m c_i u_i^{\otimes l}-\sum_{i=1}^r c_i^* [u_i^*]^{\otimes l}}.\label{obj-intro}
\end{equation}

When $m=r$, gradient descent on the objective above will empirically get stuck at a bad local minimum even for orthogonal tensors~\citep{ge2015escaping}. On the other hand, when $m =\Omega( d^{l-1})$, gradient descent provably converges to a global minimum near the initialization. This result follows straightforwardly from the Neural Tangent Kernel (NTK) technique \citep{jacot2018neural}, which was originally developed to analyze neural network training, and is referred to as the ``lazy training'' regime because essentially the algorithm is optimizing a convex function near the initialization~\citep{chizat2018note}.

The main goal of this paper is to understand whether we can go beyond the lazy training regime for the tensor decomposition problem via  better algorithm design  and analysis. In other words, we aim to use a much milder over-parametrization than  $m = \Omega(d^{l-1})$ and still converge to the global minimum of objective~\eqref{obj-intro}. We view the problem as an important first step towards analyzing neural network training beyond the lazy training regime.

We build upon the technical framework of mean-field analysis~\citep{mei2018mean}, which was developed to analyze overparameterized neural networks. It allows the parameters to move far away from the initialization and therefore has the potential to capture the realistic training regime of neural networks. However, to date, all the provable optimization results with mean-field analysis essentially operate in the infinite or exponential overparameterization regime~\citep{chizat2018global, wei2019regularization}, and applying these techniques to our problem naively would require $m$ to be exponentially large in $d$, which is even worse than the NTK result. The exponential dependency is \textit{not surprising} because the mean-field analyses in~\citep{chizat2018global, wei2019regularization} do not leverage or assume any particular structures of the data so they fail to produce polynomial-time guarantees on the worst-case data. Motivated by identifying problem structure that allows for polynomial-time guarantees, we study the mean-field analysis applied to tensor decomposition.

The main contribution of this paper is to attain nearly dimension-independent over-parametrization for the mean-field analysis in~\cite{wei2019regularization} by leveraging the particular structure of the tensor decomposition problem, 
and to show that with  $m = O^*(r^{2.5l}\log d)$, a modified version of gradient descent on a variant of objective~\eqref{obj-intro} converges to the global minimum and recovers the ground-truth tensor. 
This is a significant improvement over the NTK requirement of $m= \Omega(d^{l-1})$ and an exponential improvement upon the existing  mean-field analysis that requires $m = \exp(d)$. Our analysis shows that unlike the lazy training regime, gradient descent with small initialization and appropriate regularizer can identify the subspace that the ground-truth components lie in, and automatically exploit such structure to reduce the number of necessary components. As shown in \citet{ge2017learning}, the population-level objective of two-layer networks is a mixture of tensor decomposition objectives with different orders, so our analysis may be extendable to improve the over-parametrization necessary in analysis of two-layer networks.  
 

\subsection{Related work}

\paragraph{Neural Tangent Kernel}
There has been a recent surge of research on connecting neural networks trained via gradient descent with the neural tangent kernel (NTK)~\citep{jacot2018neural,du2018gradientb,du2018gradient,chizat2018note,allen2018learning,arora2019exact,arora2019fine,zou2019improved,oymak2020towards}. This line of analysis proceeds by coupling the training dynamics of the nonlinear network with the training dynamics of its linearization in a local neighborhood of the initialization, and then analyzing the optimization dynamics of the linearization which is convex.

Though powerful and applicable to any function class including tensor decomposition, NTK is not yet a completely satisfying theory for explaining the success of over-parametrization in deep learning. Neural tangent kernel analysis is essentially dataset independent and requires at least number of neurons $m \ge \frac{n}{d} = d^{l-1}$ to find a global optimum~\citep{zou2019improved,daniely2019neural}\footnote{Here $n$ is the number of samples which is effectively $\Theta(d^l)$ in our setting.}.

\paragraph{Beyond NTK approach}
 The gap between linearized models and the full neural network has been established in theory by~\citep{wei2019regularization,allen2019can,yehudai2019power,ghorbani2019limitations,dyer2019asymptotics,woodworth2020kernel} and observed in practice~\citep{chizat2018note,arora2019exact,lee2019wide}. Higher-order approximations of the gradient dynamics such as Taylorized Training~\citep{bai2019beyond,bai2020taylorized} and the Neural Tangent Hierarchy~\citep{huang2019dynamics} have been recently proposed towards closing this gap. Unlike this paper, existing results mostly try to improve the sample complexity instead of the level of over-parametrization for the NTK approach.

\paragraph{Mean field approach}
 For two-layer networks, a series of works used the mean field approach to establish the evolution of the network parameters~\citep{mei2018mean,chizat2018global,wei2018margin,rotskoff2018neural,sirignano2018mean}.
In the mean field regime, the parameters move significantly from
their initialization, unlike NTK regime, so it is \emph{a priori} possible for the mean field approach to exploit data-dependent structure to utilize fewer neurons. However the current analysis techniques for mean field approach need either exponential in dimension or exponential in time number of neurons to attain small training error and do not exploit any data structure. One of the main contributions of our work is to show that gradient descent can benefit from the low-rank structure in $T^\ast$.

\paragraph{Tensor decomposition} Tensor decomposition is in general an NP-hard problem~\citep{hillar2013most}. There are many algorithms that find the exact decomposition (when $m = r$) under various assumptions. In particular Jenrich's algorithm \citep{harshman1970foundations} works when $r\le d$ and the components are linearly independent. In our setting, the components may not be linearly independent, this is similar to the overcomplete tensor decomposition problem. Although there are some algorithms for overcomplete tensor decomposition (e.g., \citet{cardoso1991super,ma2016polynomial}), they require nondegeneracy conditions which we are not assuming. When the number of components $m$ is allowed to be larger than $r$, one can use spectral algorithms to find a decomposition where $m = \Theta(r^{l-1})$. In this paper our focus is to achieve similar guarantees using a direct optimization approach.

\paragraph{Neural network with polynomial activations} Another model that sits between tensor decomposition and standard ReLU neural network is neural network with polynomial activations. \citet{livni2013algorithm} gave an algorithm for training network with quadratic activations with specific algorithm. \citet{andoni2014learning} gave a way to learn degree $l$ polynomials over $d$ variables using $\Omega(d^l)$ neurons, which is similar to the guarantee of (much later) NTK approach.

\section{Notations}\label{sec:notation}

We use $[n]$ as a shorthand for $\{1,2,\cdots,n\}$. We use $O(\cdot),\Omega(\cdot)$ to hide constant factor dependencies. We use $O^*(\cdot)$ to hide constant factors and also the dependency on accuracy $\epsilon.$ We use $\poly(\cdot)$ to represent a polynomial on the relevant parameters with constant degree.

\noindent\textbf{Tensor notations:} We use $\otimes$ as the tensor product (outer product). An $l$-th order $d$-dimensional tensor is defined as an element in space $\R^d\otimes \dots \otimes \R^d$, succinctly denoted as $(\R^d)^{\otimes l}$. 
For any $i_1,\cdots,i_l\in [d],$ we use $T_{i_1,\cdots,i_l}$ to refer to the $(i_1,\cdots,i_l)$-th entry of $T\in(\R^d)^{\otimes l}$ with respect to the canonical basis. For a vector $v\in\R^{d},$ we define $v^{\otimes l}$ as a tensor in $(\R^d)^{\otimes l}$ such that $\pr{v^{\otimes l}}_{i_1,\cdots,i_l}=v_{i_1}v_{i_2}\cdots v_{i_l}.$
A tensor is symmetric if the entry values remain unchanged for any permutation of its indices. We define $\text{vec}(\cdot)$ to be the vectorize operator for tensors, mapping a tensor in $(\R^d)^{\otimes l}$ to a vector in $\R^{d^l}$: $\vvv(T)_{(i_1-1)d^{l-1}+(i_2-1)d^{l-2}+\cdots+(i_{l-1}-1)d+i_l}:=T_{i_1,i_2,\cdots,i_l}$.


A tensor $T\in (\R^d)^{\otimes l}$ is rank-1 if it can be written as $T=w\cdot v_1\otimes v_2\otimes\cdots\otimes v_l$ for some $w\in\R\text{ and }v_1,\cdots,v_l\in \R^d$, and the rank of a tensor is defined as the minimum integer $k$ such that this tensor equals the sum of $k$ rank-1 tensors.

\paragraph{Norm and inner product:} We use $\n{v}$ to denote the $\ell_2$ norm of a vector $v$. For $l$-th order tensors $T,T'\in (\R^d)^{\otimes l}$ (vectors and matrices can be viewed as tensors with order $1$ and $2$, respectively), we define the inner product as $\inner{T}{T'}:=\sum_{i_1,\cdots,i_l\in[d]}T_{i_1,\cdots,i_l}T_{i_1,\cdots,i_l}'$ and the Frobenius norm as $\fn{T}=\sqrt{\sum_{i_1,\cdots,i_l\in[d]}T_{i_1,\cdots,i_l}^2}$.

\section{Problem setup and challenges}
In this section we discuss the objective for over-parameterized tensor decomposition and explain the challenges in optimizing this objective.

We consider tensor decomposition problems with general order $l\geq 3$. Throughout the paper we consider $l$ as a constant. Specifically, we assume that the ground-truth tensor is $T^*$ of rank at most $r$:
\[
T^* := \sum_{i=1}^r c_i^* [u_i^*]^{\otimes l},
\]
where $\forall i\in[r], c_i^*\in \R \text{ and } u_i^*\in\R^d$.
Without loss of generality, we assume that $\fn{T^*}=1$. We focus on the low rank setting where $r$ is much smaller than $d$. Note we don't assume that $u_i^*$'s are linearly independent. 

The vanilla over-parameterized model we use consists of $m$ components (where $m\ge r$):
\[
T_v := \sum_{i=1}^m c_i u_i^{\otimes l},
\]
where $\forall i\in[m], c_i\in\R\text{ and }u_i\in\R^d$. We use $U\in\mathbb{R}^{d\times m}$ to denote the matrix whose $i$-th column is $u_i$, and denote $C\in \R^{m\times m}$ as the diagonal matrix with $C_{i,i}=c_i$. The vanilla loss function we are considering is the square loss:
\begin{equation}
f_v(U,C) = \frac{1}{2}\fns{T_v-T^*}=\frac{1}{2}\fns{\sum_{i=1}^m c_i u_i^{\otimes l}-\sum_{i=1}^r c_i^* [u_i^*]^{\otimes l}}.\label{eq:vanillaloss}
\end{equation}

In other words, we are looking for a rank $m$ approximation to a rank $r$ tensor.
When $m = r$, the problem of finding a decomposition is known to be NP-hard\cite{}. Therefore, our goal is to get a small objective value with small $m$ (which corresponds to the rank of $T_v$).
In the following sub-sections, we will see that there are many challenges to directly optimize the vanilla over-parameterized model over the vanilla objective, so we will need to modify the parametrization of the tensor $T_v$ and the optimization algorithm to overcome them.

\subsection{Challenge 0: lazy training requires immense over-parameterization}

We show lazy training requires $\Omega(d^{l-1})$ components to fit a rank-one tensor in the following theorem.

\begin{restatable}
{theorem}{kernel}\label{thm:kernel}
Suppose the ground truth tensor $T^*=[u^*]^{\otimes l}$, where $u^*$ is uniformly sampled from the unit sphere $\mathbb{S}^{d-1}$. Lazy training (defined as below) requires $\Omega(d^{l-1})$ components to achieve $o(1)$ error in expectation.
\end{restatable}

In the lazy training regime, all the $u_i$'s stay very close to the initialization. Assuming the final $u_i'$ is equal to $u_i+\delta_i$, all the higher-order terms in $\delta_i$ can be ignored. Therefore, the model can only capture tensors in the linear subspace $S_U = \text{span}\{P_{sym}\text{vec}(u_i^{\otimes l-1}\otimes \delta_i)\}_{i=1}^m$ (here $P_{sym}$ is the projection to the space of vectorized symmetric tensors, $u_i$'s are the initialization and $\delta_i$'s are arbitrary vectors in $\R^d$). The dimension of this subspace is upperbounded by $dm$. Let $W_l$ be the space of all vectorized symmetric tensors in $(\R^d)^{\otimes l}$ (with dimension $\Omega(d^{l})$), and $S_U^\perp$ be the subspace of $W_l$ orthogonal to $S_U$. We show that for a random rank-1 tensor $T^*$, it will often have a large projection in $S_U^\perp$ unless $m = \Omega(d^{l-1})$. Basically, the subspace $S_U$ has to cover the whole space $W_l$ to approximate a random rank-1 tensor.
The proof of Theorem~\ref{thm:kernel} is in Appendix~\ref{sec:kernel}.



\subsection{Challenge 1: zero is a high-order saddle point for vanilla objective}
\label{sec:challenge:saddle}
As \citet{chizat2018note} pointed out, lazy training regime corresponds to the case where the initialization has large norm. A natural way to get around lazy training is to use a much smaller initialization. However, for the vanilla objective, 0 will be a saddle point of order $l$ on the loss landscape. This makes gradient descent really slow at the beginning. In Section~\ref{sec:alg}, we  fix this issue by re-parameterizing the model into a 2-homogeneous model. 


\subsection{Challenge 2: existence of bad local minima far away from 0}
\label{bad_local_min}

It was shown that no bad local minima exist in matrix decomposition problems \citep{ge2016matrix}. Therefore, (stochastic) gradient descent is guaranteed to find a global minimum. In this section, we show that in contrary tensor decomposition problems with order at least 3 have bad local minima.



\begin{restatable}
{theorem}{badLocalMin}\label{thm:bad_local_min}
Let $f_v(U,C)$ be as defined in Equation~\ref{eq:vanillaloss}. Assume $l\geq 3, d > r\geq 1$ and $m\geq r(l+1)+1.$ There exists a symmetric ground truth tensor $T^*$ with rank at most $r(l+1)+1$ such that a local minimum with function value $l(l-1)r/4$ exists while the global minimum has function value zero.
\end{restatable}


In the construction, we set all the $u_i$'s to be $e_1/m^{1/l}$ so that $T=e_1^{\otimes l}.$ We define the ground truth tensor by setting the residual $T-T^*$ to be $\sum_{j=2}^{r+1}e_j^{\otimes 2}\otimes e_1^{\otimes l-2}$ plus its $\binom{l}{2}$ permutations. At this point, the gradient equals zero, so there is no first order change to the function value. Furthermore, we show if any component moves in one of the missing direction $e_j$ for $2\leq j\leq r+1,$ it will incur a second order function value increase. So the tensor can only moves along $e_1$ direction, which cannot further decrease the function value because $e_1^{\otimes l}$ is orthogonal with the residual. Note this is a bad local min but not a strict bad local min because we can shrink one component to zero and meanwhile increase another component so that the tensor does not change. When we have a zero component (it's a saddle point), we can add a missing direction to decrease the function value. 

In Appendix~\ref{sec:proof_local_min}, we prove Theorem~\ref{thm:bad_local_min} and also construct a bad local minimum for 2-homogeneous model defined in Section~\ref{sec:alg}.
To escape these spurious local minima, our algorithm re-initializes one component after a fixed number of iterations.

\section{Algorithms and main results}
\label{sec:alg}
In this section, we introduce our main algorithm, a modified version of gradient descent on a non-convex objective, and state our main results. 

To address the high-order saddle point issue in Section~\ref{sec:challenge:saddle}, we introduce a new variant of the parameterized models. 
\[
T := \sum_{i=1}^m a_i c_i^{l-2} u_i^{\otimes l},
\]
where $\forall i\in[m], a_i\in\{-1,1\}, c_i=\frac{1}{\n{u_i}}\text{ and }u_i\in\R^d$.

Note that since $u_i^{\otimes l}$ is homogeneous, there is a redundancy in the vanilla parametrization between the coefficient and the norm of $\|u\|$. Here we do the rescaling to make sure that the model $T$ is a 2-homogeneous function of $u_i$'s. 
Using the new formulation of $T$, 0 will no longer be a high order saddle point.  

Recall that we use $U\in\mathbb{R}^{d\times m}$ to denote the matrix whose $i$-th column is $u_i$. We use $C,A\in\mathbb{R}^{m\times m}$ to denote the diagonal matrices with $C_{ii}=c_i, A_{ii}=a_i$. The loss function we are considering is the square loss plus a regularization term:
\[
f(U,C,\hat{C},A) \triangleq \frac12\fns{\sum_{i=1}^m a_i c_i^{l-2} u_i^{\otimes l}-T^*} + \lambda\sum_{i=1}^m \hat{c}_i^{l-2} \n{u_i}^l,
\]
where $\forall i\in[m], \hat{c}_i\in\mathbb{R}^+$ and we use $\hat{C}\in\mathbb{R}^{m\times m}$ to denote the diagonal matrix with $\hat{C}_{ii}=\hat{c}_i$. For simplicity, we use $\bar{C}$ to denote the tuple $(C,\hat{C},A)$. Therefore, we can write $f(U,C,\hat{C},A)$ as $f(U,\bar{C})$. 

The algorithm contains $K$ epochs, where each epoch includes $H$ iterations. 
At the initialization, we independently sample each $u_i$ from $\delta\text{Unif}(\mathbb{S}^{d-1}),$ where the radius $\delta$ will be set to be $\poly(\epsilon,1/d).$

Denote the subspace of $\mbox{span}\{u_i^*\}$ as $S$ and its orthogonal subspace in $\R^d$ as $B$. Let $P_S, P_B$ be the projection matrices onto subspace $S$ and $B$, respectively. Since the components of the ground-truth tensor lies in the subspace $S$, ideally we want to make sure that the components of tensor $T$ lies in the same subspace $S$. We also want to make sure $c_i$ roughly equals $1/\|P_S u_i\|$ to ensure the improvement in $S$ subspace is large enough. However, the algorithm does not know the subspace $S$. We address this problem using the observation that $\|P_S u_i\| \approx \frac{\sqrt{r}}{\sqrt{d}}\|u_i\|$ at initialization; and $\|P_S u_i\|\approx \|u_i\|$ if norm of $u_i$ is large, but its projection in $B$ has not grown larger. In our algorithm we introduce a "scalar mode switch" step between these two regimes by the separation between $C$ and $\hat{C}$: For the $i$-th component, the coefficients $c_i$ and $\hat{c}_i$ are initialized as $\sqrt{d(m+K)}/\n{u_i}$ and $1/\n{u_i}$, respectively, and we reduce $c_i$ by a factor of $\sqrt{d(m+K)}$ ($c_i$ will be equal to $\hat{c}_i$ afterwards) when $\n{u_i}$ exceeds $2\sqrt{m+K}\delta$ for the first time. For each $i\in[m], a_i$ is i.i.d. sampled from Unif$\{1,-1\}$.

We also re-initialize one component at the beginning of each epoch. At each iteration, we first update $U$ by gradient descent: $U'\leftarrow U-\eta\nabla_U f(U,\bar{C})$. Note that when taking the gradient over $U$, we treat $c_i$'s and $\hat{c}_i$'s as constants. Then we update each $c_i$ and $\hat{c}_i$ using the updated value of $u_i$ to preserve 2-homogeneity, i.e., $c_i' = \frac{\n{u_i}}{\n{u_i'}}c_i$ and $\hat{c}_i' = \frac{\n{u_i}}{\n{u_i'}}\hat{c}_i$. 
We fix $a_i$'s during the algorithm except for the initialization and re-initialization steps. 

The pseudocode is given in Algorithm~\ref{alg1}. Using this variant of gradient descent, we can recover the ground truth tensor $T^*$ with high probability using only $O\pr{\frac{r^{2.5l}}{\epsilon^5}\log (d/\epsilon)}$ number of components. The formal theorem is stated below.

\begin{algorithm}[htbp]
\caption{Variant of Gradient Descent for Tensor Decomposition}
\label{alg1}
\begin{algorithmic}[1.]
\State \textbf{Input:} number of epochs $K$, number of iterations in one epoch $H$, initialization size $\delta$, step size $\eta$.
\State For each $i\in [m],$ initialize $u_i$ i.i.d. from $\delta \text{Unif}(\mathbb{S}^{d-1})$; initialize $a_i$ i.i.d. from $\text{Unif}\{1,-1\}$; initialize $c_i$ as $\frac{\sqrt{d(m+K)}}{\n{u_i}}$ and $\hat{c}_i$ as $\frac{1}{\n{u_i}}$.
\For{epoch $k:=1$ to $K$}
\State Let $u_j$ be any vector with the smallest $\ell_2$ norm among all columns of $U$.
\State Re-initialize $u_j$ from $\delta \text{Unif}(\mathbb{S}^{d-1})$, re-initialize $a_j$ from $\text{Unif}\{1,-1\}$ and set $c_j=\frac{\sqrt{d(m+K)}}{\n{u_j}}, \hat{c}_j=\frac{1}{\n{u_j}}$.\\
\For {iteration $t:=1$ to $H$}
\State $U'\leftarrow U-\eta\nabla_U f(U,\bar{C})$.
\For {$i:=1$ to $m$}
\State $c_i'\leftarrow \frac{\|u_i\|}{\|u_i'\|}c_i;$ $\hat{c}_i'\leftarrow \frac{\|u_i\|}{\|u_i'\|}\hat{c}_i$.
\If {$\|u_i\|\leq 2\sqrt{m+K}\delta<\|u_i'\|$ holds for the first time since it was (re)-initialized}
\State $c_i' \leftarrow  \frac{c_i'}{\sqrt{d(m+K)}}$. \Comment{Scalar Mode Switch}
\EndIf
\EndFor
\State $U\leftarrow U', C\leftarrow C', \hat{C}\leftarrow\hat{C}$.
\EndFor
\EndFor
\State \textbf{Output:} $T := \sum_{i=1}^m a_i c_i^{l-2} u_i^{\otimes l}$.
\end{algorithmic}
\end{algorithm}

\begin{restatable}
{theorem}{mainThm}\label{thm:main}
Given any target accuracy $\epsilon>0,$ there exists $m=O\pr{\frac{r^{2.5l}}{\epsilon^5}\log(d/\epsilon)},\lambda=O\pr{\frac{\epsilon}{r^{0.5l}}},\delta=\poly(\epsilon,1/d),\eta=\poly(\epsilon,1/d),H=\poly(1/\epsilon,d)$ such that with probability at least $0.99,$ our algorithm finds a tensor $T$ satisfying
\[\fn{T-T^*}\leq \epsilon,\]
within $K=O\pr{\frac{r^{2l}}{\epsilon^4}\log(d/\epsilon)}$ epochs.
\end{restatable}

\section{Summary of our techniques}
\label{short_proof_sketch}

In this section, we discuss the high-level ideas that we need to prove Theorem~\ref{thm:main}. The full proof is deferred into Appendix~\ref{sec:full_proof}.

Generally, doing gradient descent never increases the objective value (though this is not obvious for our algorithm as it is slightly different in handling the normalization $c_i, \hat{c}_i$'s). Our main concern is to address Challenge 2, namely, the algorithm might get stuck at a bad local minimum. We will show that this cannot happen with the re-initialization procedure.

More precisely, we rely on the following main lemma to show that as long as the objective is large, there is at least a constant probability to improve the objective within one epoch.

\begin{restatable}
{lemma}{normExplode}\label{lem:norm_explode}
In the setting of Theorem~\ref{thm:main}, let $(U_0',\bar{C}_0')$ and $(U_H,\bar{C}_H)$ be the parameters at the beginning of an epoch and the parameters at the end of the same epoch. Assume $\fn{T_0'-T^*}\geq \epsilon,$ where $T_0'$ is tensor with parameters $(U_0',\bar{C}_0')$. Then with probability at least $1/6$, we have
\[f(U_H, \bar{C}_H)-f(U_0',\bar{C}_0')= - \Omega\pr{\frac{\epsilon^4}{r^{2l}\log(d/\epsilon)}}.\]
\end{restatable}

We complement this lemma by showing that even if an epoch does not decrease the objective, it will not overly increase the objective. 

\begin{restatable}
{lemma}{fIncreaseBounded}\label{co:f_increase_bounded}
In the setting of Theorem~\ref{thm:main}, let $(U_0',\bar{C}_0')$ and $(U_H,\bar{C}_H)$ be the parameters at the beginning of an epoch and the parameters at the end of the same epoch. Assume $f(U_0',\bar{C}_0')\geq \epsilon^2$, where $\epsilon$ is the target accuracy in Theorem~\ref{thm:main}. Then, we have 
$f(U_H, \bar{C}_H)-f(U_0',\bar{C}_0')= O(\frac{1}{\lambda m}).$
\end{restatable}

From these two lemmas, we know that in each epoch, the loss function can decrease by $\Omega\pr{\frac{\epsilon^4}{r^{2l}\log(d/\epsilon)}}$ with probability at least $\frac16$, and even if we fail to decrease the function value, the increase of function value is at most $O\pr{\frac{1}{\lambda m}}$. By our choice of parameters in Theorem~\ref{thm:main}, $m=\Theta \pr{\frac{r^{2.5l}}{\epsilon^5}\log(d/\epsilon)},\lambda=\Theta \pr{\frac{\epsilon}{r^{0.5l}}}$ and then $O(\frac{1}{\lambda m})=O(\frac{\epsilon^4}{r^{2l}\log(d/\epsilon)})$. Choosing a large constant factor in $m$, we can ensure that the function value decrease will dominate the increase. This allows us to prove Theorem~\ref{thm:main}.

In the next two subsections, we will discuss how to prove Lemma~\ref{lem:norm_explode} and Lemma~\ref{co:f_increase_bounded}, respectively.

\subsection{Proof sketch for Lemma~\ref{co:f_increase_bounded} - upper bound on function increase}

To prove the increase of $f$ is bounded in one epoch, we identify all the possible ways that the loss can increase and upper bound each of them. We first show that a normal step (without scalar mode switch) of the algorithm will not increase the objective function 

\begin{lemma}\label{lem:f_value_decrease}
In the setting of Theorem~\ref{thm:main}, let $(U,\bar{C})$ be the parameters at the beginning of one iteration and let $U',\bar{C}'$ be the updated parameters (before potential scalar mode switch). Assuming $f(U,\bar{C})\leq 10,$ we have $f(U',\bar{C}')-f(U,\bar{C})\leq -\frac{\eta}{l}\fns{\nabla_U f(U,\bar{C})}.$ 
\end{lemma}
Note that we treat $C$ and $\hat{C}$ as constants when taking gradient with respect to $U$ and then update $C$ and $\hat{C}$ according to the updated value of $U$, so this lemma does not directly follow from standard optimization theory. The gradient descent on $U$ decreases the function value when the step size is small enough while updating $C,\hat{C}$ can potentially increase the function value. In order to show that overall the function value decreases, we need to bound the function value increase due to updating $C,\hat{C}$. We are able to do this because of the special regularizer we choose. In particular, our regularizer guarantees that the change introduced by updating $C$ and $\hat{C}$ is proportional to the change of the gradient step, and is smaller in scale. Therefore we maintain the decrease in the gradient step. 

Since we already know that the function value cannot increase in a normal iteration (before potential scalar mode switch), the only causes of the function value increase are the re-initialization or scalar mode switches. According to the algorithm, we only switch the scalar mode when the norm of a component reaches $2\sqrt{m+K}\delta$ for the first time, so the number of scalar mode switches in each epoch is at most $m$. Choosing $\delta$ to be small enough, the effects of scalar mode switches should be negligible. In the re-intialization, we remove the component with smallest $\ell_2$ norm, which can increase the function value by at most $O(\frac{1}{\lambda m})$. This is proved in Lemma~\ref{lem:reinit_bounded}.

\begin{lemma}\label{lem:reinit_bounded}
In the setting of Theorem~\ref{thm:main}, let $(U_0',\bar{C}_0')$ and $(U_0,\bar{C}_0)$ be the parameters before and after the reinitialization step, respectively. Assume $f(U_0',\bar{C}_0')\geq \epsilon^2$, where $\epsilon$ comes from Theorem~\ref{thm:main}. Then, we have $f(U_0, \bar{C}_0)-f(U_0',\bar{C}_0')= O(\frac{1}{\lambda m}).$
\end{lemma}

In the proof, we can show the function value is at most a constant and then $\sum_{i=1}^m \ns{u_i}=O(1/\lambda)$ due to the regularizer. Since we choose the reinitialized component $u$ as one of the component with smallest $\ell_2$ norm, we know $\ns{u}=O(\frac{1}{\lambda m}).$ This then allows us to bound the function value change from reinitialization by $O(\frac{1}{\lambda m}).$ Lemma~\ref{co:f_increase_bounded} follows from Lemma~\ref{lem:f_value_decrease} and Lemma~\ref{lem:reinit_bounded}.

\subsection{Proof sketch for Lemma~\ref{lem:norm_explode} - escaping local minima}\label{sec:escape:main}
In this section, we will show how we can escape local minima by re-initialization. 
Intuitively, we will show that when a component is randomly re-initialized, it has a positive probability of having a good correlation with the current residual $T - T^*$. However, there is a major obstacle here: because the component is re-initialized in the full $d$-dimensional space, the correlation of this new component with $T-T^*$ is going to be of the order $d^{-l/2}$. If every epoch can only improve the objective function by $d^{-l/2}$ we would need a much larger number of epochs and components.

We solve this problem by observing that both $T$ and $T^*$ are almost entirely in the subspace $S$. If we only care about the projection in $S$, the random component will have a correlation of $r^{-l/2}$ with the residual $T - T^*$. We will show that such a correlation will keep increasing until the norm of the new component is large enough, therefore decreasing the objective function. 

First of all, we need to show that the influence coming from the subspace $B$ (the orthogonal subspace of the span of $\{u_i^*\}$) is small enough so that it can be ignored.
\begin{lemma}\label{lem:bound-ortho-norm}
In the setting of Theorem~\ref{thm:main}, we have $\fns{P_BU}\leq (m+K)\delta^2$ throughout the algorithm.
\end{lemma}

We prove Lemma~\ref{lem:bound-ortho-norm} by showing the norm of $P_BU$ only increases at the (re-)initializations, so it will stay small throughout this algorithm. This lemma is also the motivation of our algorithm, i.e., we treat $C$ and $\hat{C}$ as constants when taking the gradient so that the gradient of $U$ will never have negative correlation with $P_BU$.

Now let us focus on the subspace $S$. We denote the re-initialized vector at $t$-th step as $u_t$, and its sign as $a\in \{\pm 1\}$, and we will take a look at the change of $P_Su_t$. Our analysis focuses on the correlation between $P_S u_t$ and the residual tensor:
$\langle (P_{S^{\otimes l}}T_t-T^*), a (\overline{ P_S u_t}^{\otimes l}) \rangle.$
Here $\overline{ P_S u_t}$ is the normalized version $P_S u_t$. We will show that the norm of $u_t$ will blow up exponentially if this correlation is significantly negative at every iteration.

Towards this goal, first we will show that the initial point $P_S u_0$ has a large negative correlation with the residual. We lower bound this correlation by anti-concentration of Gaussian polynomials:

\begin{lemma}\label{lem:initial_correlate}
Suppose the residual at the beginning of one epoch is $T_0'-T^*.$ Suppose $a c_0^{l-2} u_0^{\otimes l}$ is the reinitialized component. With probability at least $1/5$,
\[
\inner{P_{S^{\otimes l}}T_0'-T^*}{a\overline{ P_S u_0}^{\otimes l}} \leq -\Omega\pr{\frac{1}{r^{0.5l}}}\fn{P_{S^{\otimes l}}T_0'-P_{S^{\otimes l}}T^*},
\]
where $\ol{P_S u_0}=P_S u_0/\n{P_S u_0}.$
\end{lemma}

Our next step argues that if this negative correlation is large in every step, then the norm of $u_t$ blows up exponentially:

\begin{lemma}\label{lem:keep_correlation_exp_norm}
In the setting of Theorem~\ref{thm:main}, within one epoch, let $T_0$ be the tensor after the reinitilization and let $T_\tau$ be the tensor at the end of the $\tau$-th iteration. Assume $\n{P_S u_0}\geq \Omega(\delta/\sqrt{d}).$ For any $t\geq 1$, as long as $\inner{P_{S^{\otimes l}}T_\tau-T^*}{a\overline{ P_S u_\tau}^{\otimes l}}\leq -\Omega\pr{\frac{\epsilon}{r^{0.5l}}}$ for all $t-1\geq \tau\geq 0$, we have
\[\ns{P_Su_{t}}\geq\pr{1+\Omega\pr{\frac{\eta \epsilon}{r^{0.5l}}}}^t\ns{P_Su_0}.\]
\end{lemma}

Therefore the final step is to show that $P_S u_t$ always have a large negative correlation with $T_t - T^*$, unless the function value has already decreased. The difficulty here is that both the current reinitialized component $u_t$ and other components are moving, therefore $T_t$ is also changing. 

We can bound the change of the correlation by separating it into two terms, which are the change of the re-initialized component and the change of the residual:
\begin{align*}
&\inner{P_{S^{\otimes l}}T_t-T^*}{a\overline{ P_S u_t}^{\otimes l}}-\inner{P_{S^{\otimes l}}T_0-T^*}{a\overline{ P_S u_0}^{\otimes l}}\\
\leq& \sum_{\tau=1}^t \pr{ \inner{P_{S^{\otimes l}}T_{\tau-1}-T^*}{\overline{ P_S u_{\tau}}^{\otimes l}}-\inner{P_{S^{\otimes l}}T_{\tau-1}-T^*}{\overline{ P_S u_{\tau-1}}^{\otimes l}} }+ \sum_{\tau=1}^t\fn{T_\tau-T_{\tau-1}}.
\end{align*}

The change of the re-initialized component has a small effect on the correlation because the change in $S$ subspace can only improve the correlation, and the influence of the $B$ subspace can be bounded. This is formally proved in the following lemma. 

\begin{lemma}\label{lem:u_change}
In the setting of Theorem~\ref{thm:main}, suppose at the beginning of one iteration, the tensor $T$ has parameters $(U,\bar{C}).$ Suppose $u$ is one column vector in $U$ with $\n{P_S u}=\Omega(\frac{\delta}{\sqrt{d}})$ and $u'=u-\eta\nabla_u f(U,\bar{C}).$ We have
\[\inner{P_{S^{\otimes l}}T-T^*}{a\ol{P_S u'}^{\otimes l}}\leq \inner{P_{S^{\otimes l}}T-P_{S^{\otimes l}}T^*}{a\ol{P_S u}^{\otimes l}}+\eta\delta\poly(d),
\]
where $\poly(d)$ does not hide any dependency on $\eta,\delta.$
\end{lemma}

Therefore, the only way to change the residual term by a lot must be changing the tensor $T$, and the accumulated change of $T$ is strongly correlated with the decrease of $f$. This is similar to the technique of bounding the function value decrease in \cite{wei2019regularization}. 
The connection between them are formalized in the following lemma:

\begin{lemma}\label{lem:tensor_bounded_epoch}
In the same setting of Lemma~\ref{lem:keep_correlation_exp_norm}, within one epoch, let $(U_0,\bar{C}_0)$ be the parameters after the reinitialization step and let $(U_H,\bar{C}_H)$ be the parameters at the end of this epoch. We have
\begin{align*}
\sum_{\tau=1}^H\fn{T_\tau - T_{\tau-1}}
\leq O\pr{\sqrt{\frac{\eta H}{\lambda}}}\sqrt{f(U_0,\bar{C}_0)-f(U_H,\bar{C}_H)+\delta^2\poly(d)}+\delta^2\poly(d),
\end{align*}
where $\poly(d)$ does not hide dependency on $\delta.$
\end{lemma}

Intuitively, Lemma~\ref{lem:tensor_bounded_epoch} is true because a large accumulated change of $T$ indicates large gradients along the trajectory, which suggests a large decrease in the function value. In fact, we choose the parameters such that $\lambda = \Theta(\frac{\epsilon}{r^{0.5l}}), \eta H = \Theta(\frac{r^{0.5l}}{\epsilon}\log(d/\epsilon))$. If the accumulated change of $T$ is larger than $\Omega(\frac{\epsilon}{r^{0.5l}}),$ the function value decreases by at least $\Omega(\frac{\epsilon^4}{r^{2l}\log(d/\epsilon)}),$ as stated in Lemma~\ref{lem:norm_explode}.

Combining all the steps, we show that either the function value has already decreased (by Lemma~\ref{lem:tensor_bounded_epoch}), or the correlation remains negative and the norm $\n{P_Su_t}$ blows up exponentially (by Lemma~\ref{lem:keep_correlation_exp_norm}). The norm cannot grow exponentially because of the regularizer, so the function value must eventually decrease. This finishes the proof of Lemma~\ref{lem:norm_explode}.

\section{Conclusion}
In this paper we show that for an over-parameterized tensor decomposition problem, a variant of gradient descent can learn a rank $r$ tensor using $O^*(r^{2.5l}\log(d/\epsilon))$ components. The result shows that gradient-based methods are capable of leveraging low-rank structure in the input data to achieve lower level of over-parametrization. There are still many open problems, in particular extending our result to a mixture of tensors of different orders which would have implications for two-layer neural network with ReLU activations. We hope this serves as a first step towards understanding what structures can help gradient descent to learn efficient representations.


\section*{Acknowledgements}
RG acknowledges support from NSF CCF1704656, NSF CCF-1845171 (CAREER), NSF CCF-1934964 (TRIPODS), NSF-Simons Research Collaborations on the Mathematical and Scientific Foundations of Deep Learning, a Sloan Fellowship, and a Google Faculty Research Award. 

JDL acknowledges support of the ARO under MURI Award W911NF-11-1-0303,  the Sloan Research Fellowship, and NSF CCF 2002272. 
 This is part of the collaboration between US DOD, UK MOD and UK Engineering and Physical Research Council (EPSRC) under the Multidisciplinary University Research Initiative. 
 
 TM acknowledges support of Google Faculty Award. The work is also partially supported by SDSI and SAIL at Stanford.

\bibliography{ref,bib}
\bibliographystyle{apalike}

\newpage
\appendix

\input{kernel}
\input{proof_local_min}
\input{proof_algo}
\input{tools}

\end{document}

%% file: kernel.tex
\subsection*{Notations} Besides the notations defined in Section~\ref{sec:notation}, we also use the following notations in the proofs.

We use $\odot$ for the Khatri-Rao product. We denote $e_i$ as the $i$-th basis vector in $\R^d.$ 

We define $\text{mat}(\cdot)$ to be the matrixize operator for tensors, mapping a tensor in $(\R^d)^{\otimes l}$ to a matrix in $\R\times \R^{d^{l-1}}$: $\text{mat}(T)_{i_1,(i_2-1)d^{l-2}+\cdots+(i_{l-1}-1)d+i_l}:=T_{i_1,i_2,\cdots,i_l}$ for any $i_1,i_2,\cdots,i_l\in[d].$

We view a tensor $T\in (\R^d)^{\otimes l}$ as a multilinear form. For matrices $M_1\in\R^{d\times k_1},\cdots,M_l\in\R^{d\times k_l},$ the tensor $T(M_1,M_2,\cdots,M_l)\in\R^{k_1\times k_2\times \cdots\times k_l}$ is defined such that
\\$T(M_1,M_2,\cdots,M_l)_{j_1,\cdots,j_l}:=\sum_{i_1,\cdots,i_l\in [d]}T_{i_1,\cdots,i_l}(M_1)_{i_1,j_1}\cdots (M_l)_{i_l,j_l},$
for any $j_1\in[k_1],\cdots,j_l\in[k_l].$ For notation simplicity, we use $T(M^{\otimes k},M_{k+1},\cdots,M_{l})$ to denote $T(M,M,\cdots,M,M_{k+1},\cdots,M_l)$. In particular, for any $v\in\R^d$, $T(v^{\otimes l})$ is a scalar equals to $\inner{T}{v^{\otimes l}}=\sum_{i_1,\cdots,i_l\in [d]}T_{i_1,\cdots,i_l}v_{i_1}v_{i_2}\cdots v_{i_l}.$ 

\section{Lower Bound for the Number of Components Needed for Kernels} \label{sec:kernel}

In this section, we will prove that a lazy training model requires $\Omega(d^{l-1})$ components to fit a random rank-one tensor with $o(1)$ loss. Recall Theorem~\ref{thm:kernel} as follows.

\kernel*

Recall that in our definition, a lazy training model can only capture tensors in the linear subspace $S_U = \text{span}\{P_{sym}\text{vec}(u_i^{\otimes l-1}\otimes \delta_i)\}_{i=1}^m$ (here $P_{sym}$ is the projection to the space of vectorized symmetric tensors, $\delta_i$'s are arbitrary vectors in $\R^d$). The dimension of this subspace is upperbounded by $dm$. Let $W_l$ be the space of all vectorized symmetric tensors in $(\R^d)^{\otimes l}$, and $S_U^\perp$ be the subspace of $W_l$ orthogonal to $S_U$. We only need to show that for a random rank-one tensor, in expectation its projection on the orthogonal subspace $S_U^\perp$ is at least a constant unless $m=\Omega(d^{l-1}).$ In the following lemma, we first lower bound the projection of the ground truth tensor on a fixed direction. The proof of Lemma~\ref{lem:expectation_unit} is deferred into Section~\ref{sec:technical_proof_kernel}.

\begin{restatable}
{lemma}{expUnit}\label{lem:expectation_unit}
Let $u\in \R^d$ be a vector sampled uniformly on the unit sphere $\mathbb{S}^{d-1}$. For any vectorized symmetric $l$-th order tensor $b\in \R^{d^l}$ with unit $\ell_2$ norm, we have
\[b^\top \E[\vvv(u^{\otimes l})\vvv(u^{\otimes l})^\top]b\geq \frac{\Gamma\pr{\frac{d}{2}}}{2^l\Gamma\pr{l+\frac{d}{2}}}l!,\]
where $\Gamma(\cdot)$ is the Gamma function.
\end{restatable}

Next, we lower bound the projection of $\vvv(T^*)$ on subspace $S_U^\perp$ by summation up the projections on the subspace bases, each of which can be bounded by Lemma~\ref{lem:expectation_unit}. We give the proof of Theorem~\ref{thm:kernel} as follows.

\begin{proofof}{Theorem~\ref{thm:kernel}}
Recall that $W_l$ is the space of all vectorized symmetric tensors in $(\R^d)^{\otimes l}$. Due to the symmetry, the dimension of $W_l$ is $\binom{d+l-1}{l}$. Since the dimension of $S_U$ is at most $dm$, we know that the dimension of $S_U^\perp$ is at least $\binom{d+l-1}{l}-dm$. Assuming $S_U^\perp$ is an $\bar{m}$-dimensional space, we have $\bar{m}\geq \binom{d+l-1}{l}-dm\geq \frac{d^l}{l!}-dm$. Let $\{e_1,\cdots,e_{\bar{m}}\}$ be a set of orthonormal bases of $S_U^\perp$, and $\Pi_U^\perp$ be the projection matrix from $\R^{d^l}$ onto $S_U^\perp$, then we know that the smallest possible error that we can get given $U$ is
\[\frac12\E_{u^*}\left[\fns{\Pi_U^\perp \vvv(T^*)}\right]=\frac12\E_{u^*}\left[\sum_{i=1}^{\bar{m}}\inner{\vvv(T^*)}{e_i}^2\right]=\frac12\sum_{i=1}^{\bar{m}}\E_{u^*}\left[\inner{\vvv(T^*)}{e_i}^2\right],\]
where the expectation is taken over $u^*\sim\text{Unif}(\mathbb{S}^{d-1})$.

By Lemma~\ref{lem:expectation_unit}, we know that for any $i\in[m],$
\begin{align*}
\E_{u^*}\left[\inner{\vvv(T^*)}{e_i}^2\right]
=& e_i^\top \E_{u^*}[(\vvv([u^*]^{\otimes l})\vvv([u^*]^{\otimes l})^\top]e_i\\
\geq& \frac{\Gamma\pr{\frac{d}{2}}}{2^l\Gamma\pr{l+\frac{d}{2}}}l!\geq \mu\frac{l!}{d^l},
\end{align*}
where $\mu$ is a positive constant only related to $l$. 

Therefore,
\[\frac12\E_{u^*}\left[\fns{\Pi_U ^\perp T^*}\right]=\frac12\sum_{i=1}^{\bar{m}}\E_{u^*}\left[\inner{\vvv(T^*)}{e_i}^2\right]\geq\pr{\frac{d^l}{l!}-dm}\frac{\mu l!}{2d^l}= \frac{\mu}{2}-\frac{\mu l!}{2}\cdot\frac{m}{d^{l-1}}.\]
Note that we assume $l$ is a constant. If $m=o(d^{l-1})$, i.e., $\frac{m}{d^{l-1}}=o(1)$, then the expectation of the smallest possible error is at least some constant. Thus, if we want the error to be $o(1)$, we must have $m=\Omega(d^{l-1})$. This finishes the proof of Theorem~\ref{thm:kernel}.
\end{proofof}

\subsection{Numerical verification of the lower bound}
In this section, we plot the projection of the ground truth tensor on the orthogonal subspace $\E_{u^*}\fns{\Pi_U ^\perp T^*}$ under different dimension $d$ and number of components $m$. For convenience, we only plot the lower bound for the projection that is $\pr{\binom{d+l-1}{l}-dm}\frac{\Gamma\pr{\frac{d}{2}}}{2^l\Gamma\pr{l+\frac{d}{2}}}l!$ as we derived previously.

\begin{figure}[h]
     \centering
     \includegraphics[width=4in]{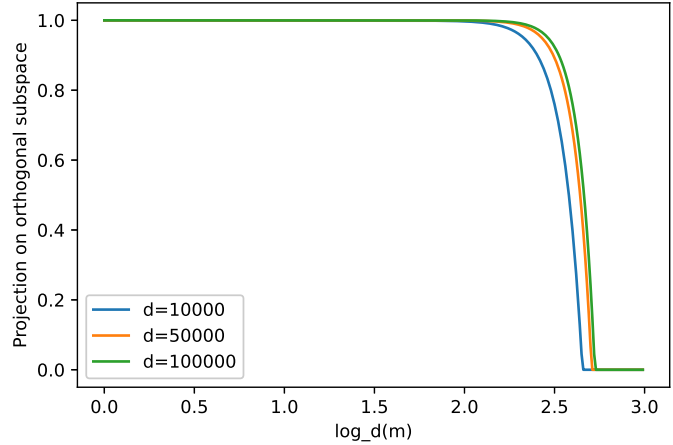}
     \caption{The projection of the ground truth tensor on the orthogonal subspace when $l=4.$}
     \label{fig:lower_bound}
\end{figure}

Figure~\ref{fig:lower_bound} shows that under different dimensions, $\E_{u^*}\fns{\Pi_U ^\perp T^*}$ is at least a constant until $\log_d m$ gets close to $l-1=3.$ As dimension $d$ increases, the threshold when the orthogonal projection significantly drops becomes closer to $3.$

\subsection{Proofs of technical lemmas}\label{sec:technical_proof_kernel}
To prove Lemma~\ref{lem:expectation_unit}, we need another technical lemma, which is stated and proved below.
\begin{lemma}\label{lem:expectation}
Let $u\in \R^d$ be a standard normal vector. For any vectorized symmetric $l$-th order tensor $b\in \R^{d^l}$ with unit norm, we have
\[b^\top \E[\vvv{(u^{\otimes l})}\vvv{(u^{\otimes l})}^\top]b\geq l!.\]
\end{lemma}
\begin{proofof}{Lemma~\ref{lem:expectation}}
First we define the notion of symmetry: For a vector $x\in \mathbb{R}^{d^B}$, where $B\in\mathbb{N}^*$, if for all permutation $\sigma$ of $[d]$, and for all $i_1,\cdots, i_B\in[d]$, $x_{i_1i_2\cdots i_B}=x_{\sigma(i_1)\sigma(i_2)\cdots \sigma(i_B)}$, then we say vector $x$ is symmetric. 

Besides, for a vector $v\in\mathbb{R}^{d^l}$,  we use $v_{i_1,i_2,\cdots,i_l}$ to refer to $\text{Tensor}(v)_{i_1,i_2,\cdots,i_l}$ where $\text{Tensor}$ is the inverse translation of $\vvv$, i.e., $\text{Tensor}$ translates a $\mathbb{R}^{d^l}$ vector back to a $\mathbb{R}^{d^{\otimes l}}$ tensor. In other words, we use $v_{i_1,i_2,\cdots,i_l}$ to refer to the entry $v_{(i_1-1)d^{l-1}+(i_2-1)d^{l-2}+\cdots+(i_{l-1}-1)d+i_l}$. Similarly, for a $d^l\times d^l$ matrix $M$, we use $M_{i_1,i_2,\cdots,i_l,j_1,j_2,\cdots,j_l}$ to denote $\text{Tensor}(M)_{i_1,i_2,\cdots,i_l,j_1,j_2,\cdots,j_l}$, or in other words, \\$M_{(i_1-1)d^{l-1}+(i_2-1)d^{l-2}+\cdots+(i_{l-1}-1)d+i_l,(j_1-1)d^{l-1}+(j_2-1)d^{l-2}+\cdots+(j_{l-1}-1)d+j_l}$.

Assume that $v=u^{\otimes l}$, then $v\in\mathbb{R}^{d^{\otimes l}}$, and $v$ is a symmetric tensor. Note that \[\forall i_1,\cdots, i_l\in[d], v_{i_1,\cdots, i_l}=u_{i_1}\cdots u_{i_l}.\]
Define $M\triangleq \vvv(v)\vvv(v)^\top = \vvv(u^{\otimes l})\vvv(u^{\otimes l})^\top$, then
\[M_{i_1,\cdots, i_l, j_1,\cdots, j_l}=u_{i_1}\cdots u_{i_l}\cdot u_{j_1}\cdots u_{j_l}.\]
By Wick's theorem, we know that
\[\mathbb{E}[M_{i_1,\cdots, i_l, j_1,\cdots, j_l}]=\sum_{\sigma\in P}\prod_{t\in[l]}\mathbb{E}[u_{\sigma(2t-1)}u_{\sigma(2t)}],\]
where $P$ is the set that containing all distinct partitions of $S=\{i_1,\cdots, i_l, j_1, \cdots, j_l\}$ into $l$ pairs. Each two variables in $S$ are considered different even if the values of them are the same, e.g., $\{(i_1, i_2), (j_1, j_2)\}$ and $\{(i_1, j_2), (j_1, i_2)\}$ are different partitions even if $i_2=j_2$. In other words, the partition is independent of the value of those variables. Thus, we can decompose matrix $M$ into the sum of $(2l-1)!!$ matrices, i.e.,
\[M=\sum_{\sigma\in P}M_{\sigma}.\]
Assume $\sigma_1$ is the partition $\{(i_1,j_1),\cdots,(i_l, j_l)\}$, so
\[\mathbb{E}[M_{\sigma_1}] = \prod_{t\in[l]}\mathbb{E}[u_{i_t}u_{j_t}].\]
Since $\mathbb{E}[u_{i_t}u_{j_t}]=\mathbb{I}\{i_t=j_t\}$($\mathbb{I}$ is the indicator function), we know that all elements on the diagonal of $\mathbb{E}[M_{\sigma_1}]$ are 1, and all other elements are 0, which means that $\mathbb{E}[M_{\sigma_1}]$ is the identity matrix. Hence, \[b^\top \mathbb{E}[M_{\sigma_1}] b=1.\]

Note that $b$ is a symmetric vector, meaning $b^\top \mathbb{E}[M_{\sigma_1}] b$ doesn't change if we permute $\{i_1,\cdots,i_l\}$. Thus, as long as each $i$ is paired with a $j$ in $\sigma$, we will have $b^\top \mathbb{E}[M_{\sigma_1}] b =  b^\top \mathbb{E}[M_{\sigma}] b$. There are $n!$ such partitions, so summing them up gives us $l!.$.

For any other partition $\sigma$, we can always permute $\{i_1,\cdots,i_l\}$ and $\{j_1,\cdots,j_l\}$ such that the partition becomes $\{(i_1,i_2),\cdots, (i_{2t-1},i_{2t}),(j_1,j_2),\cdots, (j_{2t-1},j_{2t}),(i_{2t+1},j_{2t+1}),\cdots,(i_l, j_l)\}$. Then
\[\mathbb{E}[M_\sigma]=I_{d^{l-2t}}\otimes (ww^\top),\]
where $w\in\mathbb{R}^{d^{2t}}$ and $w_{i_1,\cdots,i_{2t}}=\mathbb{I}\{i_1=i_2\}\cdots\mathbb{I}\{i_{2t-1}=i_{2t}\}$. Therefore, $\mathbb{E}[M_\sigma]$ is a positive semi-definite matrix, i.e., $b^\top \mathbb{E}[M_{\sigma}]b\geq 0$.

In a word, we can divide $\mathbb{E}[M]$ into the sum of two sets of matrices. In the symmetric sense, the first set of matrices are equivalent to identity matrices while the second set of matrices are equivalent to some semi-definite matrices. Therefore,
\[b^\top\mathbb{E}[\vvv(u^{\otimes l})\vvv(u^{\otimes l})^\top]b\geq l!.\]
\end{proofof}

\begin{proofof}{Lemma~\ref{lem:expectation_unit}}
Let $u\in\R^d$ be a standard normal vector, i.e., $u\sim\N(0,I_d)$, then from Theorem 2 in \cite{vignat2008extension} we know that
\[b^\top \E\br{\vvv\pr{\pr{u/\n{u}}^{\otimes l}}\vvv\pr{\pr{u/\n{u}}^{\otimes l}}^\top}b=\frac{\Gamma\pr{\frac{d}{2}}}{2^l\Gamma\pr{l+\frac{d}{2}}}b^\top \E[\vvv(u^{\otimes l})\vvv(u^{\otimes l})^\top]b.\]
Furthermore, from Lemma~\ref{lem:expectation} we know that
\[^\top\mathbb{E}[\vvv(u^{\otimes l})\vvv(u^{\otimes l})^\top]b\geq l!.\]
Note that $u/\n{u}$ is distributed as a uniform vector from the unit sphere $\mathbb{S}^{d-1}.$ Combining the above equality and inequality, we finish the proof of this lemma.
\end{proofof}

%% file: proof_local_min.tex
\section{Construction of Bad Local Minimum}\label{sec:proof_local_min}
In this section, we construct a bad local min for the vanilla loss function with vanilla parameterization of $T$. That is, $T := \sum_{i=1}^m c_i u_i^{\otimes l}$ and $f_v(U,C)=1/2\fns{T-T^*}$. Recall Theorem~\ref{thm:bad_local_min} as follows.

\badLocalMin*

In our example, the model fits one direction in $T^*$ but misses all the other directions. Moving any component towards one of the missing directions would actually make the approximation worse because of the cross terms. The proof of Theorem~\ref{thm:bad_local_min} is in Section~\ref{sec:proofs_technical_local_min}.

We also extend the local min to the vanilla loss function with 2-homogeneous parameterization of $T$. That is, $T := \sum_{i=1}^m a_i c_i^{l-2} u_i^{\otimes l}$ and $f(U,C)=1/2\fns{T-T^*}$. For simplicity, we assume half of the $a_i$'s are $1$'s. 

\begin{restatable}
{theorem}{badLocalMin2homo}\label{thm:bad_local_min_2_homo}
Let $f(U,C):=1/2\fns{\sum_{i=1}^m a_i c_i^{l-2} u_i^{\otimes l}-T^*}$. Assume $\floor{m/2}$ of $a_i$'s are $1$'s and the remaining are $-1$'s.
Assume $l\geq 3, d-2 \geq r\geq 1$ and $m\geq 4r(l+1)+2.$ There exists a symmetric ground truth tensor $T^*$ with rank at most $2r(l+1)+2$ such that a local minimum with function value $l(l-1)r/2$ exists while the global minimum has function value zero.
\end{restatable}

In the above Theorem, we treat $c_i$'s as separate variables from $u_i$'s. That is, at a local min $(U,C),$ we allow arbitrary perturbations to $c_i$'s regardless of the perturbations to $u_i$'s and show none of these perturbations can decrease the function value. Note our result trivially extends to the case when $c_i=1/\n{u_i}$ since the coupling between $c_i$ and $u_i$ only restricts the set of possible perturbations to $(U,C).$ The detailed proof of Theorem~\ref{thm:bad_local_min_2_homo} is in Section~\ref{sec:proofs_technical_local_min}.

\subsection{Detailed Proofs}\label{sec:proofs_technical_local_min}
\begin{proofof}{Theorem~\ref{thm:bad_local_min}}
In this proof, we first construct a ground truth tensor and a local min with non-zero loss. To further prove this local min is indeed spurious, we show there exists a global min with zero loss under the same ground truth tensor.

We first define the local min. For every $i\in[m],$ let $c_i$ be 1 and $u_i$ be $e_1/m^{\frac1l}$. Then, we know at this point $T=e_1^{\otimes l}.$

We define the ground truth tensor $T^*$ by defining the residual $R:=T-T^*.$ The residual $R$ is defined as the summation of $\hat{R}$ and all its permutation. We define $\hat{R}$ as follows,
\[\hat{R}:=\sum_{j=2}^{r+1} e_j^{\otimes 2}\otimes e_1^{\otimes l-2}.\]
Then, $R$ is defined as the summation of all $\binom{l}{2}$ permutations of $\hat{R}$. It's clear that $R$ is symmetric and therefore $T^*$ is symmetric.

Let $U$ be a $d\times m$ matrix whose $i$-th row is $u_i$, and $C$ be an $m\times m$ diagonal matrix with $C_{ii}=c_i,\forall i\in[m]$. Suppose we perform a local change to $U$ and $C$ such that $U'=U+\Delta U, C'=C+\Delta C$ and $\fn{\Delta U},\fn{\Delta C}\rightarrow 0.$ We prove that for any $\Delta U, \Delta C$, we have $f(U',C')\geq f(U,C).$ Let's first show that the gradient at $U$ is zero, which means there is no locally first-order change on the function value.

\paragraph{First-order Change}
Let's first show the gradient of $f_v$ w.r.t. $U$ and $C$ at $(U,C)$ is zero. Here we first compute the gradient in terms of one column $u_i$,
\begin{align*}
&\forall i\in[m],\nabla_{u_i} f_v(U,C)=lR(u_i^{\otimes l-1},I)c_i=\frac{l}{m^{\frac{l-1}{l}}}R(e_1^{\otimes l-1},I).\\
&\forall i\in[m],\nabla_{c_i} f_v(U,C)=R(u_i^{\otimes l})=\frac{1}{m}R(e_1^{\otimes l}).
\end{align*}
In order to compute $R(e_1^{\otimes l-1},I)$, we first consider $\hat{R}(e_1^{\otimes l-1},I).$ We have
\[\hat{R}(e_1^{\otimes l-1},I)=\sum_{j=2}^{r+1} e_j\inner{e_j}{e_1}\inner{e_1}{e_1}^{l-2}=0.\]
Similarly,
\[\hat{R}(e_1^{\otimes l})=\sum_{j=2}^{r+1} \inner{e_j}{e_1}^2\inner{e_1}{e_1}^{l-2}=0.\]
The computation for other permutations of $\hat{R}$ is the same. Overall, we have $\nabla f_v(U,C)=0$.

\paragraph{Second-order change}
The second order change of $f_v(U,C)$ is as follows,
\begin{align*}
&\frac{1}{2}\fns{\sum_{i=1}^{m}\pr{c_i(\Delta u_i)\otimes u_i^{\otimes l-1}+c_iu_i\otimes (\Delta u_i)\otimes u_i^{\otimes l-2}+\cdots +c_iu_i^{\otimes l-1}\otimes (\Delta u_i)} +(\Delta c_i)u_i^{\otimes l}}\\
&+\sum_{i=1}^{m} \pr{l(l-1)R(u_i^{\otimes l-2},\Delta u_i,\Delta u_i)c_i+lR(u_i^{\otimes l-1}, \Delta u_i)\Delta c_i}.
\end{align*}
The first term is always non-negative, and the second term can be further computed as follows:
\begin{align*}
l(l-1)\sum_{i=1}^m R(u_i^{l-2},\Delta u_i,\Delta u_i)=& l(l-1)\frac{1}{m^{(l-2)/l}}\sum_{i=1}^m R(e_1^{\otimes l-2},\Delta u_i,\Delta u_i)\\
=& l(l-1)\frac{1}{m^{(l-2)/l}}\sum_{i=1}^m \sum_{j=2}^{r+1} [\Delta u_i]_j^2.
\end{align*}
Similar to the computations in the first-order change part, we know $R(u_i^{\otimes l-1}, \Delta u_i)=0$. Therefore, the second-order change of $f(U,C)$ can be lower bounded by $l(l-1)\frac{1}{m^{(l-2)/l}}\sum_{i=1}^m \sum_{j=2}^{r+1} [\Delta u_i]_j^2$.

For any $\Delta U,\Delta C,$ if there exists $i\in[m],2\leq j\leq r+1$ such that $[\Delta u_i]_j\neq 0$, we know the second order change is positive. Combining with the fact that the gradient is zero at $(U,C)$, this implies the function value increases.

Otherwise, if $[\Delta u_i]_j= 0$ for all $i\in[m]$ and all $2\leq j\leq r+1,$ we know $\Delta u_i\in B$ for all $i\in [m]$ where $B$ is the span of $\{e_1,e_k|r+2\leq k\leq d\}.$ Let the perturbed tensor be $T',$ we know $T'-T$ lies in the $B^{\otimes l}$ subspace. Note perturbing $c_i$ introduces changes in $e_1^{\otimes l}$ direction that is also in the $B^{\otimes l}$ subspace. This type of perturbation cannot decrease the function value because the residual $R$ is orthogonal with $B^{\otimes l}$ subspace.

Overall, we have proved that $(U,C)$ is a local minimizer. Notice that residual $R$ contains $r\binom{l}{2}$ orthogonal components with unit norm. Therefore, the function value at $(U,C)$ is $f(U,C)=\frac{1}{2}\fns{R}=\frac{1}{2}\times r\times \binom{l}{2}=\frac{l(l-1)r}{4}.$

\paragraph{Construction of global minimizer:} Next we will show that when $m\geq r(l+1)+1$, there exists $U$ and $C$ such that $f(U,C)=0$. Therefore, the local minimizer we found above must be a spurious local minimizer. We only need to show that $T^*$ can be expressed as the summation of $r(l+1)+1$ rank-one symmetric tensors.

Define $\hat{R}_j:=e_j^{\otimes 2}\otimes e_1^{\otimes l-2}$, and define $R_j$ to be the sum of all $\binom{l}{2}$ permutations of $\hat{R_j}$. Then we can write $T^*$ as
\[T^*=e_1^{\otimes l}+\sum_{j=2}^{r+1} R_j.\]
Note that $R_j$ is a symmetric tensor with entries equal to 1 if the index of the entry has 2 $j$'s and $(l-2)$ 1's, and entries equal to 0 otherwise.
Define $v_{i,j}:=e_1+b_{i,j}e_j$ where $b_{i,j}\in\mathbb{R}$, and consider the tensor $\bar{T}_j:=\sum_{i=1}^{l+1}\bar{b}_{i,j}v_{i,j}^{\otimes l}$. Then we know that $\bar{T}_j$ is also a symmetric tensor with entries equal to $\sum_{i=1}^{l+1}\bar{b}_{i,j}b_{i,j}^k$ if the index of the entry has $k$ $j$'s and $(l-k)$ 1's, and entries equal to 0 otherwise. Therefore, if $\forall k\in\{0,1,\cdots,l\}\backslash\{2\}$, $\sum_{i=1}^{l+1}\bar{b}_{i,j}b_{i,j}^k=0$ and $\sum_{i=1}^{l+1}\bar{b}_{i,j}b_{i,j}^2=1$, then $\bar{T}_j=R_j$. In other words, we want
\[
\left(
\begin{array}{c}
0\\
0\\
1\\
0\\
\vdots\\
0\\
\end{array}
\right) =
\left(
  \begin{array}{cccc}
    1 & 1 & \cdots & 1\\
    b_{1,j} & b_{2,j} & \cdots & b_{l+1,j}  \\
    b_{1,j}^2 & b_{2,j}^2 & \cdots & b_{l+1,j}^2  \\
    \vdots  & \vdots  & \vdots & \vdots \\
    b_{1,j}^l & b_{2,j}^l & \cdots & b_{l+1,j}^l  \\
  \end{array}
\right)
\left(
\begin{array}{c}
\bar{b}_{1,j}\\
\bar{b}_{2,j}\\
\vdots\\
\bar{b}_{l+1,j}
\end{array}
\right).
\]
Denote the matrix in the middle by $M_j$, then
\[|M_j|=\prod_{\substack{1\leq s< t\leq l+1}}(b_{s,j}-b_{t,j}).\]
Thus, as long as the $b_{i,j}$'s are mutually different, the matrix $M_j$ will be full rank, and there must exist a set of $\bar{b}_{i,j}$'s such that the equation above holds. In other words, we have shown that there exists such $b_{i,j}$'s and $\bar{b}_{i,j}$'s that $\forall 2\leq j\leq r+1$, $\bar{T}_j=R_j$. Therefore, we know each $R_j$ can be expressed as the summation of $(l+1)$ rank-one symmetric tensors.

To summarize, when $m\geq r(l+1)+1$, we can construct $T^*$ such that there exists a local minimum with function value $\frac{l(l-1)r}{4}$ while the global minimum has function value zero.

\end{proofof}

\begin{proofof}{Theorem~\ref{thm:bad_local_min_2_homo}}
The proof is very similar as the proof of Theorem~\ref{thm:bad_local_min}. The only difference is that in the $2$-homogeneous, all the $c_i$'s are positive and we need to rely on positive and negative $a_i$'s to fit the ground truth tensors. We need to define a slightly different ground truth tensor and bad local min.

Same as in the proof of Theorem~\ref{thm:bad_local_min}, we first construct a ground truth tensor and a local min with non-zero loss. To further prove this local min is indeed spurious, we show there exists a global min with zero loss under the same ground truth tensor.

We first define the local min. Let $m'=\floor{m/2}.$ Without loss of generality, assume $a_i=1$ for all $i\in[m']$ and $a_i=-1$ for all $i\in[m'+1,m].$ For any $i\in[m'],$ let $u_i=\sqrt{1/m'}e_1$ and $c_i=1/\n{u_i}.$ For any $i\in[m'+1,m],$ let $u_i=\sqrt{1/(m-m')}e_d$ and $c_i=1/\n{u_i}.$ With this choice of parameters, it's not hard to verify that $T=e_1^{\otimes l}-e_d^{\otimes l}.$

We define the ground truth tensor $T^*$ by defining the residual $R:=T-T^*.$ The residual $R$ is defined as the summation of $\hat{R}$ and all its permutation, where $\hat{R}$ is defined as:
\[\hat{R}:=\sum_{j=2}^{r+1} \pr{e_j^{\otimes 2}\otimes e_1^{\otimes l-2}-e_j^{\otimes 2}\otimes e_d^{\otimes l-2}}.\]
Since we assume $r\leq d-2,$ we know $r+1\leq d-1$ and $e_j$ is orthogonal with $e_1,e_d$ for all $2\leq j\leq r+1.$
Then, $R$ is defined as the summation of all $\binom{l}{2}$ permutations of $\hat{R}$. It's clear that $R$ is symmetric and $T^*$ is also symmetric.

Suppose we perform a local change to $U$ and $C$ such that $U'=U+\Delta U, C'=C+\Delta C$ and $\fn{\Delta U},\fn{\Delta C}\rightarrow 0,$ where $\Delta C$ is a diagonal matrix. We prove that for all possible $\Delta U, \Delta C$, $f(U',C')\geq f(U,C).$

\paragraph{First-order Change}
Let's first show the derivative of $f$ in terms of all $u_i$'s and $c_i$'s at $(U,C)$ is zero. This means there is no first order decrease direction at $(U,C).$

For any $i\in[m'],$ we can compute the derivative in terms of $u_i$ and $c_i$:
\begin{align*}
\nabla_{u_i} f(U,C)=&l  R(u_i^{\otimes l-1},I)c_i^{l-2}=\frac{l}{\sqrt{m'}}R(e_1^{\otimes l-1},I),\\
\nabla_{c_i} f(U,C)=&(l-2) R(u_i^{\otimes l})c_i^{l-3}=\frac{l-2}{(m')^{3/2}}R(e_1^{\otimes l}).
\end{align*}
It's not hard to verify that $R(e_1^{\otimes l-1},I)=0$ and $R(e_1^{\otimes l})=0$ using the orthogonality between $e_1$ and $e_j$ for all $2\leq j\leq r+1.$ For the same reason, we also have $\nabla_{u_i} f(U,C)=0, \nabla_{c_i} f(U,C)=0$ for all $i\in[m'+1,2m].$

\paragraph{Second-order change}
The second order change of $f(U',C')$ compared with $f(U,C)$ is as follows,
\begin{align*}
&\frac{1}{2}\fns{\sum_{i=1}^{m}a_i\pr{ c_i^{l-2}\pr{(\Delta u_i)\otimes u_i^{\otimes l-1}+u_i\otimes (\Delta u_i)\otimes u_i^{\otimes l-2}+\cdots +u_i^{\otimes l-1}\otimes (\Delta u_i)} +(l-2) c_i^{l-3}(\Delta c_i)u_i^{\otimes l}}}\\
&+\sum_{i=1}^{m}a_i \pr{l(l-1)R(u_i^{\otimes l-2},\Delta u_i,\Delta u_i)c_i^{l-2}+l(l-2)R(u_i^{\otimes l-1}, \Delta u_i)c_i^{l-3}\Delta c_i + (l-2)(l-3)R(u_i^{\otimes l})c_i^{l-4}(\Delta c_i)^2}.
\end{align*}
(When $l=3$, we do not have the $c_i^{l-4}$ term)

The first term is always non-negative. By the previous argument, we also know $R(u_i^{\otimes l-1}, \Delta u_i)=0$ and $R(u_i^{\otimes l})=0$. Therefore, the second order change is lower bounded by $\sum_{i=1}^{m}a_i l(l-1)R(u_i^{\otimes l-2},\Delta u_i,\Delta u_i)c_i^{l-2}.$ Let's first consider the components from $1$ to $m'$ for which $a_i=1,$
\begin{align*}
l(l-1)a_i\sum_{i=1}^{m'} R(u_i^{\otimes l-2},\Delta u_i,\Delta u_i)c_i^{l-2}=& l(l-1)\sum_{i=1}^{m'} R(e_1^{\otimes l-2},\Delta u_i,\Delta u_i)\\
=& l(l-1)\sum_{i=1}^{m'} \sum_{j=2}^{r+1} [\Delta u_i]_j^2.
\end{align*}
For the components from $m'+1$ to $m$, we have
\begin{align*}
l(l-1)a_i\sum_{i=m'+1}^{m} R(u_i^{\otimes l-2},\Delta u_i,\Delta u_i)c_i^{l-2}=& l(l-1)\sum_{i=m'+1}^{m} -R(e_d^{\otimes l-2},\Delta u_i,\Delta u_i)\\
=& l(l-1)\sum_{i=m'+1}^{m} \sum_{j=2}^{r+1} [\Delta u_i]_j^2.
\end{align*}
Overall, we have
\[
\sum_{i=1}^{m}a_i l(l-1)R(u_i^{\otimes l-2},\Delta u_i,\Delta u_i)c_i^{l-2}\geq l(l-1)\sum_{i=1}^{m} \sum_{j=2}^{r+1} [\Delta u_i]_j^2.
\]

For any $\Delta U,\Delta C,$ if there exists $i\in[m],2\leq j\leq r+1$ such that $[\Delta u_i]_j\neq 0$, we know the second order change is positive. Combining with the fact that the gradient is zero at $(U,C)$, this implies the function value increases.

Otherwise, if $[\Delta u_i]_j= 0$ for all $i\in[m]$ and all $2\leq j\leq r+1,$ we know $\Delta u_i\in B$ for all $i\in [m]$ where $B$ is the span of $\{e_1,e_k|r+2\leq k\leq d\}.$ Let the perturbed tensor be $T',$ we know $T'-T$ lies in the $B^{\otimes l}$ subspace. Note perturbing $c_i$ introduces changes in $e_1^{\otimes l}$ direction or $e_d^{\otimes l}$ direction that are also in the $B^{\otimes l}$ subspace. This type of perturbation cannot decrease the function value because the residual $R$ is orthogonal with $B^{\otimes l}$ subspace.

Overall, we have proved that $(U,C)$ is a local minimizer. Notice that residual $R$ contains $2r\binom{l}{2}$ orthogonal components with unit norm. Therefore, the function value at $(U,C)$ is $f(U,C)=\frac{1}{2}\fns{R}=\frac{1}{2}\times 2r\times \binom{l}{2}=\frac{l(l-1)r}{2}.$

\paragraph{Construction of global minimizer:} Next, we show as long as $m\geq 4r(l+1)+2$, there exists parameters $(U,C)$ such that $f(U,C)=0.$ To prove this, we first write $T^*$ as summation of rank-one symmetric tensors.

For any $2\leq j\leq r+1$, define $\hat{R}_{j,1}:=e_j^{\otimes 2}\otimes e_1^{\otimes l-2} \text{ and } \hat{R}_{j,d}=e_j^{\otimes 2}\otimes e_d^{\otimes l-2}$, and define $R_{j,1},R_{j,d}$ to be the sum of all $\binom{l}{2}$ permutations of $\hat{R}_{j,1}$ and $\hat{R}_{j,d}$ respectively. Then we can write $T^*$ as
\[T^*=e_1^{\otimes l}-e_d^{\otimes l}-\sum_{j=2}^{r+1} R_{j,1}+\sum_{j=2}^{r+1} R_{j,d}.\]

Same as in the proof of Theorem~\ref{thm:bad_local_min}, we can show each $R_{j,1}$ or $R_{j,d}$ can be written as the sum of $(l+1)$ rank-one symmetric tensors.

Therefore, we know the ground truth $T^*$ can be expressed as the summation of $2+2r(l+1)$ rank-one symmetric tensors. For each component, we can re-scale it to make it fit the form of $a_ic_i^{l-2}u_i^{\otimes l},$ with $a_i=\pm 1$ and $c_i=1/\n{u_i}.$ In these rank-one tensors, at most $2r(l+1)+1$ has positive (negative) $a_i$. So, as long as $m'\geq 2r(l+1)+1$ or $m\geq 4r(l+1)+2$ our model is able to fit this ground truth tensor and achieve zero loss.

To summarize, when $m\geq 4r(l+1)+2$, we can construct $T^*$ with rank at most $2r(l+1)+2$ such that there exists a local minimum with function value $\frac{l(l-1)r}{2}$ while the global minimum has function value zero.
\end{proofof}

%% file: proof_algo.tex
\section{Detailed Proofs of Theorem~\ref{thm:main}}
\label{sec:full_proof}

In this section, we give the proof of Theorem~\ref{thm:main}. We first state a formal version of Theorem~\ref{thm:main}.

\begin{theorem}\label{thm:main_formal}
Given any target accuracy $\epsilon>0,$ there exists $m=O\pr{\frac{r^{2.5l}}{\epsilon^5}\log(d/\epsilon)}$, $\lambda=O\pr{\frac{\epsilon}{r^{0.5l}}}$, $\delta=O\pr{\frac{\epsilon^{5l-1.5}}{d^{l-1.5}(\log(d/\epsilon))^{l+0.5}r^{2.5l^2-0.75l}}}$, $\eta=O\pr{\frac{\epsilon^{15l-4.5}}{d^{3l-4.5}(\log(d/\epsilon))^{3l+1.5}r^{7.5l^2-2.25l}}}$ and $H=O\pr{\frac{d^{3l-4.5}(\log(d/\epsilon))^{3l+2.5}r^{7.5l^2-1.75l}}{\epsilon^{15l-3.5}}}$ such that with probability at least $0.99,$ our algorithm finds a tensor $T$ satisfying
\[\fn{T-T^*}\leq \epsilon,\]
within $K=O\pr{\frac{r^{2l}}{\epsilon^4}\log(d/\epsilon)}$ epochs.
\end{theorem}

We follow the proof strategy outlined in Section~\ref{short_proof_sketch}.

As we discussed in Challenge 2, bad local minima exist for our loss function. Therefore, gradient descent might get stuck at a bad local minima. This issue is fixed in our algorithm by re-initializing one component at the beginning of each epoch.
In Lemma~\ref{lem:norm_explode}, we show as long as the objective is large, there is at least a constant probability to improve the objective within one epoch. We state the formal version of Lemma~\ref{lem:norm_explode} as follows. The proof of Lemma~\ref{lem:norm_explode_formal} is in Section~\ref{sec:escape_local_min}.

\begin{lemma}\label{lem:norm_explode_formal}
Let $(U_0',\bar{C}_0')$ and $(U_H,\bar{C}_H)$ be the parameters at the beginning of an epoch and the parameters at the end of the same epoch. For the target accuracy $\epsilon>0$ in Theorem~\ref{thm:main_formal}, assume $K\leq\frac{\lambda m}{14}$ and $\fn{T_0'-T^*}\geq \epsilon$ where $T_0'$ is the tensor with parameters $(U_0',\bar{C}_0')$. There exists $m=O\pr{\frac{r^{2.5l}}{\epsilon^5}\log(d/\epsilon)}$, $\lambda=O\pr{\frac{\epsilon}{r^{0.5l}}}$, $\delta=O\pr{\frac{\epsilon^{5l-1.5}}{d^{l-1.5}(\log(d/\epsilon))^{l+0.5}r^{2.5l^2-0.75l}}}$, $\eta=O\pr{\frac{\epsilon^{15l-4.5}}{d^{3l-4.5}(\log(d/\epsilon))^{3l+1.5}r^{7.5l^2-2.25l}}}$, $H=O\pr{\frac{d^{3l-4.5}(\log(d/\epsilon))^{3l+2.5}r^{7.5l^2-1.75l}}{\epsilon^{15l-3.5}}}$, such that with probability at least $\frac16$, we have
\[f(U_H, \bar{C}_H)-f(U_0',\bar{C}_0')\leq -\Omega\pr{\frac{\epsilon^4}{r^{2l}\log(d/\epsilon)}}.\]
\end{lemma}

We compliment this lemma by showing that even if an epoch does not improve the objective, it will not increase the function value by too much. The formal version of Lemma~\ref{co:f_increase_bounded} is as follows. We prove Lemma~\ref{co:f_increase_bounded_formal} in Section~\ref{sec:f_increase_bounded}.

\begin{lemma}\label{co:f_increase_bounded_formal}
Assume $K\leq\frac{\lambda m}{14}$, $\delta\leq \frac{\mu_1\epsilon}{m^{\frac34}d^{\frac{l-2}{2}}(m+K)^{\frac{l-1}{2}}\lambda^{\frac12}}$, and $\eta\leq \frac{\mu_2\lambda}{m^{\frac12}  d^{\frac{l-1}{2}}(m+K)^{\frac{l-2}{2}}}$ for some constants $\mu_1,\mu_2$, and $\frac{10}{m}\leq\lambda\leq 1$. Let $(U_0',\bar{C}_0')$ and $(U_H,\bar{C}_H)$ be the parameters at the beginning of an epoch and the parameters at the end of the same epoch. Assume $f(U_0',\bar{C}_0')\geq \epsilon^2$, where $\epsilon$ is the target accuracy in Theorem~\ref{thm:main_formal}. Then we have
\[f(U_H, \bar{C}_H)-f(U_0',\bar{C}_0')\leq O\pr{\frac{1}{\lambda m}}.\]
\end{lemma}

From these two lemmas, we know that in each epoch, the loss function can decrease by $\Omega\pr{\frac{\epsilon^4}{r^{2l}\log(d/\epsilon)}}$ with probability at least $\frac16$, and even if we fail to decrease the function value, the increase of function value is at most $O\pr{\frac{1}{\lambda m}}$. Therefore, choosing a large enough $m$, the function value decrease will dominate the increase. This allows us to prove Theorem~\ref{thm:main_formal}.

\begin{proofof}{Theorem~\ref{thm:main_formal}}
We use a contradiction proof to show that with high probability our algorithm finds a tensor $T$ satisfying $\fn{T-T^*}\leq \epsilon$ within $K$ epochs. For the sake of contradiction, we assume $\fn{T-T^*}>\epsilon$ through the first $K$ epochs. Under this assumption, we show with high probability the function value will decrease below zero.

Note that under the choice of parameters of this theorem, all the conditions of Lemma~\ref{lem:norm_explode_formal} and Lemma~\ref{co:f_increase_bounded_formal} are satisfied. By Lemma~\ref{lem:norm_explode_formal}, we know that with probability at least $1/6$, the function value decreases by at least $\Lambda:=\Omega\pr{\frac{\epsilon^4}{r^{2l}\log(d/\epsilon)}}$ in each epoch. By Lemma~\ref{co:f_increase_bounded_formal}, we show that the function value at most increases by $\Lambda' := O(\frac{1}{\lambda m})$ in each epoch. Using our choice of the parameters in Theorem~\ref{thm:main_formal}, we know that $O(\frac{1}{\lambda m})=O\pr{\frac{\epsilon^4}{r^{2l}\log(d/\epsilon)}}$. Choosing a large enough constant factor for $m$ ensures that $\Lambda'\leq \frac{\Lambda}{10}$.

For each $1\leq k\leq K$, let $\calE_k$ be the event that in the beginning of the $k$-th epoch, the reinitialized component $a c_0^{l-2} u_0^l$ has good correlation with the residual (see Lemma~\ref{lem:initial_correlate_formal}) and $\n{P_S u_0}\geq \frac{\mu\delta}{\sqrt{d}},$ where $\mu$ is some constant. We know $\calE_k$'s are independent with each other and $\Pr[\calE_k]\geq 1/6.$ By Hoeffding's inequality, we know as long as $K\geq \mu'$ for certain constant $\mu',$ we have $\sum_{k=1}^K \indic{\calE_k}\geq K/7$ with probability at least $0.99$, where $\indic{\calE_k}$ is the indicator function of event $\calE_k.$

By the proof of Lemma~\ref{lem:norm_explode_formal}, we know conditioning on $\calE_k,$ the function value decreases by at least $\Lambda$ in the $k$-th epoch. Since $\sum_{k=1}^K \indic{\calE_k}\geq K/7$, we know the total function value decrease is at least $K\Lambda/7-K\Lambda/10=K\Omega\pr{\frac{\epsilon^4}{r^{2l}\log(d/\epsilon)}}.$ Therefore, there exists $K=O\pr{\frac{r^{2l}\log(d/\epsilon)}{\epsilon^4}}$ such that $K\Lambda/7-K\Lambda/10\geq 4.$

By the analysis in Lemma~\ref{lem:U_bounded_formal}, the function value is upper bounded by 3 at initialization. However, with probability at least 0.99, the decrease of the function value is at least $4$, meaning that the function value must be negative, which is a contradiction. Therefore, we know that with probability at least 0.99, our algorithm finds a tensor $T$ satisfying $\fn{T-T^*}\leq\epsilon$ within $K=O\pr{\frac{r^{2l}\log(d/\epsilon)}{\epsilon^4}}$ epochs.
\end{proofof}

\subsection{Upper bound on function increase}\label{sec:f_increase_bounded}

In this section, we prove Lemma~\ref{co:f_increase_bounded_formal}.

To prove the increase of $f$ is bounded in one epoch, we identify all the possible ways that the loss can increase and upper bound each of them. We first show that a normal step (without re-initialization or scalar mode switch) of the algorithm will not increase the objective function. Note that many parts of our proofs rely on an upperbound on function value. To get such a bound the proof includes an induction component: when we prove Lemma~\ref{lem:f_value_decrease_formal} and Lemma~\ref{lem:f_increase_bounded_formal}, we assume that the function value is upper bounded by a constant, and we will inductively prove that these conditions are satisfied in Lemma~\ref{lem:U_bounded_formal}. This induction ensures that the conclusions of all the lemmas in this section hold throughout the entire algorithm.

The following lemma is a formal version of Lemma~\ref{lem:f_value_decrease} in the main text.

\begin{lemma}\label{lem:f_value_decrease_formal}
Let $(U,\bar{C})$ be the parameters at the beginning of one iteration and let $U',\bar{C}'$ be the updated parameters (before potential scalar mode switch). Assuming $f(U,\bar{C})\leq 10,$ $\lambda\leq 1$, there exists constants $\mu_1,\mu_2$ such that
\[f(U',\bar{C}')-f(U,\bar{C})\leq -\frac{\eta}{l}\fns{\nabla_U f(U,\bar{C})}\]
as long as
$\delta\leq \frac{\mu_1}{m^{\frac14}\sqrt{\lambda}d^{\frac{l-2}{2}}(m+K)^{\frac{l-1}{2}}},\eta\leq \frac{\mu_2\lambda}{m^{\frac12} d^{\frac{l-1}{2}}(m+K)^{\frac{l-2}{2}}}.$
\end{lemma}

Recall that in an iteration, we first update $U$ by gradient descent, then update $C$ and $\hat{C}$ by the updated value of $U$. The gradient descent step on $U$ cannot increase the function value as long as the step size is small enough. The update on $C$ and $\hat{C}$ can potentially increase the function value. In the proof of Lemma~\ref{lem:f_value_decrease_formal}, we show the increase due to updating $C$ and $\hat{C}$ is proportional to the decrease by updating $U$ and smaller in scale.

\begin{proofof}{Lemma~\ref{lem:f_value_decrease_formal}}

According to the algorithm, each iteration contains two steps: update $U$ as $U'\leftarrow U-\eta\nabla_U f(U,\bar{C})$; update $c_i$ and $\hat{c}_i$ as $c_i'=c_i\frac{\n{u_i}}{\n{u_i'}}$ and $\hat{c}_i'=\hat{c}_i\frac{\n{u_i}}{\n{u_i'}}$. We can divide the function value change into these two steps: $f(U',\bar{C}')-f(U,\bar{C})=(f(U',\bar{C})-f(U,\bar{C}))+(f(U',\bar{C}')-f(U',\bar{C})).$ We will show that the function value decrease in the first step and does not increase by too much in the second step. At the end, we will combine them to show that overall the function value decreases.

Since we assume $f(U,\bar{C})\leq 10.$ According to the definition of the loss function, we know $\fn{T-T^*}\leq \sqrt{20},\sum_{i=1}^m\ns{u_i}\leq \frac{10}{\lambda}.$ We also know that $\sum_{i=1}^m\n{u_i}^4\leq \pr{\sum_{i=1}^m\ns{u_i}}^2\leq\frac{100}{\lambda^2}$. For convenience, denote $\Gamma=10,M_4^2 := \frac{100}{\lambda^2}$ and $M_2^2:=\frac{10}{\lambda}.$

\paragraph{$f(U',\bar{C})-f(U,\bar{C})$ is negative:} In the first step, we update $U$ by gradient descent, which should decrease the function value as long as we choose the step size to be small enough. To prove that an inverse polynomially step size suffices, we need to bound the second derivative of $f$ in terms of $U$ at $(U'',\bar{C})$ for any $U''\in\{(1-\theta) U +\theta U'| 0\leq \theta\leq 1\}.$ Let $\hessian''$ be the Hessian of $f$ in terms of $U$ at $(U'',\bar{C})$. We will bound the Frobenius norm of $\hessian''.$

Let's first show that $\n{u_i''}\leq (1+1/(4l))\n{u_i}$ when $\eta$ is small enough. Recall the derivative in $u_i$ is,
\begin{align*}
\nabla_{u_i} f(U,\bar{C})=l(T-T^*)(u_i^{\otimes(l-1)},I)c_i^{l-2}a_i+\lambda lu_i.
\end{align*}
Therefore, we can bound the derivative as
\[\n{\nabla_{u_i} f(U,\bar{C})}\leq \pr{l\sqrt{2\Gamma}(\sqrt{d(m+K)})^{l-2}+\lambda l}\n{u_i}.
\]
Thus, as long as $\eta\leq \frac{1}{4l^2(\sqrt{2\Gamma}(\sqrt{d(m+K)})^{l-2}+\lambda)}$, we have
\[\eta\n{\nabla_{u_i} f(U,\bar{C})}\leq \eta\pr{l\sqrt{2\Gamma}(\sqrt{d(m+K)})^{l-2}+\lambda l}\n{u_i}\leq   \frac{1}{4l}\n{u_i}.\]
Since $u_i'' = u_i-\theta\eta\nabla_{u_i} f(U,\bar{C})$ for $0\leq \theta\leq 1,$ we know that
\begin{align*}
\n{u''_i}\leq& \n{u_i}+\eta\n{\nabla_{u_i} f(U,\bar{C})}\\
\leq& (1+\frac{1}{4l})\n{u_i},
\end{align*}

Let $T''$ be the tensor parameterized by $(U'',\bar{C})$. We can bound $\fn{T''-T^*}$ as follows,
\begin{align*}
\fn{T''-T^*}\leq& \fn{T-T^*}+\sum_{i=1}^m\sum_{k=1}^l\binom{l}{k}\n{u_i}^{l-k}\n{\eta \nabla_{u_i} f(U,\bar{C})}^k c_i^{l-2}\\
\leq& \fn{T-T^*}+\sum_{i=1}^m\sum_{k=1}^l\binom{l}{k}\frac{1}{l^k}\n{u_i}^{l} c_i^{l-2}\\
\leq& \fn{T-T^*}+l\sum_{i=1}^m\pr{4(\sqrt{d(m+K)})^{l-2}(m+K)\delta^2+\|u_i\|^2 }\\
\leq& \fn{T-T^*}+4lm(\sqrt{d(m+K)})^{l-2}(m+K)\delta^2+lM_2^2\\
\leq& \sqrt{2\Gamma}+2l M_2^2,
\end{align*}
where the last inequality assumes $\delta \leq\frac{M_2}{\sqrt{4m(\sqrt{d(m+K)})^{l-2}(m+K)}}.$ For convenience, denote $\beta:=\sqrt{2\Gamma}+2l M_2^2.$

With the bound on $\n{T''-T^*}$ and $\n{u_i''},$ we are ready to bound the Frobenius norm of $\hessian''.$
For each $i\in[m],$ we have
\begin{equation}
\frac{\partial}{\partial u_i} f(U'',\bar{C})=l(T''-T^*)((u_i'')^{l-1},I)c_i^{l-2}a_i+\lambda l\pr{\frac{\n{u_i''}}{\n{u_i}}}^{l-2}u_i''.\label{eqn:df-u}
\end{equation}
We know $\hessian''$ is a $dm\times dm$ matrix that contains $m\times m$ block matrices with dimension $d\times d.$ Each block corresponds to the second-order derivative of $f(U'',\bar{C})$ in terms of $u_i,u_j.$ We will bound the Frobenius norm of $\hessian''$ by bounding the Frobenius norm of each block.

For each $i,$ we can compute $\frac{\partial^2 }{\partial u_i\partial u_i}f(U'',\bar{C})$ as follows,
\begin{align*}
\frac{\partial^2 }{\partial u_i\partial u_i}f(U'',\bar{C})
=&l(l-1)(T''-T^*)((u_i'')^{l-2},I,I)c_i^{l-2}a_i+l^2c_i^{2l-4}\n{u_i''}^{2l-4}u_i''\otimes u_i''\\
&+\lambda l\pr{\frac{\n{u_i''}}{\n{u_i}}}^{l-2}I+\lambda l(l-2)\frac{\n{u_i''}^{l-4}u_i''\otimes u_i''}{\n{u_i}^{l-2}}.
\end{align*}

For the first term, we have $l(l-1)\fn{(T''-T^*)((u_i'')^{l-2},I,I)c_i^{l-2}}\leq l^2\sqrt{e}\beta(\sqrt{d(m+K)})^{l-2}$ since $\n{u_i''}/\n{u_i}\leq (1+1/(4l))$.

For the second term, we have
\begin{align*}
l^2\fn{c_i^{2l-4}\n{u_i''}^{2l-4}u_i''\otimes u_i''}\leq& l^2\sqrt{e}\max\{4(d(m+K))^{l-2}(m+K)\delta^2,\|u_i\|^2\}.
\end{align*}

For the third term, we have
\[\fn{\lambda l\pr{\frac{\n{u_i''}}{\n{u_i}}}^{l-2}\vvv(I)}\leq \lambda l\sqrt{e}\sqrt{d}.\]

For the fourth term, we have
\[\fn{\lambda l(l-2)\frac{\n{u_i''}^{l-4}u_i''\otimes u_i''}{\n{u_i}^{l-2}}}\leq \lambda l^2\sqrt{e}.\]

Combing the bounds on these terms and assuming $\lambda\leq 1$, we have
\begin{align*}
\fn{\frac{\partial^2 }{\partial u_i\partial u_i}f(U'',\bar{C})}\leq& l^2\sqrt{e}\beta(\sqrt{d(m+K)})^{l-2}+l^2\sqrt{e}\max\{4d^{l-2}(m+K)^{l-1}\delta^2,\ns{u_i}\}+2l^2\sqrt{e}\sqrt{d}.
\end{align*}
Thus,
\begin{align*}
\fns{\frac{\partial^2 }{\partial u_i\partial u_i}f(U'',\bar{C})}\leq& 3e l^4\beta^2 d^{l-2}(m+K)^{l-2}+ 3e l^4\max\{16d^{2l-4}(m+K)^{2l-2}\delta^4,\|u_i\|^4\}+12el^4d\\
\leq& 15e l^4\beta^2 d^{l-2}(m+K)^{l-2}+ 3e l^4\max\{16d^{2l-4}(m+K)^{2l-2}\delta^4,\|u_i\|^4\}.
\end{align*}

For each pair of $i\neq j,$ we can compute $\frac{\partial^2 }{\partial u_i\partial u_j}f(U'',\bar{C})$ as follows
\begin{align*}
\frac{\partial^2 }{\partial u_i\partial u_j}f(U'',C)=l^2a_ia_jc_i^{l-2}c_j^{l-2}\inner{u_i''}{u_j''}^{l-2}u_i''\otimes u_j''.
\end{align*}
The Frobenius norm square can be bounded as
\begin{align*}
\fns{\frac{\partial^2 }{\partial u_i\partial u_j}f(U'',\bar{C})} \leq& e l^4 \max\{4d^{l-2}(m+K)^{l-1}\delta^2,\|u_i\|^2\}\max\{4d^{l-2}(m+K)^{l-1}\delta^2,\|u_j\|^2\}\\
\leq& el^4\max\{\max\{\ns{u_i},\ns{u_j}\}^2, 16d^{2l-4}(m+K)^{2l-2}\delta^4 \}\\
\leq& e l^4 \pr{\n{u_i}^4+\n{u_j}^4+16d^{2l-4}(m+K)^{2l-2}\delta^4 }
\end{align*}

Summing over the bounds on blocks, we can bound the Frobeneius norm of $\hessian'',$
\begin{align*}
\fns{\hessian''}=&\sum_{i,j}\fns{\frac{\partial^2 }{\partial u_i\partial u_j}f(U'',\bar{C})}\\
\leq& 15e ml^4\beta^2 d^{l-2}(m+K)^{l-2}+ 48em l^4d^{2l-4}(m+K)^{2l-2}\delta^4+3el^4\sum_{i=1}^m \n{u_i}^4\\
&+(m-1)el^4\sum_{i=1}^m\n{u_i}^4+16el^4m(m-1)d^{2l-4}(m+K)^{2l-2}\delta^4\\
=& 15e ml^4\beta^2 d^{l-2}(m+K)^{l-2}+16el^4m(m+2)d^{2l-4}(m+K)^{2l-2}\delta^4+(m+2)el^4\sum_{i=1}^m\n{u_i}^4\\
\leq& 15e ml^4\beta^2 d^{l-2}(m+K)^{l-2}+2(m+2)el^4M_4^2
\end{align*}
where the last inequality assumes $\delta\leq \pr{\frac{M_4^2}{16 m d^{2l-4}(m+K)^{2l-2}}}^{1/4}.$

Denoting $L_1:=\sqrt{15e ml^4\beta^2 d^{l-2}(m+K)^{l-2}+2(m+2)el^4M_4^2}$, we have
\begin{align*}
f(U',\bar{C})-f(U,\bar{C})\leq -\eta\fns{\nabla_U f(U,\bar{C})}+\frac{\eta^2 L_1}{2}\fns{\nabla_U f(U,\bar{C})}
\end{align*}

\paragraph{$f(U',\bar{C}')-f(U',\bar{C})$ is bounded:} Next, we show that setting $c_i'$ as $c_i\frac{\n{u_i}}{\n{u_i'}}$ and $\hat{c}_i'$ as $\hat{c}_i\frac{\n{u_i}}{\n{u_i'}}$ does not increase the function value by too much. We use $\nabla_{\hu_i} f$ to denote the gradient of $u_i$ through $c_i$ and $\hat{c}_i$, which means
\[\nabla_{\hu_i} f = \frac{\partial f}{\partial c_i}\frac{\partial c_i}{\partial u_i}+\frac{\partial f}{\partial \hat{c}_i}\frac{\partial \hat{c}_i}{\partial u_i}.\]
In the following we first bound the Frobenius norm of the Hessian of $f$ in terms of $\hat{U}$ evaluated at $(U',\bar{C}'')$ for any $C''\in \{\mbox{diag}(c_1'',\cdots,c_m'')| c_i''=c_i\frac{\n{u_i}}{\n{(1-\theta)u_i+\theta u_i' }},0\leq \theta\leq 1 \}$ and $\hat{C}''\in \{\diag(\hat{c}_1'',\cdots,\hat{c}_m'')| \hat{c}_i''=\hat{c}_i\frac{\n{u_i}}{\n{(1-\theta)u_i+\theta u_i' }},0\leq \theta\leq 1 \}$. We denote the Hessian at $(U',\bar{C}'')$ as $\hat{\hessian}''$, which is a $md\times md$ matrix. Hessian $\hat{\hessian}''$ contains $m\times m$ blocks with dimension $d\times d,$ each of which corresponds to $\frac{\partial^2}{\partial \hu_i\partial \hu_j}f(U',\bar{C}'')$ for some $(i,j)\in[m]\times[m]$.

Note that $\frac{\n{u_i'}}{\n{u_i''}}\leq 1+1/l$ since $\n{u'_i-u_i}\leq 1/(4l)\n{u_i}.$ Let $T'''$ be the tensor corresponds to $(U',\bar{C}''),$ we can bound $\fn{T'''-T^*}$ as follows,
\begin{align*}
\fn{T'''-T^*}\leq& \fn{\sum_{i=1}^m a_i(c_i'')^{l-2}(u_i')^{\otimes l}}+\fn{T^*}\\
\leq& e\pr{\sum_{i=1}^m \ns{u_i} + 4(\sqrt{d(m+K)})^{l-2}m(m+K)\delta^2}+1\\
\leq& 2eM_2^2+1,
\end{align*}
where the last step assumes $\delta \leq \frac{M_2}{\sqrt{4(\sqrt{d(m+K)})^{l-2}m(m+K) }}.$ For convenience, denote $\alpha:=2eM_2^2+1.$

Let's first compute the derivative of $f$ in terms of $\hu_i,$
\begin{align}
&\frac{\partial }{\partial \hu_i}f(U',\bar{C}'')\\
=&-(l-2)a_i(T'''-T^*)((u_i')^{\otimes l})\frac{u_i''}{\n{u_i''}^l}\pr{(\sqrt{d(m+K)})^{l-2}\indicc+\indiccb}\notag\\
&-\lambda(l-2)\frac{u_i''}{\n{u_i''}^{l}}\n{u_i'}^{l}.\label{eqn:df-hu}
\end{align}
For each $i,$ we have
\begin{align*}
&\frac{\partial^2 }{\partial \hu_i\partial \hu_i}f(U',\bar{C}'')\\
=&-(l-2)a_i(T'''-T^*)((u_i')^{\otimes l})\frac{I}{\n{u_i''}^l}\pr{(\sqrt{d(m+K)})^{l-2}\indicc+\indiccb}\\
&+l(l-2)a_i(T'''-T^*)((u_i')^{\otimes l})\frac{u_i''(u_i'')^\top}{\n{u_i''}^{l+2}}\pr{(\sqrt{d(m+K)})^{l-2}\indicc+\indiccb}\\
&+(l-2)^2\n{u_i'}^{2l}\frac{u_i''(u_i'')^\top}{\n{u_i''}^{2l}}\pr{d^{l-2}(m+K)^{l-2}\indicc+\indiccb}\\
&-\lambda(l-2)\frac{I}{\n{u_i''}^{l}}\n{u_i'}^{l}\\
&+\lambda l(l-2)\frac{u_i''(u_i'')^\top}{\n{u_i''}^{l+2}}\n{u_i'}^{l}.
\end{align*}
We bound its Frobenius norm square by
\begin{align*}
&\fns{ \frac{\partial^2 }{\partial \hu_i\partial \hu_i}f(U',\bar{C}'')}\\
\leq& 5\pr{l^2\alpha^2e^2d^{l-1}(m+K)^{l-2}+l^4\alpha^2e^2d^{l-2}(m+K)^{l-2}}\\
&+5\pr{l^4e^4\max\{\n{u_i}^4,16(m+K)^{2l-2}\delta^4 d^{2l-4}\}+\lambda^2l^2de^2+\lambda^2l^4e^2}\\
\leq& 20e^2\alpha^2 l^4d^{l-1}(m+K)^{l-2}+5e^4l^4\pr{\n{u_i}^4+16(m+K)^{2l-2}\delta^4 d^{2l-4}},
\end{align*}
where we assume that $\lambda\leq 1$.

For $i\neq j,$ we have
\begin{align*}
&\frac{\partial^2 }{\partial \hu_i\partial \hu_j}f(U',\bar{C}'')\\
=&(l-2)^2(\inner{u_i'}{u_j'}^{l})\frac{u_i''(u_j'')^\top}{\n{u_i''}^{l}\n{u_j''}^l}\\
&\cdot \pr{(\sqrt{d(m+K)})^{l-2}\indicc+\indiccb}\\
&\cdot \pr{(\sqrt{d(m+K)})^{l-2}\indiccj+\indiccbj}\\
\end{align*}
We can bound its Frobenius norm by
\begin{align*}
\fns{\frac{\partial^2 }{\partial \hu_i\partial \hu_j}f(U',\bar{C}'')}
\leq& l^4 e^4\max\{4d^{l-2}(m+K)^{l-1}\delta^2,\|u_i\|^2\}\max\{4d^{l-2}(m+K)^{l-1}\delta^2,\|u_j\|^2\}\\
\leq& l^4 e^4\pr{\n{u_i}^4+\n{u_j}^4+16(m+K)^{2l-2}\delta^4 d^{2l-4}}
\end{align*}

Combing the bounds on all blocks, we have
\begin{align*}
\fns{\hat{\hessian}''}\leq& \sum_{i=1}^m \pr{20e^2\alpha^2 l^4d^{l-1}(m+K)^{l-2}+5e^4l^4 \pr{\n{u_i}^4+16(m+K)^{2l-2}\delta^4 d^{2l-4}}}\\
                         &+ \sum_{\substack{i,j\in[m]\\i\neq j}}  l^4 e^4\pr{\n{u_i}^4+\n{u_j}^4+16(m+K)^{2l-2}\delta^4 d^{2l-4}} \\
                      \leq& 20me^2\alpha^2l^4d^{l-1}(m+K)^{l-2}+80ml^4e^4(m+K)^{2l-2}\delta^4d^{2l-4}+5l^4e^4\sum_{i=1}^m\n{u_i}^4\\
                      &+   16m^2l^4e^4(m+K)^{2l-2}\delta^4d^{2l-4}+2l^4e^4m\sum_{i=1}^m\n{u_i}^4\\
                      \leq& 20me^2\alpha^2l^4d^{l-1}(m+K)^{l-2}+7l^4 e^4m M_4^2+96m^2l^4e^4(m+K)^{2l-2}\delta^4 d^{2l-4}\\
\leq& 20me^2\alpha^2l^4d^{l-1}(m+K)^{l-2}+8 l^4e^4mM_4^2,
\end{align*}
where the last inequality assumes $\delta\leq \pr{\frac{M_4^2}{96m(m+K)^{2l-2}d^{2l-4}}}^{1/4}.$

Denote $L_2:=\sqrt{20me^2\alpha^2l^4d^{l-1}(m+K)^{l-2}+8 l^4e^4mM_4^2}.$
Then, we have
\begin{align*}
f(U',\bar{C}')-f(U',\bar{C})
\leq& \inner{\nabla_{\hat{U}}f(U',\bar{C})}{-\eta\nabla_U f(U,\bar{C})}+\frac{\eta^2 L_2}{2}\fns{\nabla_U f(U,\bar{C})}.
\end{align*}

From equation \ref{eqn:df-u} and \ref{eqn:df-hu} we know that $\forall i\in[m]$, $\nabla_{\hat{u}_i}f(U,\bar{C})=-\frac{l-2}{l}\cdot\frac{u_iu_i^\top(\nabla_{u_i}f(U,\bar{C}))}{\ns{u_i}}$. Thus,
\[\fn{\nabla_{\hat{U}}f(U,\bar{C})}\leq \frac{l-2}{l}\fn{\nabla_{U}f(U,\bar{C})}.\]
In order to bound $\fn{\nabla_{\hat{U}}f(U',\bar{C})},$ we still need to show that $\nabla_{\hat{U}}f(U',\bar{C})$ is close to $\nabla_{\hat{U}}f(U,\bar{C}).$
\paragraph{Bounding $\fn{\nabla_{\hat{U}}f(U',\bar{C})-\nabla_{\hat{U}}f(U,\bar{C})}: $} Define $U''$ as $(1-\theta)U+\theta U'$ for all $0\leq \theta\leq 1.$ We will show that the derivative of $\nabla_{\hat{U}}f$ in $U$ evaluated $(U'',\bar{C})$ is bounded. We denote this derivative as $\tilde{\hessian}''$ that is a $dm\times dm$ matrix. Matrix $\tilde{\hessian}''$ contains $m\times m$ blocks each of which has dimension $d\times d$ and corresponds to $\frac{\partial^2}{\partial \hu_i \partial u_j}f(U'',\bar{C}).$ Denote $T''$ as the tensor parameterized by $(U'',\bar{C}).$ Recall that,
\begin{align*}
&\frac{\partial }{\partial \hu_i}f(U'',\bar{C})\\
=&-(l-2)a_i(T''-T^*)((u_i'')^{\otimes l})\frac{u_i}{\n{u_i}^l}\pr{(\sqrt{d(m+K)})^{l-2}\indiccU+\indiccbU}\notag\\
&-\lambda(l-2)\frac{u_i}{\n{u_i}^{l}}\n{u_i''}^{l}.
\end{align*}
For any $i\in[m],$ we have
\begin{align*}
&\frac{\partial^2 }{\partial u_i\partial \hu_i}f(U'',\bar{C})\\
=&-l(l-2)a_i(T''-T^*)((u_i'')^{\otimes l-1},I)\frac{u_i^\top }{\n{u_i}^l}\pr{(\sqrt{d(m+K)})^{l-2}\indiccU+\indiccbU}\\
&-l(l-2)c_i^{l-2}\n{u_i''}^{2l-2}\frac{u_i(u_i'')^\top}{\n{u_i}^l}\pr{(\sqrt{d(m+K)})^{l-2}\indiccU+\indiccbU}\\
&-\lambda l(l-2)\frac{u_i(u_i'')^\top}{\n{u_i}^{l}}\n{u_i''}^{l-2}.
\end{align*}
The Frobenius norm of $\frac{\partial^2 }{\partial u_i\partial \hu_i}f(U'',\bar{C})$ can be bounded as follows,
\begin{align*}
\fn{\frac{\partial^2 }{\partial u_i\partial \hu_i}f(U'',\bar{C})}
\leq& \sqrt{e}l^2\beta(\sqrt{d(m+K)})^{l-2}+\sqrt{e} l^2 \max\pr{\ns{u_i},4d^{l-2}(m+K)^{l-1}\delta^2}+\sqrt{e}\lambda l^2\\
\leq& 2\sqrt{e}l^2\beta(\sqrt{d(m+K)})^{l-2}+\sqrt{e} l^2 \max\pr{\ns{u_i},4d^{l-2}(m+K)^{l-1}\delta^2},
\end{align*}
where the last inequality assumes $\lambda\leq 1$.
Therefore,
\begin{align*}
\fns{\frac{\partial^2 }{\partial u_i\partial \hu_i}f(U'',\bar{C})}
\leq& 8el^4\beta^2 d^{l-2}(m+K)^{l-2}+2e l^4 \max\pr{\n{u_i}^4,16d^{2l-4}(m+K)^{2l-2}\delta^4 }
\end{align*}

For $i\neq j,$ we have
\begin{align*}
&\frac{\partial^2 }{\partial u_j\partial \hu_i}f(U'',\bar{C})\\
=&-l(l-2)a_ia_jc_j^{l-2}\inner{u_i''}{u_j''}^{l-1}\frac{u_i(u_i'')^\top}{\n{u_i}^l}\pr{(\sqrt{d(m+K)})^{l-2}\indiccU+\indiccbU}
\end{align*}
The Frobenius norm square can be bounded as
\begin{align*}
\fns{\frac{\partial^2 }{\partial \hu_i\partial u_j}f(U'',C)} \leq& e l^4 \max\{4d^{l-2}(m+K)^{l-1}\delta^2,\|u_i\|^2\}\max\{4d^{l-2}(m+K)^{l-1}\delta^2,\|u_j\|^2\}\\
\leq& el^4\pr{\n{u_i}^4+\n{u_j}^4+16(m+K)^{2l-2}\delta^4 d^{2l-4}}.
\end{align*}

Summing over the bounds on blocks, we can bound the Frobeneius norm of $\tilde{\hessian}'',$
\begin{align*}
&\fns{\tilde{\hessian}''}\\
=&\sum_{i,j}\fns{\frac{\partial^2 }{\partial u_j\partial \hu_i}f(U'',\bar{C})}\\
\leq& 8mel^4\beta^2 d^{l-2}(m+K)^{l-2}+2el^4\sum_{i=1}^m \n{u_i}^4 + 32mel^4d^{2l-4}(m+K)^{2l-2}\delta^4\\
&+2mel^4 \sum_{i=1}^m \n{u_i}^4+16m^2el^4(m+K)^{2l-2}\delta^4d^{2l-4}\\
\leq& 8mel^4\beta^2 d^{l-2}(m+K)^{l-2}+48m^2el^4d^{2l-4}(m+K)^{2l-2}\delta^4+3el^4mM_4^2\\
\leq& 8mel^4\beta^2 d^{l-2}(m+K)^{l-2}+4el^4 mM_4^2,
\end{align*}
where the last inequality assumes $\delta\leq \pr{\frac{M_4^2}{48 m d^{2l-4}(m+K)^{2l-2}}}^{1/4}.$

Denoting $L_3:=\sqrt{8mel^4\beta^2 d^{l-2}(m+K)^{l-2}+4el^4 mM_4^2}$, we have
\begin{align*}
\fn{\nabla_{\hat{U}} f(U',\bar{C})-\nabla_{\hat{U}} f(U,\bar{C})}\leq L_3\fn{\eta \nabla_U f(U,\bar{C})}\leq \frac{1}{3l}\fn{\nabla_U f(U,\bar{C})},
\end{align*}
where the second inequality assumes $\eta\leq \frac{1}{3L_3 l}.$ Therefore, we have
\begin{align*}
\fn{\nabla_{\hat{U}} f(U',\bar{C})}\leq \fn{\nabla_{\hat{U}} f(U,\bar{C})}+\frac{1}{3l}\fn{\nabla_U f(U,\bar{C})}
\leq& \pr{\frac{l-2}{l}+\frac{1}{3l}}\fn{\nabla_U f(U,\bar{C})}\\
\leq& \pr{1-\frac{5}{3l}}\fn{\nabla_U f(U,\bar{C})}.
\end{align*}

Overall, we have proved that as long as $\eta$ is small enough,
\begin{align*}
f(U',\bar{C}')-f(U,\bar{C})
=& f(U',\bar{C})-f(U,\bar{C})+f(U',\bar{C}')-f(U',\bar{C})\\
\leq& -\eta\fns{\nabla_U f(U,\bar{C})}+\frac{\eta^2L_1}{2}\fns{\nabla_U f(U,\bar{C})}\\
& +\eta\fns{\nabla_{\hat{U}} f(U',\bar{C})}+\frac{\eta^2 L_2}{2}\fns{\nabla_{\hat{U}} f(U',\bar{C})}\\
\leq& -\eta\fns{\nabla_U f(U,\bar{C})}+\frac{\eta^2L_1}{2}\fns{\nabla_U f(U,\bar{C})}\\
& +\eta\pr{1-\frac{5}{3l}}\fns{\nabla_U f(U,\bar{C})}+\frac{\eta^2 L_2}{2}\fns{\nabla_U f(U,\bar{C})}\\
\leq& -\frac{\eta}{l}\fns{\nabla_U f(U,\bar{C})},
\end{align*}
where the last inequality assumes $\eta\leq \frac{4}{3l(L_1+L_2)}.$ Combining all the bounds on $\delta, \eta,$ we know there exists constant $\mu_1,\mu_2$ such that as long as $\delta\leq \frac{\mu_1}{m^{\frac14}\sqrt{\lambda}d^{\frac{l-2}{2}}(m+K)^{\frac{l-1}{2}}},\eta\leq \frac{\mu_2\lambda}{m^{\frac12} d^{\frac{l-1}{2}}(m+K)^{\frac{l-2}{2}}},$ we have
\[f(U',\bar{C}')-f(U,\bar{C})\leq -\frac{\eta}{l}\fns{\nabla_U f(U,\bar{C})}.\]
\end{proofof}

Then, we know in an epoch, the function value can only increase because of the initialization and the scalar mode switches. In Lemma~\ref{lem:f_increase_bounded_formal}, we show these operations cannot increase the function by too much. Note that Lemma~\ref{lem:f_increase_bounded_formal} is the formal version of Lemma~\ref{lem:reinit_bounded} together with the bound for scalar mode switches in the main text (these two arguments in the main text correspond to the two claims in Lemma~\ref{lem:f_increase_bounded_formal}).

\begin{lemma}\label{lem:f_increase_bounded_formal}
Assume $f(U_0',\bar{C}_0')\leq\tilde{\Gamma}\leq 10$ at the beginning of an epoch, $\delta\leq \frac{\mu_1 \sqrt{\tilde{\Gamma}}}{m^{\frac34}d^{\frac{l-2}{2}}(m+K)^{\frac{l-1}{2}}\lambda^{\frac12}}$, and $\eta\leq \frac{\mu_2\lambda}{m^{\frac12}  d^{\frac{l-1}{2}}(m+K)^{\frac{l-2}{2}}}$ for some constants $\mu_1,\mu_2$. Also assume that $\lambda m\geq 10$. Denote the parameters at the end of this epoch as $(U_H,\bar{C}_H)$, then
\[f(U_H,\bar{C}_H)\leq\exp\pr{\frac{14}{\lambda m}}\tilde{\Gamma}.\]
\end{lemma}

\begin{proofof}{Lemma~\ref{lem:f_increase_bounded_formal}}
By Lemma~\ref{lem:f_value_decrease_formal}, we know the function value does not increase in any iteration (before potential scalar mode switch) as long as the initial function value is at most $10$ and $\delta\leq \frac{\mu_1}{m^{\frac14}\sqrt{\lambda}d^{\frac{l-2}{2}}(m+K)^{\frac{l-1}{2}}},\eta\leq \frac{\mu_2\lambda}{m^{\frac12}  d^{\frac{l-1}{2}}(m+K)^{\frac{l-2}{2}}}$ for some constants $\mu_1, \mu_2.$ Thus, the function value can only increase when we reinitialize a component or when we switch the scaling from $\sqrt{d(m+K)}/\n{u_i}$ to $1/\n{u_i}.$ In the following, we first show that reinitializing a component can only increase the function value by a small factor.

\begin{claim}
\label{claim:re-init}
Suppose $f(U,\bar{C})\leq\hat{\Gamma}\leq 10$. Reinitialize any vector with the smallest $\ell_2$ norm among all columns of $U$, and let the updated parameters be $(U',\bar{C}')$, then
\[f(U',\bar{C}')\leq \pr{1+ \frac{13}{\lambda m}}\hat{\Gamma}.\]
\end{claim}
According to the definition of the function value, we know $\fn{T-T^*}\leq \sqrt{2\hat{\Gamma}}\leq\sqrt{20}$, $\sum_{j=1}^m \ns{u_j}\leq\frac{\hat{\Gamma}}{\lambda}$, and $\sum_{j=1}^m\n{u_j}^4\leq\pr{\sum_{j=1}^m\ns{u_j}}^2\leq \frac{\hat{\Gamma}^2}{\lambda^2}$. Suppose $u_i$ is one of the vectors in $U$ with the smallest $\ell_2$ norm, then $\ns{u_i}\leq \frac{\hat{\Gamma}}{\lambda m}.$ Suppose $u_i',c_i',\hat{c}_i',a_i'$ are the corresponding reinitialized vector and coefficients, and we have
\begin{align*}
&\fn{a_ic_i^{l-2}u_i^{\otimes l} - a_i'(c_i')^{l-2}(u_i')^{\otimes l} }\\
\leq& \fn{a_ic_i^{l-2}u_i^{\otimes l}}+\fn{a_i'(c_i')^{l-2}(u_i')^{\otimes l}}\\
\leq& \max\pr{\ns{u_i},(\sqrt{d(m+K)})^{l-2}4(m+K)\delta^2}+(\sqrt{d(m+K)})^{l-2}\delta^2\\
\leq& \frac{\hat{\Gamma}}{\lambda m}+(\sqrt{d(m+K)})^{l-2}4(m+K)\delta^2+(\sqrt{d(m+K)})^{l-2}\delta^2\leq \frac{\hat{2\Gamma}}{\lambda m},
\end{align*}
where the last inequality assumes $\delta^2\leq \frac{\hat{\Gamma}}{5\lambda m(m+K)^{\frac{l}{2}}d^{\frac{l-2}{2}}}.$ Therefore, we can bound $f(U',\bar{C}')$ as
\begin{align*}
&f(U',\bar{C}')\\
=&\frac{1}{2}\fns{T-T^*-a_ic_i^{l-2}u_i^{\otimes l}+a_i'(c_i')^{l-2}(u_i')^{\otimes l}}+\lambda\sum_{j=1}^m \hat{c}_j^{l-2}\n{u_j}^{l}-\lambda \hat{c}_i^{l-2}\n{u_i}^{2l}+\lambda(\hat{c}_i')^{l-2}\n{u_i'}^{l}\\
\leq& f(U,C)+\fn{T-T^*}\fn{a_ic_i^{l-2}u_i^{\otimes l}-a_i'(c_i')^{l-2}(u_i')^{\otimes l}}+\frac{1}{2}\fns{a_ic_i^{l-2}u_i^l-a_i'(c_i')^{l-2}(u_i')^l}+\lambda(\hat{c}_i')^{l-2}\n{u_i'}^{l}\\
\leq& f(U,C)+\sqrt{20}\cdot\frac{2\hat{\Gamma}}{\lambda m}+\frac{1}{2}\pr{2\frac{\hat{\Gamma}}{\lambda m}}^2+\lambda \delta^2\\
\leq& \pr{1+ \frac{12}{\lambda m}}\hat{\Gamma}+\lambda \delta^2\\
\leq& \pr{1+ \frac{13}{\lambda m}}\hat{\Gamma},
\end{align*}
where the second last inequality assumes $\lambda m\geq \hat{\Gamma}$ and the last inequality assumes $\delta^2\leq \frac{\hat{\Gamma}}{\lambda^2m}.$

Switching the scaling from $\sqrt{d(m+K)}/\n{u_i}$ to $1/\n{u_i}$ can also potentially increase the function value. In the following, we show that the function value increase is small because we only switch the scaling mode when $\n{u_i}\leq 2\sqrt{m+K}\delta.$
\begin{claim}
\label{claim:scaling-switch}
Assume $f(U',\bar{C}')\leq \bar{\Gamma}.$ Suppose at this iteration we switch the scaling of $c_i'$, i.e., we set $c_i'$ as $c_i'/\sqrt{d(m+K)}.$ Let the updated parameters be $(U',C'',\hat{C}',A'),$ we have
\[f(U',C'',\hat{C}',A')\leq \pr{1+\frac{1}{\lambda m^2}}\bar{\Gamma}.\]
\end{claim}
Suppose $u_i$ is the parameter which is one step of gradient descent before $u_i'$. According to the algorithm, we know $\n{u_i}\leq 2\sqrt{m+K}\delta.$ According to the proof in Lemma~\ref{lem:f_value_decrease_formal} where we bound the derivative of $f$ with respect to $u_i$, we know that as long as $\eta\leq \frac{1}{l(\sqrt{2\Gamma}(\sqrt{d(m+K)})^{l-2}+\lambda)}$, we have $\n{u_i'}\leq 2\n{u_i}\leq 4\sqrt{m+K}\delta$. Therefore,
\begin{align*}
\fn{(c_i')^{l-2}(u_i')^{\otimes l}-(c_i'')^{l-2}(u_i')^{\otimes l}}=&\fn{(c_i')^{l-2}(u_i')^{\otimes l}-\frac{1}{(\sqrt{d(m+K)})^{l-2}}(c_i')^{l-2}(u_i')^{\otimes l}}\\
\leq& \fn{(c_i')^{l-2}(u_i')^{\otimes l}}\leq 16(m+K)\delta^2(\sqrt{d(m+K)})^{l-2}.
\end{align*}
Suppose the tensor at $(U',\bar{C}')$ is $T',$ then $\fn{T'-T^*}\leq \sqrt{2\bar{\Gamma}}.$ 

Thus, we can bound $f(U',C'',\hat{C}',A')$ as follows:
\begin{align*}
&f(U',C'',\hat{C}',A')\\
\leq& f(U',\bar{C}'')+\fn{T'-T^*}\fn{(c_i')^{l-2}(u_i')^{\otimes l}-(c_i'')^{l-2}(u_i')^{\otimes l}}\\
&+\frac{1}{2}\fns{(c_i')^{l-2}(u_i')^{\otimes l}-(c_i'')^{l-2}(u_i')^{\otimes l}}\\
\leq& \bar{\Gamma}+\sqrt{2\bar{\Gamma}}\pr{16(m+K)\delta^2(\sqrt{d(m+K)})^{l-2}}+\frac{1}{2}\pr{16(m+K)\delta^2(\sqrt{d(m+K)})^{l-2}}^2\\
\leq& \pr{1+\frac{1}{\lambda m^2}}\bar{\Gamma},
\end{align*}
where the last inequality assumes $\delta^2\leq \frac{\sqrt{\bar{\Gamma}}}{32\sqrt{2}\lambda m^2(m+K)^{\frac{l}{2}}d^{\frac{l-2}{2}}}$ and $\lambda m^2\geq 1$.

We are now ready to bound the increase of the function value during this epoch. According to the algorithm, each epoch contains at most $m$ scaling mode switches. Therefore, following Claim \ref{claim:scaling-switch}, all the scaling switches in one epoch can increase the upper bound of the function value by at most a factor of $\pr{1+\frac{1}{\lambda m^2}}^m\leq\exp\pr{\frac{1}{\lambda m}}.$ Combining with Claim \ref{claim:re-init} which considers the re-initialization, we know that in each epoch, the upper bound of the function value increase by at most a factor of $\exp\pr{\frac{1}{\lambda m}}\pr{1+ \frac{13}{\lambda m}}\leq\exp\pr{\frac{14}{\lambda m}}.$
\end{proofof}

Following Lemma~\ref{lem:f_value_decrease_formal} and Lemma~\ref{lem:f_increase_bounded_formal}, we are ready to show that the function value is upper bounded by a constant by an induction proof. At the beginning of our algorithm, the function value is bounded by a constant as long as the initialization radius $\delta$ is small enough. According to Lemma~\ref{lem:f_increase_bounded_formal}, the increase in each epoch is bounded by a factor of $O(\frac{1}{\lambda m})$. Therefore, as long as the total number of epochs do not exceed $O(\lambda m)$, $f$ will always be bounded by a constant. As a consequence, the Frobenius norm square of $U$ must be bounded by $O(\frac{1}{\lambda})$ due to the design of the regularizer. These results are summarized into Lemma~\ref{lem:U_bounded_formal}.

\begin{lemma}\label{lem:U_bounded_formal}
Assume $\delta\leq \frac{\mu_1}{m^{\frac34}d^{\frac{l-2}{2}}(m+K)^{\frac{l-1}{2}}\lambda^{\frac12}}$, and $\eta\leq \frac{\mu_2\lambda}{m^{\frac12} l^4 d^{\frac{l-1}{2}}(m+K)^{\frac{l-2}{2}}}$ for some constants $\mu_1,\mu_2$. Also assume $K\leq \frac{\lambda m}{14}$ and $\frac{10}{m}\leq\lambda\leq 1$. We know throughout the algorithm
\[f(U,\bar{C})\leq 10\text{ and }\sum_{i=1}^m\ns{u_i}\leq \frac{10}{\lambda}.\]
\end{lemma}
\begin{proofof}{Lemma~\ref{lem:U_bounded_formal}}
Let's first show that the function value is bounded at the initialization if we choose $\delta$ to be small enough. At initialization, we have
\begin{align*}
f(U,\bar{C})=& \frac{1}{2}\fns{T-T^*}+\lambda\sum_{i=1}^m \hat{c}_i^{l-2}\n{u_i}^{l}\\
\leq& \frac{1}{2}\pr{\sum_{i=1}^m\fn{c_i^{l-2}u_i^{\otimes l}}+\fn{T^*}}^2+\lambda\sum_{i=1}^m \ns{u_i}\\
\leq& \frac{1}{2}\pr{m(\sqrt{d(m+K)})^{l-2}\delta^2+1}^2+\lambda m\delta^2\\
\leq& m^2d^{l-2}(m+K)^{l-2}\delta^4+1+\lambda m\delta^2 \leq 3,
\end{align*}
where the last inequality assumes $\delta^4\leq\frac{1}{m^2d^{l-2}(m+K)^{l-2}}$ and $\lambda\leq 1$.

We use an inductive proof to prove that the function value at the end of the $k$-th ($k\leq K$) iteration is at most $3\exp(\frac{14k}{\lambda m})$: At the initialization, the function value is at most $3$. For every epoch, assume that our induction hypothesis is true, then at each step (a step can be a re-initialization, a gradient descent update, or a scalar mode switch), from Lemma~\ref{lem:f_increase_bounded_formal} we know that the function value is upper bounded by $3\exp(\frac{14k}{\lambda m})\leq 10$, so at this step Lemma~\ref{lem:f_value_decrease_formal} is correct, meaning that Lemma~\ref{lem:f_increase_bounded_formal} is still correct at the next step.

Therefore, throughout the algorithm, we have
\[f(U,C)\leq 3\exp\pr{\frac{14K}{\lambda m}}\leq 10,\]
where we assume $K\leq \frac{\lambda m}{14}.$ This immediately implies that $\sum_{i}\ns{u_i}\leq \frac{10}{\lambda}$ by the design of our regularizer.
\end{proofof}

Now we are ready to prove Lemma~\ref{co:f_increase_bounded_formal}.

\begin{proofof}{Lemma~\ref{co:f_increase_bounded_formal}}
From Lemma~\ref{lem:U_bounded_formal}, we know that the function value is upper bounded by 10 throughout our algorithm. Besides, from Lemma~\ref{lem:f_increase_bounded_formal} we know that the function value increase is at most $(\exp(\frac{14}{\lambda m})-1)$ times the function value at the beginning of this epoch. Choosing $\tilde{\Gamma}=f(U_0',\bar{C}_0'),$ we know the function value increase at each epoch cannot exceed $10(\exp(\frac{14}{\lambda m})-1)=O(\frac{1}{\lambda m})$, which finishes the proof of Lemma~\ref{co:f_increase_bounded_formal}.
\end{proofof}

\subsection{Escaping local minima}\label{sec:escape_local_min}
In this section, we will give a formal proof of Lemma~\ref{lem:norm_explode_formal}. We again follow the proof ideas outlined in Section~\ref{sec:escape:main}. Recall that the proof goes in the following steps:
\begin{enumerate}
    \item We first show that the projection of $U$ in the $B$ subspace must be very small, therefore the influence from incorrect subspace $B$ is small (Lemma~\ref{lem:bound_ortho_norm_formal}).
    \item We then focus on the correlation in the correct subspace $S$. First we show that the correlation can be significantly negative at re-initialization with constant probability (Lemma~\ref{lem:initial_correlate_formal}).
    \item If the correlation is always significantly negative, then the re-initialized component will grow exponentially and eventually decrease the function value (Lemma~\ref{lem:keep_correlation_exp_norm_formal}).
    \item If the correlation changes significantly, the function value must also decrease (Lemma~\ref{lem:u_change_formal} and Lemma~\ref{lem:tensor_bounded_epoch_formal}).
\end{enumerate}

First of all, we need to show that the influence coming from $B$ is small enough so that it can be ignored. The following lemma is the formal version of Lemma~\ref{lem:bound-ortho-norm}. Note that the assumption $\fn{U}\leq \sqrt{\frac{10}{\lambda}}$ has been verified in Lemma~\ref{lem:U_bounded_formal} so Lemma~\ref{lem:bound_ortho_norm_formal} holds for the entire algorithm.

\begin{lemma}\label{lem:bound_ortho_norm_formal}
Assume $\fn{U}\leq \sqrt{\frac{10}{\lambda}}$ throughout the algorithm.
Assume $\lambda\leq \sqrt{10}, \delta\leq \frac{\sqrt{10}}{2\sqrt{\lambda md^{l-2}(m+K)^{l-1}}}$ and $\eta\leq \frac{\lambda}{20l}.$
Then, we know $\fns{P_BU}\leq (m+K)\delta^2$ throughout the algorithm.
\end{lemma}
\begin{proofof}{Lemma~\ref{lem:bound_ortho_norm_formal}}
At the initialization,
\[ \fns{P_BU} \leq \fns{U} = m\delta^2. \]
At the beginning of each epoch, we re-initialize one column of $U$, which at most increases $\fns{P_BU}$ by $\delta^2$. Thus, the total increase due to the re-initialization process is at most $K\delta^2$.

Then, we only need to show that running gradient descent does not increase the norm of $P_B U.$ Suppose at the beginning of one iteration, the tensor $T$ is parameterized by $(U,\bar{C}).$ Let $U'$ be the updated parameter, which means $U'=U-\eta \nabla_U f(U,\bar{C}).$ We have,
\begin{align}
&\indent \fns{P_BU'} - \fns{P_BU} \notag\\
&= \fns{P_B(U-\eta\nabla_U f(U,\bar{C}))} - \fns{P_BU} \notag \\
&= -2\eta\inner{P_BU}{P_B\nabla_U f(U,\bar{C})} + \eta^2\fns{P_B\nabla_U f(U,\bar{C})}.\label{eqn:gradient_b}
\end{align}

We will show that the first term is negative and dominates the second term when $\eta$ is small enough, which implies that gradient descent never increases the norm of $P_B U$. We first compute the gradient as follows,
\[\nabla_U f(U,\bar{C})=l\mat(T-T^*)U^{\odot l-1}C^{l-2}A +l\lambda U.\]
where $\text{mat}(T-T^*)=UC^{l-2}A(U^{\odot l-1})^\top - U^*C^*[(U^*)^{\odot l-1}]^\top$ is a $d\times d^{l-1}$ matrix. Therefore, the projection of the gradient on $B$ subspace is
\[
P_B\nabla_U f(U,\bar{C}) = lP_B \mat(T)U^{\odot l-1}C^{l-2}A +l\lambda P_B U.
\]

Now, we show that the first term in (\ref{eqn:gradient_b}) is negative.
\begin{align*}
-2\eta\inner{P_BU}{P_B\nabla_U f(U,\bar{C})}
=& -2l\eta\inner{P_BU}{P_B \mat(T)U^{\odot l-1}C^{l-2}A +\lambda P_B U}\\
=& -2l\eta\inner{P_BU C^{l-2}A[U^{\odot l-1}]^\top }{P_B \mat(T)}-2l\eta\inner{P_B U}{\lambda P_B U}\\
=& -2l\eta\fns{P_B \mat(T)} - 2l\lambda \eta\fns{P_B U}.
\end{align*}

Next, we show the second term in (\ref{eqn:gradient_b}) is bounded. We have,
\begin{align*}
\eta^2\fns{P_B\nabla_U f(U,\bar{C})}
=& \eta^2\fns{lP_B \mat(T)U^{\odot l-1}C^{l-2}A +l\lambda P_B U}\\
\leq& 2\eta^2l^2 \pr{\fns{P_B \mat(T)U^{\odot l-1}C^{l-2}A}+\lambda^2\fns{ P_B U}}
\end{align*}

Recall that $M_2=\sqrt{\frac{10}{\lambda}}$. Note that
\begin{align*}
\fns{P_B\text{mat}(T)U^{\odot l-1}C^{l-2}A} &\leq \fns{P_B\text{mat}(T)}\fns{U^{\odot l-1}C^{l-2}}\\
                                            &= \fns{P_B\text{mat}(T)}\sum_{i=1}^mc_i^{2l-4}\n{u_i}^{2l-2}\\
                                            &\leq \fns{P_B\text{mat}(T)}\sum_{i=1}^m\max\{4(d(m+K))^{l-2}(m+K)\delta^2,\ns{u_i}\}\\
                                            &\leq M_2^2\fns{P_B\text{mat}(T)}\\
\end{align*}
where the second inequality holds since $c_i=\frac{\sqrt{d(m+K)}}{\n{u_i}}$ only when $\n{u_i}\leq 2\sqrt{m+K}\delta$, and otherwise $c_i=\frac{1}{\n{u_i}}$. The last inequality assumes $\delta\leq \frac{M_2}{2\sqrt{md^{l-2}(m+K)^{l-1}}}.$

Overall, we have
\begin{align*}
& \fns{P_BU'} - \fns{P_BU} \\
=& -2\eta\inner{P_BU}{P_B\nabla_U f(U,\bar{C})} + \eta^2\fns{P_B\nabla_U f(U,\bar{C})}\\
\leq& -2l\eta\pr{\fns{P_B \mat(T)} + \lambda \fns{P_B U}} + 2\eta^2 l^2\pr{M_2^2\fns{P_B \mat(T)} + \lambda^2 \fns{P_B U }}\\
\leq& -l\eta\pr{\fns{P_B \mat(T)} + \lambda \fns{P_B U}},
\end{align*}
where the last inequality assumes $\lambda\leq M_2^2$ and $\eta\leq \frac{1}{2 lM_2^2}.$
\end{proofof}

Lemma~\ref{lem:bound_ortho_norm_formal} shows that the norm of $P_BU$ only increases at the (re-)initializations, so it will stay small throughout this algorithm. This allows us to bound the influence to our algorithm from the orthogonal subspace $B$ and only focus on subspace $S$. We denote the re-initialized vector at $t$-th step as $u_t$, and its sign as $a\in \{\pm 1\}$. Our analysis focuses on the correlation between $P_S u_t$ and the residual tensor
\[\langle P_{S^{\otimes l}}T_t-T^*, a \overline{ P_S u_t}^{\otimes l} \rangle.\]
Here $\overline{ P_S u_t}$ is the normalized version $P_S u_t$. We will show that if this correlation is significantly negative at every iteration the norm of $u_t$ will blow up exponentially.

Towards this goal, first we will show that the initial point $P_S u_0$ has a large negative correlation with the residual. We will lower bound this correlation by anti-concentration of Gaussian polynomials, and the following lemma is the formal version of Lemma~\ref{lem:initial_correlate}. Note in our notation, we have $\langle P_{S^{\otimes l}}T_t-T^*, a \overline{ P_S u_t}^{\otimes l} \rangle=a\pr{P_{S^{\otimes l}}T_t-T^*}\pr{\overline{ P_S u_t}^{\otimes l}}.$

\begin{lemma}\label{lem:initial_correlate_formal}
Suppose the residual at the beginning of one epoch is $T_0'-T^*.$ Suppose $a c_0^{l-2} u_0^{\otimes l}$ is the reinitialized component. There exists absolute constant $\mu$ such that with probability at least $1/5$,
\[
\inner{P_{S^{\otimes l}}T_0'-T^*}{a\overline{ P_S u_0}^{\otimes l}} \leq -\frac{1}{(\mu rl)^{l/2}}\fn{P_{S^{\otimes l}}T_0'-P_{S^{\otimes l}}T^*},
\]
where $\ol{P_S u_0}=P_S u_0/\n{P_S u_0}.$
\end{lemma}

\begin{proofof}{Lemma~\ref{lem:initial_correlate_formal}}
Let's restrict into the $r^l$-dimensional space $S^{\otimes l}$, and let $P_{S^{\otimes l}}$ be the projection operator that projects a $d^l$-dimensional tensor to the $r^l$-dimensional space $S^{\otimes l}$. Then, we can think of $(P_{S^{\otimes l}}T-P_{S^{\otimes l}}T^*)$ as an $r^l$ dimensional vector, and $\overline{ P_S u}$ comes from uniform distribution on $\mathbb{S}^{r-1}.$ Let $v$ be an $r$-dimensional standard normal vector, then $a(P_{S^{\otimes l}}T-P_{S^{\otimes l}}T^*)(\overline{ P_S u}^{\otimes l})$ has the same distribution as $a(P_{S^{\otimes l}}T-P_{S^{\otimes l}}T^*)(v^l)\frac{1}{\n{v}^l}.$

Let's first show that the variance of $a(P_{S^{\otimes l}}T-P_{S^{\otimes l}}T^*)(v^l)$ is large:
\begin{align*}
\var\br{a(P_{S^{\otimes l}}T-P_{S^{\otimes l}}T^*)(v^l)}
=& \E\absr{a(P_{S^{\otimes l}}T-P_{S^{\otimes l}}T^*)(v^l)}^2\\
\geq& l!\fns{P_{S^{\otimes l}}T-P_{S^{\otimes l}}T^*},
\end{align*}
where the equality holds because $\E \left[a(P_{S^{\otimes l}}T-P_{S^{\otimes l}}T^*)(v^l)\right]=0$ and the inequality follows from Lemma~\ref{lem:expectation}. It's not hard to see that $a(P_{S^{\otimes l}}T-P_{S^{\otimes l}}T^*)(v^l)$ is an $l$-th order polynomial of standard Gaussian vectors.
By anti-concentration inequality of Gaussian polynomials (Lemma~\ref{lem:anti-concentration}), we know there exists constant $\kappa$ such that
\[\Pr\br{\absr{a(P_{S^{\otimes l}}T-P_{S^{\otimes l}}T^*)(v^l)}\leq \epsilon\sqrt{l!}\fn{P_{S^{\otimes l}}T-P_{S^{\otimes l}}T^*} } \leq \kappa l\epsilon^{1/l}.\]
Choosing $\epsilon=\frac{1}{2^l \kappa^l l^l},$ we know with probability at least half,
\begin{align*}
\absr{a(P_{S^{\otimes l}}T-P_{S^{\otimes l}}T^*)(v^l)}
\geq& \frac{1}{2^l \kappa^l l^l}\sqrt{l!}\fn{P_{S^{\otimes l}}T-P_{S^{\otimes l}}T^*}\\
\geq& \frac{1}{2^l \kappa^l l^l}\pr{\frac{l}{e}}^{l/2}\fn{P_{S^{\otimes l}}T-P_{S^{\otimes l}}T^*}\\
=& \frac{1}{2^le^{l/2} \kappa^l l^{l/2}}\fn{P_{S^{\otimes l}}T-P_{S^{\otimes l}}T^*}.
\end{align*}
Since the distribution of $a(P_{S^{\otimes l}}T-P_{S^{\otimes l}}T^*)(v^l)$ is symmetric, we know with probability at least $1/4,$
\[a(P_{S^{\otimes l}}T-P_{S^{\otimes l}}T^*)(v^l)\leq -\frac{1}{2^le^{l/2} \kappa^l l^{l/2}}\fn{P_{S^{\otimes l}}T-P_{S^{\otimes l}}T^*}.\]

According to Lemma~\ref{lem:norm_vector}, we know that with probability at least $19/20,$
\[\n{v}\leq \kappa'\sqrt{r},\]
where $\kappa'$ is some constant. This further implies that with probability at least $1/5,$
\[a(P_{S^{\otimes l}}T-P_{S^{\otimes l}}T^*)(v^l)\frac{1}{\n{v}^l}\leq -\frac{1}{2^le^{l/2} (\kappa\kappa')^l (rl)^{l/2}}\fn{P_{S^{\otimes l}}T-P_{S^{\otimes l}}T^*}.\]
Choosing $\mu=4e\kappa^2(\kappa')^2$ finishes the proof.
\end{proofof}

Our next step argues that if this negative correlation is large in every step, then the norm of $u_t$ blows up exponentially. Intuitively, this is due to the fact that the correlation is basically the dominating term in the gradient, so when it is significantly negative the vector $u_t$ behaves similar to a vector doing matrix power method (here it is important that our model is 2-homogeneous so the behavior of power method is similar to the matrix setting). Below is the formal version of Lemma~\ref{lem:keep_correlation_exp_norm}.

\begin{lemma}\label{lem:keep_correlation_exp_norm_formal}
In the setting of Theorem~\ref{thm:main_formal}, within one epoch, let $T_0$ be the tensor after the reinitilization and let $T_\tau$ be the tensor at the end of the $\tau$-th iteration. Assume $\n{P_S u_0}\geq \frac{\mu_2\delta }{\sqrt{d}}$ for some constant $\mu_2\in(0,1).$ For any $H\geq t\geq 1$, as long as $\inner{P_{S^{\otimes l}}T_\tau-T^*}{a\overline{ P_S u_\tau}^{\otimes l}}\leq \frac{-\epsilon}{5(\mu_1 rl)^{l/2}}$ for some constant $\mu_1$ for all $t-1\geq \tau\geq 0$, we have
\[\ns{P_Su_{t}}\geq\pr{1+\eta\pr{\frac{\mu_2}{2}}^{l-2}\frac{\epsilon}{10(\mu_1 rl)^{\frac{l}{2}}}}^t\ns{P_Su_0}.\]
\end{lemma}

\begin{proofof}{Lemma~\ref{lem:keep_correlation_exp_norm_formal}}
We will use inductive proof for this lemma. At the first step, we have
\begin{align*}
\ns{P_S u_1} =& \ns{P_S u_0 - \eta P_S \nabla_u f(U_0,\bar{C}_0)}\\
=& \ns{P_S u_0} -\eta\inner{P_S u_0}{P_S \nabla_u f(U_0,\bar{C}_0)}+ \eta^2 \ns{P_S \nabla_u f(U_0,\bar{C}_0)}\\
\geq& \ns{P_S u_0} -\eta\inner{P_S u_0}{P_S \nabla_u f(U_0,\bar{C}_0)}.
\end{align*}

We can write down the $P_S \nabla_u f(U_0,\bar{C}_0)$ as follows,
\begin{align*}
P_S \nabla_u f(U_0,\bar{C}_0) = al(T_0-T^*)(u_0^{\otimes (l-1)},P_S)c_0^{l-2} + \lambda lP_S u_0.
\end{align*}

Let's first consider $al(T_0-T^*)((u_0)^{\otimes (l-1)},P_S)c_0^{l-2}$. We can decompose $u_0$ into $P_Su_0$ and $P_Bu_0$, so we can divide $al(T_0-T^*)(u_0^{\otimes (l-1)},P_S)c_0^{l-2}$ into $2^{l-1}$ terms, each of which corresponds to the projection of $u_0^{\otimes (l-1)}$ on a subspace in $\{S,B\}^{\otimes l-1}$. For subspace $S^{\otimes (l-1)},$ the projection is $al(P_{S^{\otimes l}}T_0-P_{S^{\otimes l}} T^*)((P_S u_0)^{\otimes l-1},P_S)c_0^{l-2}.$ Its inner product with $-P_S u_0$ is
\begin{align*}
&\inner{-P_S u_0 }{al(P_{S^{\otimes l}}T_0-P_{S^{\otimes l}} T^*)((P_S u_0)^{\otimes l-1},P_S)c_0^{l-2}}\\
=& -al(P_{S^{\otimes l}}T_0-P_{S^{\otimes l}} T^*)((P_S u_0)^{\otimes l})c_0^{l-2}\\
=& -al(P_{S^{\otimes l}}T_0-P_{S^{\otimes l}} T^*)((\overline{P_S u_0})^{\otimes l})\pr{\n{P_S u_0}c_0}^{l-2}\ns{P_S u_0}.
\end{align*}
Now, we only need to show that $\n{P_S u_0}c_0$ is lower bounded. We have
\[\n{P_S u_0}c_0=\frac{\sqrt{d(m+K)}\n{P_S u_0}}{\n{u_0}}\geq \frac{\sqrt{d(m+K)}\mu_2\delta/\sqrt{d}}{\delta}\geq \frac{\mu_2}{2},\]
where the first inequality uses $\n{P_S u_0}\geq \mu_2\delta/\sqrt{d}.$ Therefore,
\begin{align*}
\inner{-P_S u_0 }{al(P_{S^{\otimes l}}T_0-P_{S^{\otimes l}} T^*)((P_S u_0)^{\otimes l-1},P_S)c_0^{l-2}}\geq \pr{\frac{\mu_2}{2}}^{l-2}\frac{\epsilon}{5(\mu_1 rl)^{l/2}}\ns{P_S u_0}.
\end{align*}

We then bound the remaining terms in $al(T_0-T^*)((u_0)^{\otimes l-1},P_S)c_0^{l-2}$: For any $l-1\geq k\geq 1$, we consider the subspace $B^{\otimes k}\otimes S^{\otimes l-1-k}$ and all of its permutations, we bound the norm of $al(T_0-T^*)((P_B u_0)^{\otimes k},(P_S u_0)^{\otimes l-1-k},P_S)c_0^{l-2}$ as follows.
\begin{align*}
&\n{al(T_0-T^*)\pr{(P_B u_0)^{\otimes k},(P_S u_0)^{\otimes l-1-k},P_S}c_0^{l-2}}\\
=& l\n{T_0((P_B u_0)^{\otimes k},(P_S u_0)^{\otimes l-1-k},P_S)c_0^{l-2}}\\
\leq& l\sum_{i=1}^m c_{0,i}^{l-2}\n{P_B u_{0,i}}^{k}\n{P_S u_{0,i}}^{l-k}\n{P_B u_0}^{k}\n{P_S u_0}^{l-1-k}c_0^{l-2}\\
\leq& l\sum_{i=1}^m c_{0,i}^{l-2}\n{P_B u_{0,i}}\n{ u_{0,i}}^{l-1}\n{P_B u_0}\n{ u_0}^{l-2}c_0^{l-2}\\
\leq& l\sum_{i=1}^m d^{l-2}(m+K)^{l-1}\delta^2\n{u_{0,i}}\\
\leq& l\sqrt{m}d^{l-2}(m+K)^{l-1}\delta^2 M_2,
\end{align*}
where $M_2=\sqrt{\frac{10}{\lambda}}$ is the upper bound of $\fn{U}$.
Denote $R_0$ as the summation of terms in all subspaces except for $S^{\otimes l-1}.$ We have
$\n{R_0}\leq (2^{l-1}-1)l\sqrt{m}d^{l-2}(m+K)^{l-1}\delta^2 M_2.$ Therefore, we have
\begin{align*}
|\inner{P_S u_0}{R_0}|\leq \n{P_S u_0}\n{R_0}\leq& 2^ll\sqrt{m}d^{l-2}(m+K)^{l-1}\delta^2 M_2\n{P_S u_0}\\
\leq& \pr{\frac{\mu_2}{2}}^{l-2}\frac{\epsilon}{20(\mu_1 rl)^{l/2}}\ns{P_S u_0}
\end{align*}
where the last inequality uses $\n{P_S u_0}\geq \frac{\mu_2 \delta}{\sqrt{d}}$ and assumes $\delta\leq \frac{1}{2^ll\sqrt{m}d^{l-2}(m+K)^{l-1} M_2}\cdot\frac{\mu_2}{\sqrt{d}} \pr{\frac{\mu_2}{2}}^{l-2}\frac{\epsilon}{20(\mu_1 rl)^{l/2}}.$

Next, let's analyze the regularizer $\lambda lP_S u_0.$ Its norm can be bounded as follows,
\[
\n{\lambda lP_S u_0}\leq \pr{\frac{\mu_2}{2}}^{l-2}\frac{\epsilon}{20(\mu_1 rl)^{l/2}}\n{P_S u_0},
\]
where we assume $\lambda\leq \frac{1}{l}\cdot \pr{\frac{\mu_2}{2}}^{l-2}\frac{\epsilon}{20(\mu_1 rl)^{l/2}}.$

Overall, we have
\begin{align*}
\ns{P_S u_1} &\geq \ns{P_Su_0}-\eta\inner{P_Su_0}{\nabla_uf(U_0,\bar{C}_0)}\\
             &\geq \ns{P_Su_0}+\eta\pr{\frac{\mu_2}{2}}^{l-2}\pr{\frac{\epsilon}{5(\mu_1 rl)^{l/2}}-\frac{\epsilon}{20(\mu_1 rl)^{l/2}}-\frac{\epsilon}{20(\mu_1 rl)^{l/2}}}\ns{P_Su_0}\\
             &= \pr{1+\eta \pr{\frac{\mu_2}{2}}^{l-2}\frac{\epsilon}{10(\mu_1 rl)^{l/2}}}\ns{P_S u_0}.
\end{align*}

\paragraph{Induction Step:} Suppose $\ns{P_S u_t}\geq \pr{1+\eta \pr{\frac{\mu_2}{2}}^{l-2}\frac{\epsilon}{10(\mu_1 rl)^{l/2}}}^t\ns{P_S u_0},$ we will prove that $\ns{P_S u_{t+1}}\geq  \pr{1+\eta \pr{\frac{\mu_2}{2}}^{l-2}\frac{\epsilon}{10(\mu_1 rl)^{l/2}}}^{t+1}\ns{P_S u_0}.$ Actually, we have assumed that $a(P_{S^{\otimes l}}T_t-P_{S^{\otimes l}}T^*)(\overline{ P_S u_t}^{\otimes l})\leq -\frac{\epsilon}{5(\mu_1 rl)^{l/2}}$, so we only need to show that
$c_t\n{P_S u_t}\geq \frac{\mu_2}{2}$. Based on these two properties, the remaining proofs are exactly the same as that for $t=0.$

The latter property is not hard to verify:

If $\n{u_t}> 2\sqrt{m+K}\delta$,
we know $c_t=1/\n{u_t}.$ Then, we have $\n{P_S u_t}c_t = \frac{\n{P_S u_t}}{\n{u_t}}\geq \frac{\n{u_t}-\n{P_B u_t}}{\n{u_t}}\geq \frac12,$ where we use $\n{P_B u_t}\leq \sqrt{m+K}\delta$.

If $\n{u_t}\leq 2\sqrt{m+K}\delta,$ we do not necessarily have $c_t=\frac{\sqrt{d(m+K)}}{\n{u_t}}$ because the norm of $\n{u_t}$ might first exceed the threshold and then drop below the threshold later. Note, by the induction proof, we only know the norm of $P_S u_t$ monotonically increase, which does not imply that $\n{u_t}$ monotonically increases. So, we have to consider both cases here. If $c_t=\frac{\sqrt{d(m+K)}}{\n{u_t}}$, we have $\n{P_S u_t}c_t=\frac{\sqrt{d(m+K)}\n{P_S u_t}}{\n{u_t}}\geq \frac{\mu_2}{2},$ which is because $\n{P_S u_t}\geq \n{P_S u_0}\geq \mu_2\delta/\sqrt{d}.$ If $c_t=1/\n{u_t},$ we know there exists $\tau\leq t$ such that $\n{u_\tau}>2\sqrt{m+K}\delta$. Since $\n{P_B u_\tau}\leq \sqrt{m+K}\delta,$ we know $\n{P_S u_\tau}\geq \sqrt{m+K}\delta.$ By the induction proof, we know $\n{P_S u_t}\geq \n{P_S u_\tau}\geq \sqrt{m+K}\delta.$ Then, we have $\n{P_S u_t}c_t=\frac{\n{P_S u_t}}{\n{u_t}}\geq \frac{1}{2}.$

This finishes the proof of Lemma~\ref{lem:keep_correlation_exp_norm_formal}.
\end{proofof}

Therefore the final step is to show that $a \overline{ P_S u_t}^{\otimes l}$ always has a large negative correlation with $P_{S^{\otimes l}}T_t-P_{S^{\otimes l}}T^*$, unless the function value has already decreased. The difficulty here is that both the current reinitialized component $u_t$ and other components are moving, therefore $T_t$ is also changing.

We can bound the change of $T - T^*$ by separating it into two terms, which are the change of the re-initialized component and the change of the residual:

\begin{align*}&
\absr{a(P_{S^{\otimes l}}T_t-P_{S^{\otimes l}}T^*)(\overline{ P_S u_t}^{\otimes l})-a(P_{S^{\otimes l}}T_0-P_{S^{\otimes l}}T^*)(\overline{ P_S u_0}^{\otimes l})}\\&
\leq \absr{\sum_{\tau=1}^t \pr{(P_{S^{\otimes l}}T_{\tau-1}-P_{S^{\otimes l}}T^*)(\overline{ P_S u_{\tau}}^{\otimes l})-(P_{S^{\otimes l}}T_{\tau-1}-P_{S^{\otimes l}}T^*)(\overline{ P_S u_{\tau-1}}^{\otimes l})} }\\&\qquad + \sum_{\tau=1}^t\fn{T_\tau-T_{\tau-1}}.
\end{align*}

The change of the re-initialized component has a small effect on the correlation because the change in $S$ subspace can only improve the correlation, and the influence of the $B$ subspace can be bounded. This is formally proved in the following lemma, which is the formal version of Lemma~\ref{lem:u_change}.

\begin{lemma}\label{lem:u_change_formal}
Assume $\delta\leq \frac{\mu_1}{m^{\frac34}\sqrt{\lambda}d^{\frac{l-2}{2}}(m+K)^{\frac{l-1}{2}}},\eta\leq \frac{\mu_2}{\lambda^{\frac32}m^{\frac94}  d^{\frac{3l-6}{2}}(m+K)^{\frac{3l-3}{2}}}$ for some constants $\mu_1,\mu_2.$ Assume $K\leq \frac{\lambda m}{14}$ and $\frac{10}{m}\leq\lambda\leq 1$.
Suppose at the beginning of one iteration, the tensor $T$ is parameterized by $(U,\bar{C}).$ Suppose $u$ is one column vector in $U$ with $\n{P_S u}\geq \frac{\mu_3\delta}{\sqrt{d}}$ where $\mu_3$ is a constant. Suppose $u'$ is $u$ after one step of gradient descent: $u'=u-\eta\nabla_u f(U,\bar{C}).$ We have
\[a(P_{S^{\otimes l}}T-P_{S^{\otimes l}}T^*)(\ol{P_S u'}^{\otimes l})\leq a(P_{S^{\otimes l}}T-P_{S^{\otimes l}}T^*)(\ol{P_S u}^{\otimes l})+\mu l^4 2^l d^{l-1.5}m^{1/2}(m+K)^{l-1}\eta\delta\lambda,\]
where $\mu$ is some constant.
\end{lemma}

\begin{proofof}{Lemma~\ref{lem:u_change_formal}}
Define $g(u):=a(P_{S^{\otimes l}}T-P_{S^{\otimes l}}T^*)(\ol{P_S u}^{\otimes l}).$ Note that in function $g(u),$ we view $T$ as fixed. We will show that the change of $g$ is bounded when the input changes from $u$ to $u'.$

\paragraph{Bounding first order change:} Let's first compute the gradient of $g$ at $u.$
\begin{align*}
\nabla g(u)
=& a l(P_{S^{\otimes l}}T-P_{S^{\otimes l}}T^*)((P_S u)^{\otimes l-1},P_S )\frac{1}{\n{P_S u}^l}
-al(P_{S^{\otimes l}}T-P_{S^{\otimes l}}T^*)((P_S u)^{\otimes l})\frac{P_S u}{\n{P_S u}^{l+2}}\\
=& al(P_{S^{\otimes l}}T-P_{S^{\otimes l}}T^*)((P_S u)^{\otimes l-1},P_S )\frac{1}{\n{P_S u}^l}
-al(P_{S^{\otimes l}}T-P_{S^{\otimes l}}T^*)((P_S u)^{\otimes l-1},\ol{P_S u})\frac{\ol{P_S u}}{\n{P_S u}^{l}}\\
=& al(P_{S^{\otimes l}}T-P_{S^{\otimes l}}T^*)((P_S u)^{\otimes l-1},P_S -\ol{P_S u}\cdot \ol{P_S u}^\top)\frac{1}{\n{P_S u}^l}\\
=& al(P_{S^{\otimes l}}T-P_{S^{\otimes l}}T^*)((P_S u)^{\otimes l-1},P_D)\frac{1}{\n{P_S u}^l},
\end{align*}
where $P_D$ is the projection matrix on the span of $S\setminus \{u\}.$ We can also compute the projection of $\nabla_u f(U,\bar{C})$ on $D$ as follows,
\begin{align*}
P_D \nabla_u f(U,\bar{C}) =& al(T-T^*)(u^{\otimes l-1},P_D)c^{l-2}.
\end{align*}
We can divide $l(T-T^*)(u^{\otimes l-1},P_D)c^{l-2}$ into $2^{l-1}$ terms, each of which corresponds to the projection of $u^{l-1}$ on a subspace. For subspace $S^{\otimes l-1},$ we have
\[al(T-T^*)((P_S u)^{\otimes l-1},P_D)c^{l-2}=al(P_{S^{\otimes l}}T-P_{S^{\otimes l}}T^*)((P_S u)^{\otimes l-1},P_D)c^{l-2},\]
which has non-negative inner product with $\nabla g(u).$ We can bound the norm of all the other terms. For any $l-1\geq k\geq 1,$ consider subspace $B^{\otimes k}\otimes S^{\otimes (l-1-k)},$ we can bound the norm of $al(T-T^*)((P_B u)^{\otimes k},(P_S u)^{\otimes (l-1-k)},P_D)c^{l-2}$ as follows:
\begin{align*}
&\n{al(T-T^*)((P_B u)^{\otimes k},(P_S u)^{\otimes l-1-k},P_D)c^{l-2}}\\
=& l\n{T((P_B u)^{\otimes k},(P_S u)^{\otimes l-1-k},P_D)c^{l-2}}\\
\leq& l\sum_{i=1}^m c_i^{l-2}\n{P_B u_i}^{k}\n{P_S u_i}^{l-k}\n{P_B u}^{k}\n{P_S u}^{l-1-k}c^{l-2}\\
\leq& l\sum_{i=1}^m c_i^{l-2}\n{P_B u_i}\n{ u_i}^{l-1}\n{P_B u}\n{ u}^{l-2}c^{l-2}\\
\leq& l\sum_{i=1}^m d^{l-2}(m+K)^{l-1}\delta^2\n{u_i}\\
\leq& l\sqrt{m}d^{l-2}(m+K)^{l-1}\delta^2 M_2,
\end{align*}
where the second last inequality comes from Lemma \ref{lem:bound_ortho_norm_formal}.

Denote $R$ as the summation of terms in all subspaces except for $S^{\otimes l-1}$, then
\[\n{R}\leq (2^{l-1}-1)l\sqrt{m}d^{l-2}(m+K)^{l-1}\delta^2 M_2.\]

Therefore, the first order change of $g$ can be bounded as follows,
\begin{align*}
&\inner{\nabla g(u)}{-\eta \nabla_u f(U,\bar{C})} \\
=& \inner{al(P_{S^{\otimes l}}T-P_{S^{\otimes l}}T^*)((P_S u)^{\otimes l-1},P_D)\frac{1}{\n{P_S u}^l}}{-\eta a l(P_{S^{\otimes l}}T-P_{S^{\otimes l}}T^*)((P_S u)^{\otimes l-1},P_D)c_u^{l-2}-\eta R}\\
\leq& \eta\n{l(P_{S^{\otimes l}}T-P_{S^{\otimes l}}T^*)((P_S u)^{\otimes l-1},P_D)\frac{1}{\n{P_S u}^l}}\n{R}\\
\leq& \eta l\sqrt{20}\frac{\sqrt{d}}{\mu_3\delta}\cdot (2^{l-1}-1)l\sqrt{m}d^{l-2}(m+K)^{l-1}\delta^2 M_2\\
\leq& \frac{10l^2 2^{l}}{\mu_3}\eta d^{l-1.5}\sqrt{m}(m+K)^{l-1}\delta M_2,
\end{align*}
where the second last inequality assumes $||P_Su||\geq\frac{\mu_3\delta}{\sqrt{d}}$.

\paragraph{Bounding higher order change:} For all $u''=(1-\theta)u+\theta u'$ with $0\leq \theta\leq 1,$ we prove a uniform upper bound for $\fn{\nabla^2 g(u'')}.$ Recall the gradient of $g$ at $u'',$
\[\nabla g(u'')
= a l(P_{S^{\otimes l}}T-P_{S^{\otimes l}}T^*)((P_S u'')^{\otimes l-1},P_D'')\frac{1}{\n{P_S u''}^l},\]
where $P_D''$ is the projection matrix to $S\setminus \{u''\}.$ We compute $\n{\nabla^2 g(u'')}$ as follows,
\begin{align*}
\nabla^2 g(u'')
=& a l(l-1)(P_{S^{\otimes l}}T-P_{S^{\otimes l}}T^*)((P_S u'')^{\otimes l-2},P_S,P_D'')\frac{1}{\n{P_S u''}^l}\\
&-a l^2(P_{S^{\otimes l}}T-P_{S^{\otimes l}}T^*)((P_S u'')^{\otimes l-1},P_D'')\otimes \frac{P_S u''}{\n{P_S u''}^{l+2}}.
\end{align*}
Therefore,
\[\fn{\nabla^2 g(u'')}\leq 2l^2\sqrt{20} \frac{1}{\ns{P_Su''}}.\]
Assume that $\eta\leq\frac{\mu_3\delta\sqrt{\lambda}}{2\sqrt{10d}\pr{\sqrt{20}l(\sqrt{d(m+K)})^{l-2}+\lambda l}}$ and from the proof of Lemma \ref{lem:f_value_decrease_formal} where we bound the gradient, we know that
\[\n{\eta\nabla_u f(U,\bar{C})}\leq\eta\pr{l\sqrt{20}(\sqrt{d(m+K)})^{l-2}+\lambda l}\n{u}\leq \frac{\sqrt{\frac{\lambda}{10}}\mu_3\delta}{2\sqrt{d}}\sqrt{\frac{10}{\lambda}}= \frac{\mu_3\delta}{2\sqrt{d}}.\]
Thus,
\begin{align*}
\n{P_Su''} &\geq \n{P_Su} - \n{P_Su''-P_Su}\\
           &\geq \n{P_Su} - \n{\theta\eta\nabla_u f(U,\bar{C})}\\
           &\geq \frac{\mu_3\delta}{\sqrt{d}} - \frac{\mu_3\delta}{2\sqrt{d}}= \frac{\mu_3\delta}{2\sqrt{d}}.
\end{align*}
Therefore,
\[\fn{\nabla^2 g(u'')}\leq 2l^2\sqrt{20} \frac{4d}{\mu_3^2\delta^2}.\]

Overall, we have
\begin{align*}
g(u')-g(u)\leq& \inner{\nabla g(u)}{-\eta \nabla_u f(U,\bar{C})}+\frac{\eta^2 }{2}2l^2\sqrt{20} \frac{4d}{\mu_3^2\delta^2}\ns{ \nabla_u f(U,\bar{C})}.
\end{align*}
Recall that,
\[\nabla_u f(U,\bar{C})=a l(T-T^*)(u^{\otimes l-1},I)c^{l-2}+\lambda l u,\]
we have
\begin{align*}
\fn{\nabla_u f(U,\bar{C})}\leq& l\sqrt{20}\max\pr{\n{u},2\sqrt{m+K}\delta(\sqrt{d(m+K)})^{l-2}}+\lambda l\n{u}\\
\leq& l\sqrt{20}\max\pr{M_2,2\sqrt{m+K}\delta(\sqrt{d(m+K)})^{l-2}}+\lambda lM_2\\
\leq& l\sqrt{20} M_2+\lambda l M_2,
\end{align*}
where the last inequality assumes $\delta\leq \frac{M_2}{2\sqrt{m+K}(\sqrt{d(m+K)})^{l-2}}$.

Finally, we have
\begin{align*}
g(u')-g(u)\leq& \inner{\nabla g(u)}{-\eta \nabla_u f(U,\bar{C})}+\frac{\eta^2 }{2}2l^2\sqrt{20} \frac{4d}{\mu_3^2\delta^2}\ns{ \nabla_u f(U,\bar{C})}\\
\leq& \frac{10l^2 2^{l}}{\mu_3}\eta d^{l-1.5}\sqrt{m}(m+K)^{l-1}\delta M_2+\frac{\eta^2 }{2}2l^2\sqrt{20} \frac{4d}{\mu_3^2\delta^2}\pr{ l\sqrt{20} M_2+\lambda l M_2}^2\\
\leq& \frac{10l^2 2^{l}}{\mu_3}\eta d^{l-1.5}\sqrt{m}(m+K)^{l-1}\sqrt{\frac{10}{\lambda}}\delta + \sqrt{20}\eta^2 l^2 \frac{4d}{\mu_3^2\delta^2}\pr{40l^2\frac{10}{\lambda}+2\lambda^2 l^2\frac{10}{\lambda}}\\
\leq& \mu l^4 2^l d^{l-1.5}m^{1/2}(m+K)^{l-1}\eta\delta\lambda,
\end{align*}
where the last inequality assumes $l\geq 3$, $\eta\leq \delta^3$ and $\mu$ is some constant.
\end{proofof}

Therefore, the only way to change the residual term by a lot must be changing the tensor $T$, and the accumulated change of $T$ is strongly correlated with the decrease of $f$. This is similar to the technique of bounding the function value decrease in \cite{wei2019regularization}. The connection between them are formalized in the following lemma, which is the formal version of Lemma~\ref{lem:tensor_bounded_epoch}:

\begin{lemma}\label{lem:tensor_bounded_epoch_formal}
Assume that $\delta\leq \frac{\mu_1}{m^{\frac34}\sqrt{\lambda}d^{\frac{l-2}{2}}(m+K)^{\frac{l-1}{2}}},\eta\leq \frac{\mu_2\lambda}{m^{\frac12} l^4 d^{\frac{l-1}{2}(m+K)^{\frac{l-2}{2}}}}$ for some constants $\mu_1,\mu_2$, and $\frac{10}{m}\leq\lambda\leq 1$.
Within one epoch, let $T_0$ be the tensor after reinitialization, and let $T_t$ be the tensor at the end of the $t$-th iteration. Let $(U_0,\bar{C}_0)$ be the parameters after the reinitialization step and let $(U_H,\bar{C}_H)$ be the parameters at the end of this epoch. We have
\begin{align*}
&\sum_{\tau=1}^H\fn{T_\tau - T_{\tau-1}}\\
\leq& 200l^{2.5}\sqrt{\frac{1}{\lambda}}\sqrt{\eta H}\sqrt{f(U_0,\bar{C}_0)-f(U_H,\bar{C}_H)+160m(m+K)\delta^2(\sqrt{d(m+K)})^{l-2}}\\
&+ 16m(m+K)\delta^2(\sqrt{d(m+K)})^{l-2}.
\end{align*}
\end{lemma}

Intuitively, if we are doing a standard gradient descent, at each step the change in function value is going to be proportional to the square of the change in the tensor $T$, and the guarantee similar to the Lemma above can be proved by applying Cauchy-Schwartz. However, in our setting the proof becomes more complicated because of the normalization steps and in particular the scalar mode switch.

Before proving Lemma~\ref{lem:tensor_bounded_epoch_formal}, we first prove the following lemma which guarantees the function value decrease in one step (without scalar mode switch): 

\begin{lemma}\label{lem:tensor_bounded_iteration_formal}
Assume $\delta\leq \frac{\mu_1}{m^{\frac34}\sqrt{\lambda}d^{\frac{l-2}{2}}(m+K)^{\frac{l-1}{2}}},\eta\leq \frac{\mu_2\lambda}{m^{\frac12} l^4 d^{\frac{l-1}{2}(m+K)^{\frac{l-2}{2}}}}$ for some constants $\mu_1,\mu_2$, and $\eta\leq\delta^3$. Assume $K\leq \frac{\lambda m}{14}.$
Starting from $T$ parameterized by $(U,\bar{C})$, suppose after one iteration (before potential scalar mode switch) the tensor becomes $T'$ parameterized by $(U',\bar{C}').$ We have
\[\fn{T'-T}\leq 200l^2 \sqrt{\frac{1}{\lambda}}\fn{\eta \nabla_U f(U,\bar{C})}.\]
\end{lemma}

\begin{proofof}{Lemma~\ref{lem:tensor_bounded_iteration_formal}}
According to the algorithm, we know each iteration is composed of two steps: update $U$ by gradient descent ($U'=U-\eta\nabla_U f(U,\bar{C})$) and update $C$ and $\hat{C}$ according to $U'.$ Let $\hat{T}$ be the intermediate tensor parameterized by $(U',\bar{C}).$ We will bound $\fn{T'-T}$ by bounding $\fn{\hat{T}-T}$ and $\fn{T'-\hat{T}}$ separately.

According to Lemma~\ref{lem:U_bounded_formal}, we know $\sum_{i=1}^m\ns{u_i},\sum_{i=1}^m\ns{u_i'}\leq \frac{10}{\lambda}.$ Denote $M_2^2 = \frac{10}{\lambda}.$

\paragraph{Bounding $\fn{\hat{T}-T}$:} From $T$ to $\hat{T},$ $U$ is updated to $U'=U-\eta\nabla_U f(U,\bar{C})$ while $C$ and $\hat{C}$ remains the same. Therefore,
\begin{align*}
\fn{\hat{T}-T}
=& \fn{\sum_{i=1}^m a_ic_i^{l-2}\pr{ u_i-\eta\nabla_{u_i}f(U,\bar{C})}^{\otimes l}-\sum_{i=1}^m a_ic_i^{l-2}u_i^{\otimes l}}\\
\leq& \sum_{i=1}^m l\n{u_i}^{l-1}\n{\eta\nabla_{u_i}f(U,\bar{C})}c_i^{l-2} + \sum_{i=1}^m \sum_{k=2}^l \binom{l}{k}\n{u_i}^{l-k}\n{\eta\nabla_{u_i}f(U,\bar{C})}^kc_i^{l-2}.
\end{align*}
We can further bound the linear term as follows:
\begin{align*}
\sum_{i=1}^m l\n{u_i}^{l-1}\n{\eta\nabla_{u_i}f(U,\bar{C})}c_i^{l-2}
\leq& l\sum_{i=1}^m \n{\eta\nabla_{u_i}f(U,\bar{C})}\max(\n{u_i}, 2\sqrt{m+K}\delta(\sqrt{d(m+K)})^{l-2})\\
\leq& l\sqrt{\sum_{i=1}^m \ns{\eta\nabla_{u_i}f(U,\bar{C})}}\sqrt{\sum_{i=1}^m\max(\ns{u_i}, 4(m+K)^{l-1}\delta^2 d^{l-2})}\\
\leq& \sqrt{2}lM_2\eta \fn{\nabla_U f(U,\bar{C})},
\end{align*}
where the last inequality assumes $\delta^2\leq \frac{M_2^2}{4m(m+K)^{l-1} d^{l-2}}.$

According to the proof in Lemma~\ref{lem:f_value_decrease_formal}, we know $\n{\eta\nabla_{u_i}f(U,\bar{C})}\leq \frac{1}{l}\n{u_i}$. Therefore, for the higher order terms, for each $k\geq 2,$
\begin{align*}
\sum_{i=1}^m \binom{l}{k}\n{u_i}^{l-k}\n{\eta\nabla_{u_i}f(U,\bar{C})}^kc_i^{l-2}
\leq& \sum_{i=1}^m l^k \n{u_i}^{l-k} \frac{\n{u_i}^{k-1}}{l^{k-1}}\n{\eta\nabla_{u_i}f(U,\bar{C})} c_i^{l-2}\\
\leq& \sum_{i=1}^m l \n{u_i}^{l-1} \n{\eta\nabla_{u_i}f(U,\bar{C})} c_i^{l-2}\\
\leq& \sqrt{2}lM_2\eta \fn{\nabla_U f(U,\bar{C})}.
\end{align*}

Overall, we have
\[\fn{\hat{T}-T}\leq 2\sqrt{2}l^2 M_2\eta \fn{\nabla_U f(U,\bar{C})}.\]

\paragraph{Bounding $\fn{T'-\hat{T}}$:} From $\hat{T}$ to $T',$ we update $C$ to $C'$ and $\hat{C}$ to $\hat{C}'$ such that $\forall i\in[m], c_i'=c_i\frac{\n{u_i}}{\n{u_i'}}$ and $\hat{c}_i'=\hat{c}_i\frac{\n{u_i}}{\n{u_i'}}$. Thus,
\begin{align*}
\fn{T'-\hat{T}} =& \fn{\sum_{i=1}^m a_i(c_i')^{l-2}(u_i')^{\otimes l}-\sum_{i=1}^m a_ic_i^{l-2}(u_i')^{\otimes l} }\\
\leq& \sum_{i=1}^m\absr{(c_i')^{l-2}-c_i^{l-2}}\n{u_i'}^l.
\end{align*}
Now, let's focus on the change in $c_i^{l-2}.$ Define $g(u)=\frac{1}{\n{u}^{l-2}}.$ We have,
\begin{align*}
\nabla g(u)=-(l-2)\frac{u}{\n{u}^l}\text{ and }
\nabla^2 g(u)= -(l-2)\frac{I}{\n{u}^l}+l(l-2)\frac{uu^\top}{\n{u}^{l+2}}.
\end{align*}
Therefore, the spectral norm of $\nabla^2 g(u)$ is bounded by $l^2/\n{u}^l.$

For any $i\in[m],$ let $u_i''$ be any point on the line segment between $u_i$ and $u_i',$ then $\n{\nabla^2 g(u_i'')}_2\leq l^2/\n{u_i''}^l.$ If $c_i=1/\n{u_i},$ we have
\begin{align*}
\absr{(c_i')^{l-2}-c_i^{l-2}}
\leq& \n{\nabla g(u_i)}\n{\eta \nabla_{u_i}f(U,\bar{C}) }+\frac{1}{2}\max_{u_i''} \n{\nabla^2 g(u_i'')}_2\ns{\eta \nabla_{u_i}f(U,\bar{C})}\\
\leq& \frac{l-2}{\n{u_i}^{l-1}}\n{\eta \nabla_{u_i}f(U,\bar{C})} + \frac{1}{2}\max_{u_i''} \frac{l\n{u_i}}{\n{u_i''}^l} \n{\eta \nabla_{u_i}f(U,\bar{C})}.
\end{align*}
If $c_i=\sqrt{d(m+K)}/\n{u_i},$ we have
\begin{align*}
\absr{(c_i')^{l-2}-c_i^{l-2}}
\leq \frac{l-2}{\n{u_i}^{l-1}}\n{\eta \nabla_{u_i}f(U,\bar{C})}(\sqrt{d(m+K)})^{l-2} + \frac{1}{2}\max_{u_i''} \frac{l\n{u_i}}{\n{u_i''}^l} \n{\eta \nabla_{u_i}f(U,\bar{C})}(\sqrt{d(m+K)})^{l-2}.
\end{align*}
Therefore, we have
\begin{align*}
\fn{T'-\hat{T}}\leq& 2el\sum_{i=1}^m \n{\eta \nabla_{u_i}f(U,\bar{C})}\max(\n{u_i},2\sqrt{m+K}\delta(\sqrt{d(m+K)})^{l-2})\\
\leq& 2\sqrt{2}elM_2 \fn{\eta \nabla_U f(U,\bar{C})},
\end{align*}
where the first inequality holds because $\n{u_i'}\leq \pr{1+\frac1l}\n{u_i}$ and the second inequality assumes $\delta^2 \leq \frac{M_2^2}{4m(m+K)^{l-1} d^{l-2}}.$

Overall, combing the bounds on $\fn{\hat{T}-T}$ and $\fn{T'-\hat{T}},$ we have
\begin{align*}
\fn{T'-T}\leq 200l^2 \sqrt{\frac{1}{\lambda}}\fn{\eta \nabla_U f(U,\bar{C})}.
\end{align*}
\end{proofof}

Now we are ready to prove Lemma~\ref{lem:tensor_bounded_epoch_formal}.

\begin{proofof}{Lemma~\ref{lem:tensor_bounded_epoch_formal}}
Let's first bound the tensor change and function value change due to scalar mode switches. Following the proof of Claim \ref{claim:scaling-switch} in Lemma~\ref{lem:f_increase_bounded_formal}, setting $\tilde{\Gamma}=10$ and assuming $\eta\leq \frac{1}{l(\sqrt{2\Gamma}(\sqrt{d(m+K)})^{l-2}+\lambda)}$, we know each scalar mode switch can at most change the tensor Frobenius norm by $16(m+K)\delta^2(\sqrt{d(m+K)})^{l-2}$. Furthermore, using the same argument as Claim \ref{claim:scaling-switch}, the function value can increase by at most $\sqrt{20}\pr{16(m+K)\delta^2(\sqrt{d(m+K)})^{l-2}}+\frac12\pr{16(m+K)\delta^2(\sqrt{d(m+K)})^{l-2}}^2\leq160(m+K)\delta^2(\sqrt{d(m+K)})^{l-2},$ where we assume $\delta^2\leq\frac{5}{8(m+K)(\sqrt{d(m+K)})^{l-2}}$.

According to the algorithm, we know each epoch contains at most $m$ scalar mode switches. Suppose $T_\tau'$ be the tensor before potential scalar mode switch in the $\tau$-th iteration. Then, we have
\begin{align*}
\sum_{\tau=1}^t\fn{T_\tau - T_{\tau-1}}\leq& \sum_{\tau=1}^t\fn{T_\tau' - T_{\tau-1}} + \sum_{\tau=1}^t\fn{T_\tau - T_\tau'}\\
\leq& \sum_{\tau=1}^t\fn{T_\tau' - T_{\tau-1}} + 16m(m+K)\delta^2(\sqrt{d(m+K)})^{l-2}.
\end{align*}
According to Lemma~\ref{lem:tensor_bounded_iteration_formal}, we know
\[\fn{T_\tau' - T_{\tau-1}}\leq 200l^2 \sqrt{\frac{1}{\lambda}}\fn{\eta \nabla_U f(U_{\tau-1},\bar{C}_{\tau-1})}.\]
Therefore, we have
\begin{align*}
\sum_{\tau=1}^t\fn{T_\tau' - T_{\tau-1}}
\leq& 200l^2 \sqrt{\frac{1}{\lambda}}\sum_{\tau=1}^t\fn{\eta \nabla_U f(U_{\tau-1},\bar{C}_{\tau-1})}\\
\leq& 200l^2 \sqrt{\frac{1}{\lambda}}\sqrt{t}\sqrt{\sum_{\tau=1}^t\fns{\eta \nabla_U f(U_{\tau-1},\bar{C}_{\tau-1})}}.
\end{align*}
According to Lemma~\ref{lem:f_value_decrease_formal}, we know $f(U_\tau',\bar{C}_\tau')-f(U_{\tau-1},\bar{C}_{\tau-1})\leq -\frac{\eta}{l}\fns{\nabla_U f(U_{\tau-1},\bar{C}_{\tau-1})}.$ Therefore, we have
\begin{align*}
\sum_{\tau=1}^t\fn{T_\tau' - T_{\tau-1}}
\leq& 200l^2 \sqrt{\frac{1}{\lambda}}\sqrt{t}\sqrt{\sum_{\tau=1}^t\eta l\pr{f(U_{\tau-1},\bar{C}_{\tau-1})-f(U_\tau',\bar{C}_\tau') }}.
\end{align*}
Since scalar mode switches in total change the function value by at most $160m(m+K)\delta^2(\sqrt{d(m+K)})^{l-2},$ we know
\begin{align*}
&\sum_{\tau=1}^t\pr{f(U_{\tau-1},\bar{C}_{\tau-1})-f(U_\tau',\bar{C}_\tau')}\\
\leq& f(U_0,\bar{C}_0)-f(U_t,\bar{C}_t)+160m(m+K)\delta^2(\sqrt{d(m+K)})^{l-2}.
\end{align*}
Overall, we have
\begin{align*}
&\sum_{\tau=1}^t\fn{T_\tau - T_{\tau-1}}\\
\leq& 200l^{2.5}\sqrt{\frac{1}{\lambda}}\sqrt{\eta H}\sqrt{f(U_0,\bar{C}_0)-f(U_t,\bar{C}_t)+160m(m+K)\delta^2(\sqrt{d(m+K)})^{l-2}}\\
&+ 16m(m+K)\delta^2(\sqrt{d(m+K)})^{l-2}.
\end{align*}
\end{proofof}

Combining all the steps above, we are now ready to prove Lemma~\ref{lem:norm_explode_formal}.

\begin{proofof}{Lemma~\ref{lem:norm_explode_formal}}
Let $u_0$ be the reinitialized vector. According to Lemma~\ref{lem:initial_correlate_formal}, we know with probability at least $1/5$,
\[a(P_{S^{\otimes l}}T_0'-P_{S^{\otimes l}}T^*)(\overline{ P_S u_0}^{\otimes l})\leq \frac{-1}{(\mu_1 rl)^{l/2}}\fn{P_{S^{\otimes l}}T_0'-P_{S^{\otimes l}}T^*}\leq -\frac{\epsilon}{(\mu_1 rl)^{l/2}},\]
where $\mu_1$ is some constant. According to Lemma \ref{lem:projection}, we know with probability at least $1-1/30,$
$\n{P_S u_0}\geq \frac{\mu_2\delta}{\sqrt{d}}$
for some constant $\mu_2<1.$ Taking a union bound, we know both properties hold with probability at least $1/6.$

Conditioning on both properties, we will prove that
\[f(U_0,\bar{C}_0)-f(U_H,\bar{C}_H)\geq \frac{ \lambda}{32000000(\mu_1 rl)^l\eta H l^5} \epsilon^2.\]
For the sake of contradiction, assume that $f(U_0,\bar{C}_0)-f(U_H,\bar{C}_H)\leq \frac{ \lambda}{32000000(\mu_1 rl)^l\eta H l^5} \epsilon^2$. According to Lemma~\ref{lem:tensor_bounded_epoch_formal}, we know
\begin{align*}
\sum_{\tau=1}^H\fn{T_\tau - T_{\tau-1}}\leq \frac{\epsilon}{10(\mu_1 rl)^{l/2}}
\end{align*}
as long as $\delta^2\leq \frac{\epsilon}{320(\mu_1 rl)^{l/2}m(m+K)^{\frac{l}{2}}d^{\frac{l-2}{2}}}$ and $\delta^2\leq \frac{\lambda\epsilon^2}{32000000(\mu_1 rl)^l\eta H l^5\cdot 160m(m+K)^{\frac{l}{2}}d^{\frac{l-2}{2}}}.$

We will prove that $a(P_{S^{\otimes l}}T_t-P_{S^{\otimes l}}T^*)(\overline{ P_S u_t}^{\otimes l})\leq -\frac{\epsilon}{5(C_1 rl)^{l/2}}$ for all $0\leq t\leq H-1$, so from Lemma~\ref{lem:keep_correlation_exp_norm_formal} we know that the norm of $P_S u_t$ must increase exponentially.

Let's first prove the case at the beginning of an epoch: Let $T_0$ be the tensor after reinitialization. According to the proof of Claim \ref{claim:re-init} in Lemma~\ref{lem:f_increase_bounded_formal}, we know
\begin{align*}
\fn{T_0-T_0'}\leq 2\sqrt{\frac{10}{\lambda m}}\leq \frac{\epsilon}{2(\mu_1 rl)^{l/2}},
\end{align*}
where the last inequality assumes $\lambda m\geq \frac{160(\mu_1 rl)^l}{\epsilon^2}.$ This implies that
\[a(P_{S^{\otimes l}}T_0-P_{S^{\otimes l}}T^*)(\overline{ P_S u_0}^{\otimes l})\leq a(P_{S^{\otimes l}}T_0'-P_{S^{\otimes l}}T^*)(\overline{ P_S u_0}^{\otimes l}) + \fn{T_0-T_0'}\leq -\frac{\epsilon}{2(\mu_1 rl)^{l/2}}.\]

For later steps, we will show that $a(P_{S^{\otimes l}}T_t-P_{S^{\otimes l}}T^*)(\overline{ P_S u_t}^{\otimes l})$ is close to $a(P_{S^{\otimes l}}T_0-P_{S^{\otimes l}}T^*)(\overline{ P_S u_0}^{\otimes l}).$ Actually,
\begin{align*}
&\absr{a(P_{S^{\otimes l}}T_t-P_{S^{\otimes l}}T^*)(\overline{ P_S u_t}^{\otimes l})-a(P_{S^{\otimes l}}T_0-P_{S^{\otimes l}}T^*)(\overline{ P_S u_0}^{\otimes l})}\\
\leq& \absr{\sum_{\tau=1}^t \pr{(P_{S^{\otimes l}}T_{\tau-1}-P_{S^{\otimes l}}T^*)(\overline{ P_S u_{\tau}}^{\otimes l})-(P_{S^{\otimes l}}T_{\tau-1}-P_{S^{\otimes l}}T^*)(\overline{ P_S u_{\tau-1}}^{\otimes l})} }\\
&+ \absr{\sum_{\tau=1}^t \pr{(P_{S^{\otimes l}}T_{\tau}-P_{S^{\otimes l}}T^*)(\overline{ P_S u_{\tau}}^{\otimes l})-(P_{S^{\otimes l}}T_{\tau-1}-P_{S^{\otimes l}}T^*)(\overline{ P_S u_{\tau}}^{\otimes l})} }\\
\leq& H\mu l^4 2^l d^{l-1.5}m^{1/2}(m+K)^{l-1}\eta\delta\lambda + \sum_{\tau=1}^t\n{T_\tau-T_{\tau-1}}\\
\leq& H\mu l^4 2^l d^{l-1.5}m^{1/2}(m+K)^{l-1}\eta\delta\lambda + \frac{\epsilon}{10(\mu_1 rl)^{l/2}}
\leq \frac{\epsilon}{5(\mu_1 rl)^{l/2}}.
\end{align*}
The second inequality above comes from Lemma~\ref{lem:u_change_formal}, and the last inequality assumes $\delta\leq \frac{1}{\mu l^4 2^l d^{l-1.5}m^{1/2}(m+K)^{l-1}\eta\lambda H} \cdot \frac{\epsilon}{10(\mu_1 rl)^{l/2}}.$

This then implies that for all $0\leq t\leq H-1$,
\[a(P_{S^{\otimes l}}T_t-P_{S^{\otimes l}}T^*)(\overline{P_S u_t}^{\otimes l})\leq -\frac{\epsilon}{2(\mu_1 rl)^{l/2}}+\frac{\epsilon}{5(\mu_1 rl)^{l/2}}\leq-\frac{\epsilon}{5(\mu_1 rl)^{l/2}}.\]

Then according to Lemma~\ref{lem:keep_correlation_exp_norm_formal},
\begin{align*}
\ns{P_S u_{H}}\geq& \pr{1+\eta \pr{\frac{\mu_2}{2}}^{l-2}\frac{\epsilon}{10(\mu_1 rl)^{l/2}}}^H\ns{P_S u_0}\\
\geq& \pr{1+\eta \pr{\frac{\mu_2}{2}}^{l-2}\frac{\epsilon}{10(\mu_1 rl)^{l/2}}}^H\frac{\mu_2^2 \delta^2}{d}\\
\geq& \exp\pr{\frac{1}{2}\eta H \pr{\frac{\mu_2}{2}}^{l-2}\frac{\epsilon}{10(\mu_1 rl)^{l/2}} }\frac{\mu_2^2 \delta^2}{d},
\end{align*}
where the last inequality assumes $\eta\leq\pr{\frac{2}{\mu_2}}^{l-2}\frac{10(\mu_1 rl)^{l/2}}{\epsilon}$.
Therefore, $\ns{P_S u_{H}}$ exceeds $M_2$ as long as $\eta H\geq 2\pr{\frac{2}{\mu_2}}^{l-2}\frac{10(\mu_1 rl)^{l/2} }{\epsilon}\log\pr{\frac{d M_2}{\mu_2^2\delta^2}}.$ Since $M_2=\sqrt{\frac{10}{\lambda}}$ is the upper bound of $\fn{U}$, this finishes the contradiction proof.

We have shown that
\[f(U_0,\bar{C}_0)-f(U_H,\bar{C}_H)\geq \frac{\lambda}{32000000(\mu_1 rl)^l\eta H l^5} \epsilon^2.\]
In order to show $f(U_0',\bar{C}_0')-f(U_H,\bar{C}_H)$ is large, we still need to bound $|f(U_0',\bar{C}_0')-f(U_0,\bar{C}_0)|$ that comes from reinitialization. According to Lemma~\ref{lem:f_increase_bounded_formal}, we know
\[|f(U_0',\bar{C}_0')-f(U_0,\bar{C}_0)|\leq \frac{200}{\lambda m}\leq \frac{\lambda}{64000000(\mu_1 rl)^l\eta H l^5} \epsilon^2,\]
where the second inequality assumes $\lambda^2m\geq 1.28\times 10^{11}(\mu_1 rl)^l\eta H l^5.$
Therefore,
\begin{align*}
    &f(U_H,\bar{C}_H)-f(U_0', \bar{C}_0')\\
\leq&\pr{f(U_0, \bar{C}_0)-f(U_H,\bar{C}_H)}+|f(U_0,\bar{C}_0)-f(U_0',\bar{C}_0')|\\
\leq&-\frac{\lambda}{3.2\times 10^7(\mu_1 rl)^l\eta H l^5} \epsilon^2+\frac{\lambda}{6.4\times 10^8(\mu_1 rl)^l\eta H l^5} \epsilon^2\\
\leq&-\frac{\lambda}{6.4\times 10^7(\mu_1 rl)^l\eta H l^5}\epsilon^2.
\end{align*}

We choose $m=O\pr{\frac{r^{2.5l}}{\epsilon^5}\log(d/\epsilon)}$, $\lambda=O\pr{\frac{\epsilon}{r^{0.5l}}}$, $\delta=O\pr{\frac{\epsilon^{5l-1.5}}{d^{l-1.5}(\log(d/\epsilon))^{l+0.5}r^{2.5l^2-0.75l}}}$, $\eta=O\pr{\frac{\epsilon^{15l-4.5}}{d^{3l-4.5}(\log(d/\epsilon))^{3l+1.5}r^{7.5l^2-2.25l}}}$, $H=O\pr{\frac{d^{3l-4.5}(\log(d/\epsilon))^{3l+2.5}r^{7.5l^2-1.75l}}{\epsilon^{15l-3.5}}}$ and $K=O\pr{\frac{r^{2l}}{\epsilon^4}\log(d/\epsilon)}$ such that all the conditions are satisfied and the function value decreases by $\Omega(\frac{\epsilon^{4}}{r^{2l}\log(d/\epsilon)})$ in each epoch. Note that there does exist some circular dependency between the parameters. This turns out to be not an issue in our proof because for example $\delta$ depends on $\frac{1}{\eta H}$ while $\eta H$ only depends logarithmically on $1/\delta$. Other circular dependencies can be resolved in the same manner. 
\end{proofof}

%% file: tools.tex
\section{Tools}
In this section, we give the technical lemmas we use in the proof.
\subsection{Random projection on a subspace}
We use the following lemma to show that with good probability, the projection of the reinitialized component on the good subspace is lower bounded.

\begin{lemma}[Lemma 2.2 in~\cite{dasgupta2003elementary}]\label{lem:projection}
Let $Y$ be a $d$-dimensional vector uniformly sampled from sphere $\mathbb{S}^{d-1}.$ Let $Z\in \R^k$ be the projection of $Y$ onto its first $k$ coordinates ($k<d$). For any $\beta<1,$ we have
\[\Pr\br{\ns{Z}\leq \frac{\beta k}{d} }\leq \exp\pr{\frac{k}{2}(1-\beta+\ln\beta)}.\]
\end{lemma}

\subsection{Norm of random Gaussian vectors}
The following lemma gives the concentration of $\ell_2$ norm of a random Gaussian vector.
\begin{lemma}[Theorem 3.1.1 in~\cite{vershynin2018high}]\label{lem:norm_vector}
Let $X=(X_1, X_2, \cdots, X_n)\in \R^n$ be a random vector with each entry independently sampled from $\mathcal{N}(0,1).$
Then
\[\Pr\br{\absr{\n{x}-\sqrt{n}}\geq t}\leq 2\exp\pr{-t^2/C^2},\]
where $C$ is an absolute constant.
\end{lemma}

\subsection{Anti-concentration of Gaussian polynomials}
We use anti-concentration of Gaussian polynomials to argue that a randomly initialized component has good correlation with the residual.
\begin{lemma}[Theorem 8 in~\cite{carbery2001distributional}]\label{lem:anti-concentration}
Let $x\in\R^n$ be a Gaussian variable $x\in N(0,I)$, for any polynomial $p(x)$ of degree $l$, there exists a constant $\kappa$ such that
\[\Pr\big[|p(x)|\leq \epsilon\sqrt{Var[p(x)]}\big]\leq \kappa l\epsilon^{1/l}.\]
\end{lemma}

%% file: ms.bbl
\begin{thebibliography}{}

\bibitem[Allen-Zhu and Li, 2019]{allen2019can}
Allen-Zhu, Z. and Li, Y. (2019).
\newblock What can resnet learn efficiently, going beyond kernels?
\newblock {\em arXiv preprint arXiv:1905.10337}.

\bibitem[Allen-Zhu et~al., 2018]{allen2018learning}
Allen-Zhu, Z., Li, Y., and Liang, Y. (2018).
\newblock Learning and generalization in overparameterized neural networks,
  going beyond two layers.
\newblock {\em arXiv preprint arXiv:1811.04918}.

\bibitem[Andoni et~al., 2014]{andoni2014learning}
Andoni, A., Panigrahy, R., Valiant, G., and Zhang, L. (2014).
\newblock Learning polynomials with neural networks.
\newblock In {\em International conference on machine learning}, pages
  1908--1916.

\bibitem[Arora et~al., 2019a]{arora2019exact}
Arora, S., Du, S.~S., Hu, W., Li, Z., Salakhutdinov, R., and Wang, R. (2019a).
\newblock On exact computation with an infinitely wide neural net.
\newblock {\em arXiv preprint arXiv:1904.11955}.

\bibitem[Arora et~al., 2019b]{arora2019fine}
Arora, S., Du, S.~S., Hu, W., Li, Z., and Wang, R. (2019b).
\newblock Fine-grained analysis of optimization and generalization for
  overparameterized two-layer neural networks.
\newblock {\em arXiv preprint arXiv:1901.08584}.

\bibitem[Bai et~al., 2020]{bai2020taylorized}
Bai, Y., Krause, B., Wang, H., Xiong, C., and Socher, R. (2020).
\newblock Taylorized training: Towards better approximation of neural network
  training at finite width.
\newblock {\em arXiv preprint arXiv:2002.04010}.

\bibitem[Bai and Lee, 2019]{bai2019beyond}
Bai, Y. and Lee, J.~D. (2019).
\newblock Beyond linearization: On quadratic and higher-order approximation of
  wide neural networks.
\newblock {\em arXiv preprint arXiv:1910.01619}.

\bibitem[Carbery and Wright, 2001]{carbery2001distributional}
Carbery, A. and Wright, J. (2001).
\newblock Distributional and $l^q$ norm inequalities for polynomials over
  convex bodies in $r^n$.
\newblock {\em Mathematical research letters}, 8(3):233--248.

\bibitem[Cardoso, 1991]{cardoso1991super}
Cardoso, J.-F. (1991).
\newblock Super-symmetric decomposition of the fourth-order cumulant tensor.
  blind identification of more sources than sensors.
\newblock In {\em International Conference on Acoustics, Speech, \& Signal
  Processing, Icassp}.

\bibitem[Chizat and Bach, 2018a]{chizat2018note}
Chizat, L. and Bach, F. (2018a).
\newblock A note on lazy training in supervised differentiable programming.
\newblock {\em arXiv preprint arXiv:1812.07956}, 8.

\bibitem[Chizat and Bach, 2018b]{chizat2018global}
Chizat, L. and Bach, F. (2018b).
\newblock On the global convergence of gradient descent for over-parameterized
  models using optimal transport.
\newblock In {\em Advances in neural information processing systems}, pages
  3040--3050.

\bibitem[Daniely, 2019]{daniely2019neural}
Daniely, A. (2019).
\newblock Neural networks learning and memorization with (almost) no
  over-parameterization.
\newblock {\em arXiv preprint arXiv:1911.09873}.

\bibitem[Dasgupta and Gupta, 2003]{dasgupta2003elementary}
Dasgupta, S. and Gupta, A. (2003).
\newblock An elementary proof of a theorem of johnson and lindenstrauss.
\newblock {\em Random Structures \& Algorithms}, 22(1):60--65.

\bibitem[Du et~al., 2018a]{du2018gradientb}
Du, S.~S., Lee, J.~D., Li, H., Wang, L., and Zhai, X. (2018a).
\newblock Gradient descent finds global minima of deep neural networks.
\newblock {\em arXiv preprint arXiv:1811.03804}.

\bibitem[Du et~al., 2018b]{du2018gradient}
Du, S.~S., Zhai, X., Poczos, B., and Singh, A. (2018b).
\newblock Gradient descent provably optimizes over-parameterized neural
  networks.
\newblock {\em arXiv preprint arXiv:1810.02054}.

\bibitem[Dyer and Gur-Ari, 2019]{dyer2019asymptotics}
Dyer, E. and Gur-Ari, G. (2019).
\newblock Asymptotics of wide networks from feynman diagrams.
\newblock {\em arXiv preprint arXiv:1909.11304}.

\bibitem[Ge et~al., 2015]{ge2015escaping}
Ge, R., Huang, F., Jin, C., and Yuan, Y. (2015).
\newblock Escaping from saddle points—online stochastic gradient for tensor
  decomposition.
\newblock In {\em Proceedings of The 28th Conference on Learning Theory}, pages
  797--842.

\bibitem[Ge et~al., 2016]{ge2016matrix}
Ge, R., Lee, J.~D., and Ma, T. (2016).
\newblock Matrix completion has no spurious local minimum.
\newblock In {\em Advances in Neural Information Processing Systems}, pages
  2973--2981.

\bibitem[Ge et~al., 2018]{ge2017learning}
Ge, R., Lee, J.~D., and Ma, T. (2018).
\newblock Learning one-hidden-layer neural networks with landscape design.
\newblock In {\em International Conference on Learning Representations}.

\bibitem[Ghorbani et~al., 2019]{ghorbani2019limitations}
Ghorbani, B., Mei, S., Misiakiewicz, T., and Montanari, A. (2019).
\newblock Limitations of lazy training of two-layers neural networks.
\newblock {\em arXiv preprint arXiv:1906.08899}.

\bibitem[Harshman, 1970]{harshman1970foundations}
Harshman, R. (1970).
\newblock Foundations of the parafac procedure: Model and conditions for an
  explanatory factor analysis.
\newblock {\em Technical Report UCLA Working Papers in Phonetics 16, University
  of California, Los Angeles, Los Angeles, CA}.

\bibitem[Hillar and Lim, 2013]{hillar2013most}
Hillar, C.~J. and Lim, L.-H. (2013).
\newblock Most tensor problems are np-hard.
\newblock {\em Journal of the ACM (JACM)}, 60(6):1--39.

\bibitem[Huang and Yau, 2019]{huang2019dynamics}
Huang, J. and Yau, H.-T. (2019).
\newblock Dynamics of deep neural networks and neural tangent hierarchy.
\newblock {\em arXiv preprint arXiv:1909.08156}.

\bibitem[Jacot et~al., 2018]{jacot2018neural}
Jacot, A., Gabriel, F., and Hongler, C. (2018).
\newblock Neural tangent kernel: Convergence and generalization in neural
  networks.
\newblock In {\em Advances in neural information processing systems}, pages
  8571--8580.

\bibitem[Lee et~al., 2019]{lee2019wide}
Lee, J., Xiao, L., Schoenholz, S.~S., Bahri, Y., Sohl-Dickstein, J., and
  Pennington, J. (2019).
\newblock Wide neural networks of any depth evolve as linear models under
  gradient descent.
\newblock {\em arXiv preprint arXiv:1902.06720}.

\bibitem[Livni et~al., 2013]{livni2013algorithm}
Livni, R., Shalev-Shwartz, S., and Shamir, O. (2013).
\newblock An algorithm for training polynomial networks.
\newblock {\em arXiv preprint arXiv:1304.7045}.

\bibitem[Livni et~al., 2014]{livni2014computational}
Livni, R., Shalev-Shwartz, S., and Shamir, O. (2014).
\newblock On the computational efficiency of training neural networks.
\newblock In {\em Advances in Neural Information Processing Systems}, pages
  855--863.

\bibitem[Ma et~al., 2016]{ma2016polynomial}
Ma, T., Shi, J., and Steurer, D. (2016).
\newblock Polynomial-time tensor decompositions with sum-of-squares.
\newblock In {\em 2016 IEEE 57th Annual Symposium on Foundations of Computer
  Science (FOCS)}, pages 438--446. IEEE.

\bibitem[Mei et~al., 2018]{mei2018mean}
Mei, S., Montanari, A., and Nguyen, P.-M. (2018).
\newblock A mean field view of the landscape of two-layer neural networks.
\newblock {\em Proceedings of the National Academy of Sciences},
  115(33):E7665--E7671.

\bibitem[Oymak and Soltanolkotabi, 2020]{oymak2020towards}
Oymak, S. and Soltanolkotabi, M. (2020).
\newblock Towards moderate overparameterization: global convergence guarantees
  for training shallow neural networks.
\newblock {\em IEEE Journal on Selected Areas in Information Theory}.

\bibitem[Rotskoff and Vanden-Eijnden, 2018]{rotskoff2018neural}
Rotskoff, G.~M. and Vanden-Eijnden, E. (2018).
\newblock Neural networks as interacting particle systems: Asymptotic convexity
  of the loss landscape and universal scaling of the approximation error.
\newblock {\em arXiv preprint arXiv:1805.00915}.

\bibitem[Sirignano and Spiliopoulos, 2018]{sirignano2018mean}
Sirignano, J. and Spiliopoulos, K. (2018).
\newblock Mean field analysis of neural networks.
\newblock {\em arXiv preprint arXiv:1805.01053}.

\bibitem[Vershynin, 2018]{vershynin2018high}
Vershynin, R. (2018).
\newblock {\em High-dimensional probability: An introduction with applications
  in data science}, volume~47.
\newblock Cambridge University Press.

\bibitem[Vignat and Bhatnagar, 2008]{vignat2008extension}
Vignat, C. and Bhatnagar, S. (2008).
\newblock An extension of wick’s theorem.
\newblock {\em Statistics \& probability letters}, 78(15):2404--2407.

\bibitem[Wei et~al., 2018]{wei2018margin}
Wei, C., Lee, J.~D., Liu, Q., and Ma, T. (2018).
\newblock On the margin theory of feedforward neural networks.
\newblock {\em arXiv preprint arXiv:1810.05369}.

\bibitem[Wei et~al., 2019]{wei2019regularization}
Wei, C., Lee~Jason, D., Liu, Q., and Ma, T. (2019).
\newblock Regularization matters: Generalization and optimization of neural
  nets vs their induced kernel.
\newblock {\em arXiv preprint arXiv:1810.05369}.

\bibitem[Woodworth et~al., 2020]{woodworth2020kernel}
Woodworth, B., Gunasekar, S., Lee, J.~D., Moroshko, E., Savarese, P., Golan,
  I., Soudry, D., and Srebro, N. (2020).
\newblock Kernel and rich regimes in overparametrized models.
\newblock {\em arXiv preprint arXiv:2002.09277}.

\bibitem[Yehudai and Shamir, 2019]{yehudai2019power}
Yehudai, G. and Shamir, O. (2019).
\newblock On the power and limitations of random features for understanding
  neural networks.
\newblock {\em arXiv preprint arXiv:1904.00687}.

\bibitem[Zou and Gu, 2019]{zou2019improved}
Zou, D. and Gu, Q. (2019).
\newblock An improved analysis of training over-parameterized deep neural
  networks.
\newblock In {\em Advances in Neural Information Processing Systems}, pages
  2053--2062.

\end{thebibliography}
